\def\year{2022}\relax
\documentclass[letterpaper]{article} 
\usepackage{aaai22}  
\usepackage{times}  
\usepackage{helvet}  
\usepackage{courier}  
\usepackage[hyphens]{url}  
\usepackage{graphicx} 
\usepackage{natbib}  
\usepackage{caption} 
\DeclareCaptionStyle{ruled}{labelfont=normalfont,labelsep=colon,strut=off} 
\usepackage{amsmath}
\usepackage{booktabs}
\usepackage{cleveref}
\usepackage{todonotes}
\usepackage{amssymb}
\usepackage{subcaption}
\usepackage[detect-all,per-mode=symbol]{siunitx}
\usepackage[ruled,vlined]{algorithm2e}
\newcommand{\sign}{\text{sign}}
\usepackage{todonotes}
\usepackage{wrapfig}

\title{Recursive Reasoning Graph \\for Multi-Agent Reinforcement Learning}

\author{
    Xiaobai Ma\textsuperscript{\rm 1}, 
    David Isele\textsuperscript{\rm 2}, 
    Jayesh K. Gupta\textsuperscript{\rm 1}, 
    Kikuo Fujimura\textsuperscript{\rm 2}, 
    Mykel J. Kochenderfer\textsuperscript{\rm 1}
}
\affiliations{
    \textsuperscript{\rm 1}Stanford University\\
    \textsuperscript{\rm 2}Honda Research Institute US\\
    maxiaoba@stanford.edu, disele@honda-ri.com, jkg@cs.stanford.edu,
    kfujimura@honda-ri.com, mykel@stanford.edu
}

\begin{document}

\maketitle

\begin{abstract}
Multi-agent reinforcement learning (MARL) provides an efficient way for simultaneously learning policies for multiple agents interacting with each other. 
However, in scenarios requiring complex interactions, 
existing algorithms can suffer from an inability to accurately anticipate the influence of self-actions on other agents. 
Incorporating an ability to reason about other agents' potential responses can allow an agent to formulate more effective strategies.
This paper adopts a recursive reasoning model in a centralized-training-decentralized-execution framework to help learning agents better cooperate with or compete against others. 
The proposed algorithm, referred to as the Recursive Reasoning Graph (R2G), shows state-of-the-art performance on multiple multi-agent particle and robotics games.
\end{abstract}

\section{Introduction}
\label{sec:intro}

Recent advances in deep reinforcement learning have shown impressive success in single-agent scenarios including games~\citep{rainbow} and robotics~\citep{johannink2019residual}. 
However, many real-world problems involve interactions between multiple agents with limited information exchange, where multi-agent reinforcement learning (MARL) is needed~\citep{vinyals2019grandmaster,gr2,yang2019cm3}.
The simplest approach to MARL is independent reinforcement learning~\citep{indq}, which trains each agent independently by treating the other learning agents as part of the environment. 
Unfortunately, these methods often have stability issues since the environment dynamics from the perspective of each learning agent is non-stationary due to the learning of other agents~\citep{mohseni2019interaction}. To account for this, we can build behavior models of other agents and thus separate this unstable component out of the environment dynamics. 
Many decentralized MARL algorithms follow this idea~\citep{he2016opponent,pr2,shen2021robust}.

However, modeling other agents' could be difficult as they are continuously learning~\cite{albrecht2018autonomous}. In the \emph{centralized-training-decentralized-execution} (CTDE) framework, such problems are avoided by allowing the learning algorithm to have direct access to all agents' internal information (policy, value network, etc.) at training time. Algorithms adopting this setting often demonstrate better interactive strategies~\citep{vdn,qmix,gupta2017cooperative}.
However, most existing algorithms are limited due to their lack of ability at explicitly considering the reasoning process of other agents~\citep{maddpg,yang2019cm3,masoftq}. They therefore cannot develop plans that account for the influence of their behavior changes on opponents' response~\cite{pr2}.

Reasoning about others' reasoning, referred to as \emph{recursive reasoning}, is important for humans interacting with each other. 
\citet{von2017minds} use recursive nested beliefs to predict the actions of other agents, but their method is limited to few state variables and discrete action spaces.
The PR2 method~\citep{pr2} applies recursive reasoning by explicitly modeling the other agent's response to the ego agent's action. However, as PR2 adopts decentralized-training, each learning agent uses a single model to learn the response of all the opponents by assuming that the other agents share the same reward as itself, which limits its application to fully-cooperative games.
\citet{m3ddpg} augment the multi-agent deep deterministic policy gradient (MADDPG)~\citep{maddpg} using adversarial training with linear approximations to model the minimax optimization, which is limited to zero-sum games.

In this paper, we propose to augment the existing CTDE framework with auxiliary central actors to learn the optimal response of each agent given opponents' actions. Based on this, we build a graph structure to model the recursive reasoning procedure between interacting agents.
This graph structure allows us to model the relationships between agents and explicitly consider the their responses. The recursive actions of each agent are efficiently sampled and shared through message passing in the graph.
The proposed method, which we call the Recursive Reasoning Graph (R2G), works in both competitive and cooperative games. Our contributions are as follows:

\begin{itemize}
    \item We propose R2G, a multi-agent reinforcement learning framework that explicitly models the recursive reasoning process of the agents in general-sum games.
    \item We augment the existing centralized-training-decentralized-execution algorithms with centralized actors and graph-like message passing to efficiently train the learning agents under R2G framework.
    \item We demonstrate state-of-the-art performance on multiple Particle World~\citep{maddpg} and RoboSumo~\citep{robosumo} environments with complex reward structure and non-trivial equilibrium.
\end{itemize}

\section{Background}

\subsection{Markov Game}
\label{sec:markov_game}
A Markov Game (MG) is commonly used to model multi-agent reinforcement learning problems~\citep{MG}. A MG is specified by $(S,\{A^i\}^n_{i=1},T,\{r^i\}^n_{i=1},\gamma,s_0)$, where $n$ is the number of agents; $S$ is the state space containing the state for all agents; $A^i$ represents the action space for agent $i$;  $T: S \times \prod^n_{i=1}A^i \rightarrow S$ represents the (stochastic) transition model conditioned on the current state as well as the actions of all agents; $r^i: S \times \prod^n_{i=1}A^i \times S \rightarrow \mathbb{R}$ represents the reward for agent $i$; and $s_0$ represents the initial state distribution of all agents. 

The learning objective in MGs is to get a set of polices $\{\pi^i\}^n_{i=1}$, where for each agent $i$, $\pi^i: S \rightarrow A^i$ maps the state to its action. However, unlike the clear optimization goal in the single-agent reinforcement learning, the concept of optimality in MARL is more complex. In fully cooperative games where all the agents share the same reward function, the optimization goal is to maximize the joint return for all agents. When agents have conflict of interests, the concept of equilibrium is introduced. A common objective is to find the Nash Equilibrium (NE), where all agents act in best response to each others' current strategy. However, a Nash Equilibrium might be difficult to compute and can diverge from human behavior~\citep{wright2010beyond}. 

\subsection{Multi-Agent Actor-Critic}
\label{sec:ma_ac}
The actor-critic framework is a common training structure in single-agent reinforcement learning. 
In this framework, a critic, $Q_\theta(s,a)$, is trained to estimate the return value of the state-action pair $(s,a)$ with the loss 
$J_{Q_\theta}=\mathbb{E}_{s,a\sim\mathcal{D}}[(Q_\theta(s,a)-\hat{Q})^2]$, where $\mathcal{D}$ is the replay buffer storing the exploration experiences and $\hat{Q}$ is an empirical estimate of the return value.
An actor, $\pi_\phi(s)$, is trained to maximize the return value with the loss
$J_{\pi_\phi}=\mathbb{E}_{s\sim\mathcal{D},a\sim\pi_\phi(s)}[-Q_\theta(s,a)]$.
Additional terms like the policy entropy could also be added to $J_{\pi_\phi}$ to improve the training~\citep{sac}.
The actor-critic framework could be naturally generalized to the multi-agent setting with centralized training. 
A central critic, $Q^i_\theta(s,a^i,a^{-i})$, for each agent $i$, where $a^{-i}$ indicates the actions of agents except agent $i$, is trained to estimate the return value of agent $i$ given the state and the joint-action; i.e., $J_{Q^i_\theta}=\mathbb{E}_{s,a^i,a^{-i}\sim\mathcal{D}}[(Q_\theta(s,a^i,a^{-i})-\hat{Q})^2]$.
Each actor, $\pi^i_\phi(s)$, is then trained to minimize the loss $J_{\pi^i_\phi}=\mathbb{E}_{s\sim\mathcal{D},a^i\sim\pi^i_\phi(s)}[-Q_\theta(s,a^i,a^{-i})]$.

There are multiple choices for sampling $a^{-i}$ during the training of actors.
For example, in MADDPG~\citep{maddpg}, $a^{-i}$ is from the stored experiences in the replay buffer. The stored actions are sampled from the policies earlier in the training or from an exploration strategy. In this case, the actor is actually learning the best response with respect to the action distribution stored in the replay buffer, whose performance largely depends on the exploration strategy.

An alternative is to sample $a^{-i}$ directly from the other agents' current policies. 
Unfortunately, this can lead to the problem of relative overgeneralization~\citep{wei2016lenient}.
During training and exploration, a suboptimal Nash Equilibrium (NE) could be preferred over the optimal NE, when each agent's action in the suboptimal NE is estimated with a higher return against the current action distributions from the other agents. 
\citet{masoftq} address this problem by having each agent learn the optimal joint action of all agents. However, this assumes that all agents are optimizing the same reward function and thus is limited to fully cooperative games.

Sampling from the other agents' current policy could also lead to oscillatory learning as all agents learn concurrently, and the best response at this iteration might be suboptimal in the next iteration. For example, in the \textsc{Rock-Paper-Scissors} game, if we know the opponent's current policy is playing \textsc{Rock} with probability 1, then our best response is to play \textsc{Paper}. However, when the opponent finds out our new policy at the next iteration, they would change to play \textsc{Scissor}. As one player changes its policy completely at each iteration, the equilibrium where each player plays the three options uniformly randomly could never be reached. 

\section{Methods}

We propose to use a recursive reasoning model to sample the opponents' response during policy training in multi-agent actor-critic.

\subsection{Logit Level-K Reasoning}
\label{sec:recursive_reasoning}
The recursive reasoning refers to the process of reasoning about the other agent's reasoning process during decision making. 
It allows the ego agent to consider the potential change in strategy of other agents instead of treating them as fixed. 
A classic model of recursive reasoning is the logit level-$k$ model: At level 0, all agents choose their actions based on some base policies, $\pi^{(0)}$. At each level $k$, each agent chooses the best policy by assuming the others follow their level $k-1$ policies. In multi-agent RL, a natural level-0 policy is the agent's current policy, i.e. $\pi^{i,(0)}=\pi^i$. Given the actions of other agents at level $k-1$: $a^{-i,(k-1)}$, the best level-$k$ action for agent $i$ should be 
\begin{equation}
\label{eq:aik*}
a^{i,(k)} = \arg\max_{a^i}Q^i(s,a^i,a^{-i,(k-1)})
\end{equation}
where $Q^i$ is the estimated return of agent $i$. This formulation holds for general-sum games.

Solving the optimization in \cref{eq:aik*} is not trivial in continuous action spaces. Thus, we introduce a central actor $\pi^i_c(s,a^{-i})$ which learns the best response for agent $i$ given state $s$ and the other agents' actions $a^{-i}$. 
We train $\pi^i_{c,\psi}(s,a^{-i})$ to approximate $\arg\max_{a^i} Q^i_\theta(s,a^i,a^{-i})$ by minimizing the loss:
\begin{equation}
\label{eq:Jpic}
    J_{\pi^i_{c,\psi}}=\mathbb{E}_{s,a^{-i}\sim \mathcal{D},a^i\sim \pi^i_{c,\psi}(s,a^{-i})}[-Q_\theta^i(s,a^i,a^{-i})]
\end{equation}

\subsection{Recursive Reasoning Graph}
With the help of $\pi^i_c$, we can calculate $a^{-i,(k)}$ using a message passing process in a recursive reasoning graph (R2G): $\mathcal{G} = (\mathcal{V},\mathcal{E})$. The node set $\mathcal{V}=\{\pi^1_c,...,\pi^n_c\}$ contains the central actor node for each agent, and the edge set $\mathcal{E}$ contains edges between all interacting agents. We use an undirected, fully-connected graph 
by assuming all agents are interacting with each other. A more sparse graph could be used if we have prior knowledge on the interacting structure between agents. The messages in the edges are the sampled actions $a^{i,(k)}$ from the central actors. 

\begin{figure*}
  \centering
  \includegraphics[width=1.6\columnwidth]{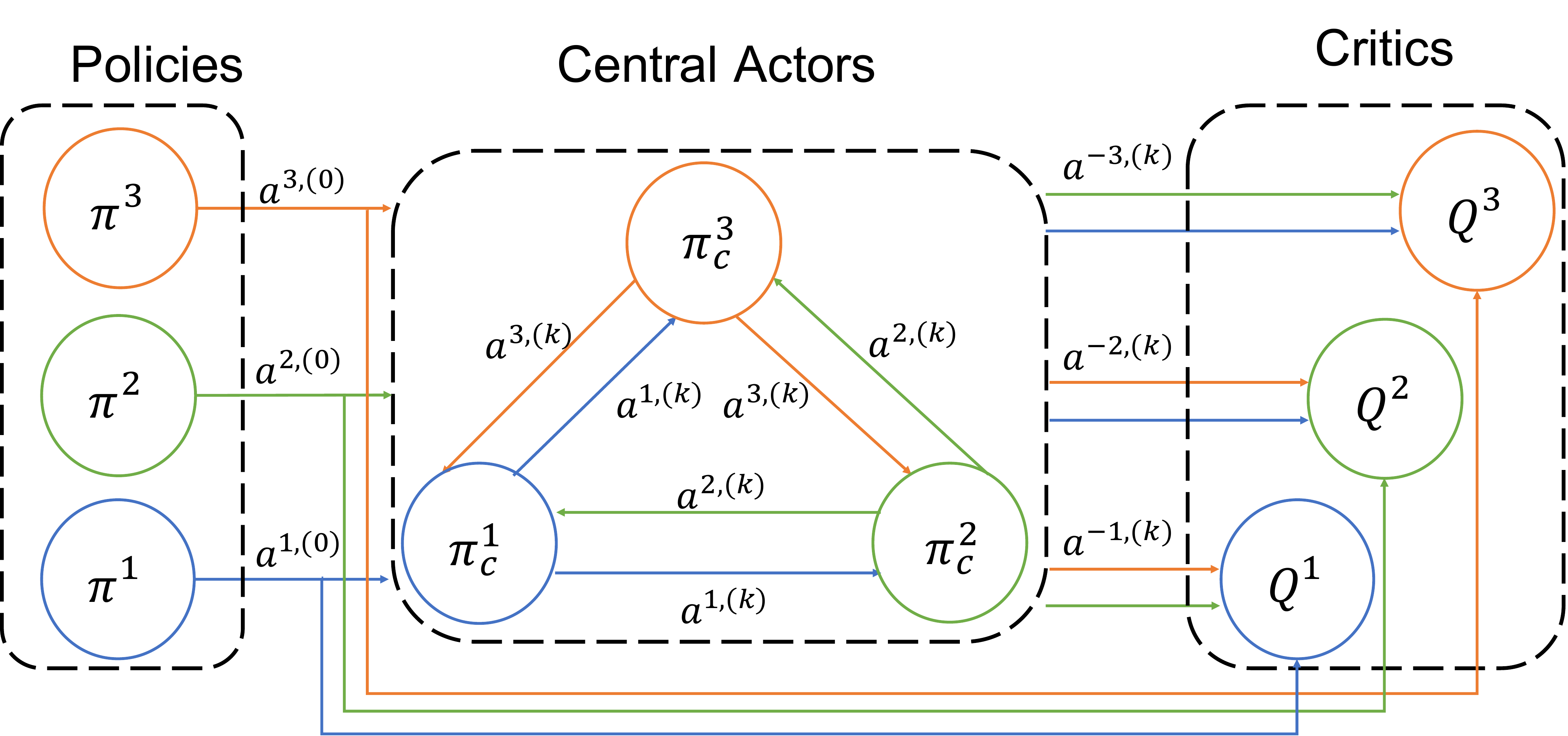}
  \caption{Recursive Reasoning Graph: The forward computation graph for Q values in the policy loss in a 3-agent game.}
  \label{fig:r2g}
\end{figure*}

\Cref{fig:r2g} gives an illustration of the recursive reasoning graph in a 3-agent game.
The initial level-0 actions are sampled from the individual policies:
\begin{equation}
\label{eq:ai0}
    a^{i,(0)} \sim \pi^i(s)
\end{equation}
At each level $k\ge 1$, we have:
\begin{equation}
\label{eq:aik}
    a^{i,(k)} \sim \pi^i_c(s,\textsc{Agg}_{j\in\mathcal{N}(i)}a^{j,(k-1)})
\end{equation}
where $\textsc{Agg}$ is the aggregation function and $\mathcal{N}$ is the node neighborhood function. In practice, we use concatenation for $\textsc{Agg}$. Thus, $\textsc{Agg}_{j\in\mathcal{N}(i)}a^{j,(k-1)}$ is interchangeable with $a^{-i}$ in fully-connected graphs. 
The generalization from fully-connected graph to sparse graph is straightforward by limiting the message passing between central actors. 
An attention mechanism~\cite{gat} could also be used to dynamically learn the interaction structures.

At level $k$, each central actor node takes the input message of $a^{-i,(k-1)}$ and outputs its best response $a^{i,(k)}$. Thus, one complete message passing through the graph gives one-level up in the recursion. 
Hypothetically, the level of recursion could go to infinity. In the following discussion, we focus on the level-1 recursion. The comparison of different recursion levels are provided in the appendix.

The output level $k$ actions are then fed to the central critics to calculate the estimated Q-values of the policy actions.
The policy loss, $J_{\pi^i_\phi}$, is then formulated as the KL-divergence of the policy action distribution to the energy-based distribution represented by $Q^i_{\theta}$:
\begin{equation}
\label{eq:Jpi}
    \begin{split}
        J_{\pi^i_\phi} = \mathbb{E}_{a^{i,(0)}\sim \pi^i_{\phi},a^{-i,(k)}\sim \mathcal{G},s\sim\mathcal{D}}[\alpha^i \log(\pi^i_\phi(a^{i,(0)}|s))\\ -Q_\theta^i(s,a^{i,(0)},a^{-i,(k)})]
    \end{split}
\end{equation}
where $\alpha^i$ is the temperature variable trained similarly as in SAC~\citep{sac}.

For the training of the central critic $Q_\theta^i$, we adopt the soft Bellman residual~\citep{sac}:
\begin{equation}
\label{eq:JQ}
    J_{Q_\theta^i}=\mathbb{E}_{\mathcal{D}}
    [(Q_\theta^i(s,a^i,a^{-i})-(r^i(s,a^i,a^{-i})+\gamma \hat{V}(s')))^2]
\end{equation}
where the next state value $\hat{V}(s')$ is estimated by
\begin{equation}
    \begin{split}
        \hat{V}(s') = \mathbb{E}_{a'^{i,(0)}\sim\pi_{\phi}^i,a'^{-i,(k)}\sim\mathcal{G}}[Q^i_{\hat{\theta}}(s',a'^{i,(0)},a'^{-i,(k)})\\ -\alpha^i \log \pi_\phi^i(a'^{i,(0)}|s')]
    \end{split}
\end{equation}
where $Q^i_{\hat{\theta}}$ is the delayed updated version of the critic network for agent $i$.

The central actor $\pi^i_{c,\psi}$ is trained by the loss given in \cref{eq:Jpic}.
We can omit the entropy term in $J_{\pi^i_{c,\psi}}$ since the exploration does not rely on $\pi^i_{c,\psi}$, and there always exists a deterministic optimal response given the strategies of other agents.

The training process is outlined in algorithm~\ref{alg:r2g}. The convergence of the algorithm are discussed in the appendix.

\begin{algorithm}
\SetAlgoLined
\KwResult{Policy $\pi^i_\phi$, $\forall i\in 1,..,n$}
 Initialize $\pi^i_\phi$, $Q^i_\theta$, $Q^i_{\hat{\theta}}$, and $\pi^i_{c,\psi}$ $\forall i\in 1,..,n$\;
 $\mathcal{D}\leftarrow$ empty reply buffer\;
 \For{each epoch}{
  Collect exploration experiences using $\pi^{i=1:n}_\phi$\;
  Add tuples $(s,a^{i=1:n},r^{i=1:n},s')$ to $\mathcal{D}$\;
  \For{each training iteration}{
    Sample $\{(s_j,a^{i=1:n}_j,r^{i=1:n}_j,s'_j)\}_{j=1}^B$ from $\mathcal{D}$ with batch-size $B$\;
    Calculate $\{a_j^{i=1:n,(k)}\}_{j=1}^B$ at $\{s_j\}_{j=1}^B$ and $\{{a'_j}^{i=1:n,(k)}\}_{j=1}^B$ at $\{s'_j\}_{j=1}^B$ from \cref{eq:aik}\;
    \For{each agent i}{
        Update $\pi^i_\phi$ with $J_{\pi^i_\phi}$ in \cref{eq:Jpi}\;
        Update $\pi^i_{c,\psi}$ with $J_{\pi^i_{c,\psi}}$ in \cref{eq:Jpic}\;
        Update $Q^i_\theta$ with $J_{Q^i_\theta}$ in \cref{eq:JQ}\;
        Update target Q network as $\hat{\theta}\leftarrow \tau\theta+(1-\tau)\hat{\theta}$ with factor $\tau$\;
    }
  }
 }
 \caption{Recursive Reasoning Graph (R2G)}
 \label{alg:r2g}
\end{algorithm}

The relative over-generalization problem in R2G is mitigated as learning the conditional best response in $\pi^i_{c}(s,a^{-i})$ is much easier than learning the marginal best response in $\pi^i(s)$, which makes the recursive actions a better approximation to the opponents' optimal response.
Let $\pi^{i,(0)}$ denote the policy trained with opponents' current strategies, where $J_{\pi^{i,(0)}}=\mathbb{E}_{s\sim\mathcal{D},a^{-i}\sim\pi^{-i,(0)}_\phi(s),a^i\sim\pi^i_\phi(s)}[-Q_\theta(s,a^i,a^{-i})]$. We first notice that $J_{\pi^{i,(0)}}$ has a much larger variance than $J_{\pi^i_{c}}$ in \cref{eq:Jpic} due to the additional variance introduced by $a^{-i}\sim\pi^{-i,(0)}(s)$, where $\pi^{-i,(0)}$ could be both stochastic and learning. 
Without loss of generality, let us consider the influence of using $a^{2,(1)}$ in the training of $\pi^1$.
In a two-player game, from the above argument, $a^{2,(1)}\sim\pi^2_c(s,a^1)$ is a better approximation to the optimal response of player 2 to $a^1$ than $a^{2,(0)}\sim\pi^{2,(0)}(s)$.
In games with more than 2 players, the above argument is not as clear, since $a^{2,(1)}\sim\pi^2_c(s,a^{-i})$ is best responding not only $a^1$ but also other players. However, in most games, the importance of different players is mutual, i.e., if the strategy of player 2 is important to player 1, then so is player 1 to player 2.
Thus, if $a^{2,(1)}$ is important for the training of $\pi^1$, then $\pi^2_c(s,a^{-i})$ should also consider $a^1$ more than other players' actions.
Thus, $Q_\theta^i(s,a^{i,(0)},a^{-i,(1)})$ gives a generally closer approximation than $Q_\theta^i(s,a^{i,(0)},a^{-i,(0)})$ on the return of the ego agent's action when opponents response optimally.

The oscillatory learning problem is also avoided by training $\pi^i$ with $a^{-i,(1)}$. The change of other agents' actions due to the ego agent's action change is accounted in the policy loss using recursive actions.


A critical concern for an MARL algorithm is its scalability with respect to the number of agents, $n$.
The computation complexity of most of the centralized-training-decentralized-execution approaches scales quadratically with the number of agents: each agent is trained through a centralized critic taking inputs of joint actions whose forward and backward propagation scales linearly with respect to $n$ (for general neural network structures).
While R2G adds an additional centralized component, the central actor, for each agent, the overall scalability does not degrade much as recursive actions are shared during policy training.
By the message passing mechanism, the recursive action for each agent is only calculated once at each recursion level and is used repeatedly for $n-1$ opponents.
Thus, the overall training still scales quadratically with $n$ and linearly with the recursion level, $k$.

\section{Experiments}
We compare R2G with several baselines spanning centralized and decentralized learning, as well as on-policy and off-policy algorithms applicable in the continuous action apace context:
\begin{itemize}
    \item PPO: The proximal policy optimization (PPO)~\cite{ppo} is one of the most successful on-policy policy optimization algorithm for single-agent continuous control. In the experiments, PPO is applied to multi-agent problems through independent learning.
    \item COMA: The counterfactual multi-agent policy gradients (COMA)~\citep{coma} is originally developed for multi-agent on-policy learning in discrete action spaces. To extend it to continuous control problems, we use a surrogate loss and likelihood ratio clipping similar as in PPO, as well as a Monte Carlo estimation for the counterfactual baseline.
    \item SAC: Soft actor-critic~\cite{sac} is a single-agent off-policy learning algorithm. Similarly as PPO, an independent learning scheme is applied for multi-agent problems.
    \item MADDPG: The multi-agent deep deterministic policy gradient (MADDPG)~\citep{maddpg} is a generalization from DDPG~\citep{ddpg} by introducing the central critic. In MADDPG, the $a^{-i}$ is sampled from the replay buffer.
    \item MASAC: MASAC is a similar generalization from SAC as MADDPG. During the training, the $a^{-i}$ is sampled from the current policies of the other agents. Thus, MASAC could be regarded as R2G without the recursive reasoning.
    \item PR2: Probabilistic recursive reasoning (PR2)~\citep{pr2} is a decentralized off-policy learning algorithm which also uses recursive reasoning in its training. The derivation of PR2 is based on the fully-cooperative assumption, but it also shows decent performance on general-sum games.
\end{itemize}

\subsection{Differential Games}
\label{sec:differential}
\begin{figure}[!ht]
  \centering
  \begin{subfigure}{0.4\columnwidth}
  \centering
    {\includegraphics[width=\columnwidth]{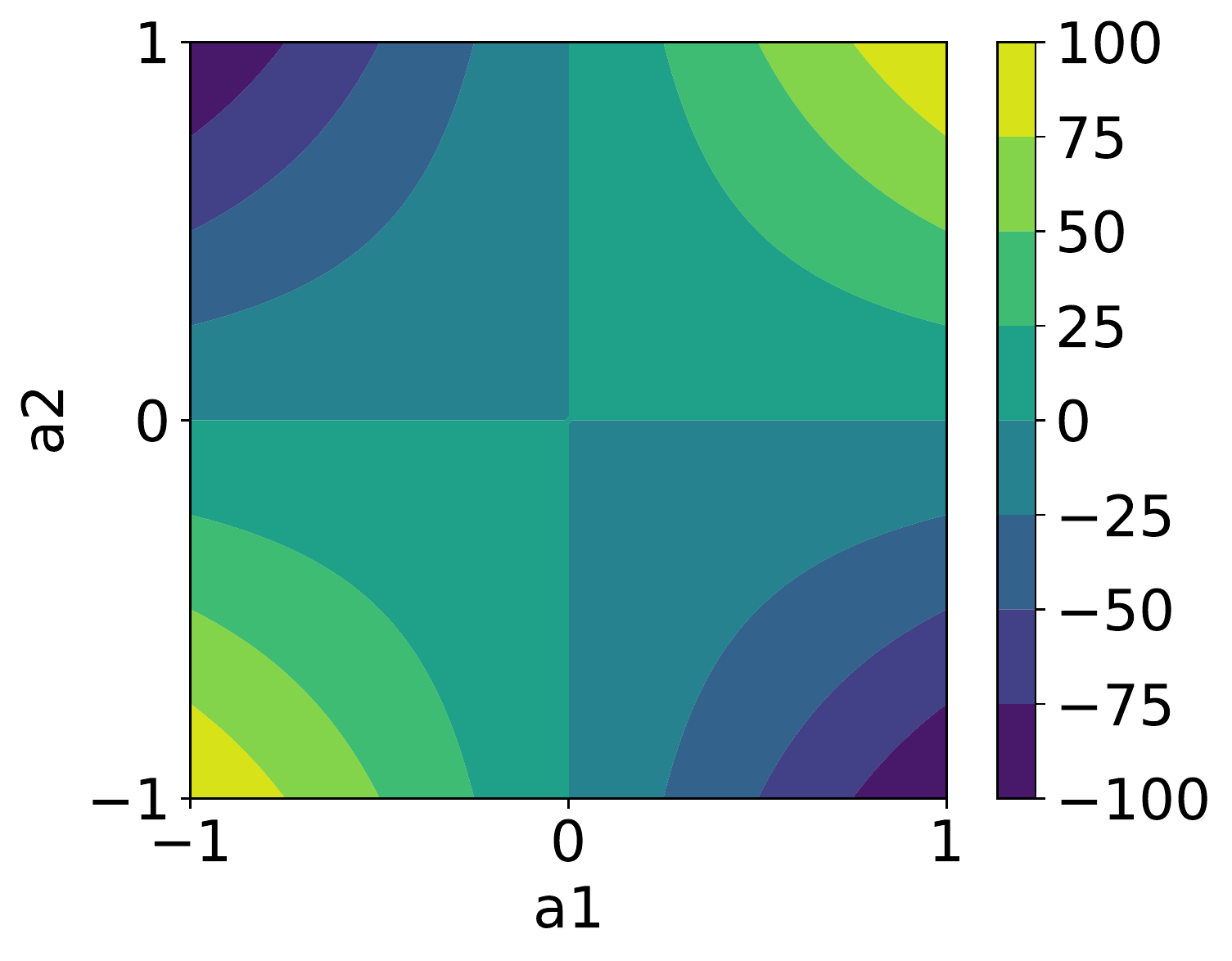}}
    \caption{Zero Sum ($r^1=-r^2$)}\label{fig:zero_sum_reward}
  \end{subfigure}
  \begin{subfigure}{0.4\columnwidth}
  \centering
    {\includegraphics[width=\columnwidth]{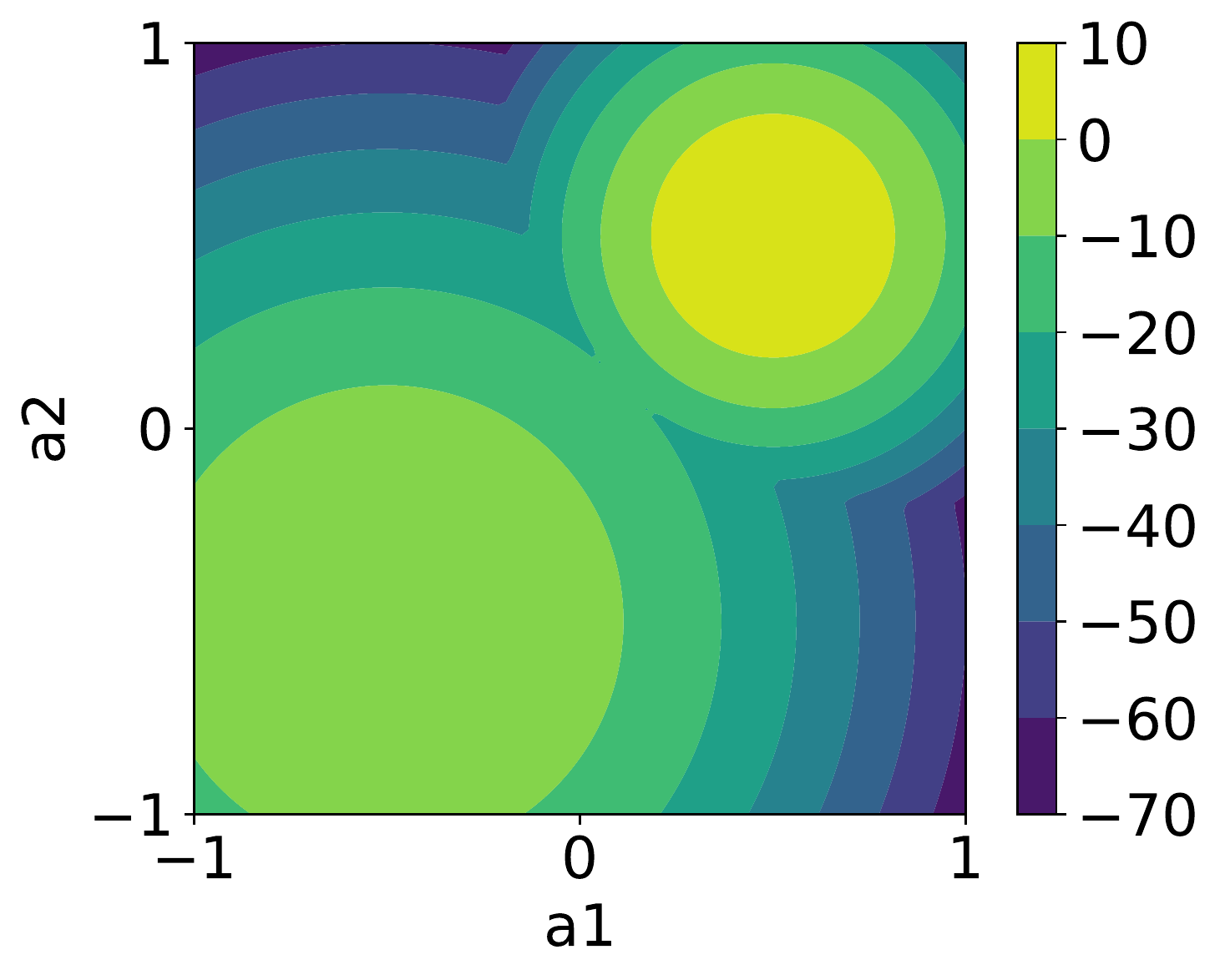}}
    \caption{Max of Two ($r^1=r^2$)}\label{fig:max2_reward}
  \end{subfigure}
  \vspace{-0.2cm}
  \caption{Differential Game: Reward landscapes}
\end{figure}

We first demonstrate the advantage of R2G analytically by focusing on two-player single-state Differential Games: the Zero Sum and the Max of Two (as in \citet{pr2}).
The reward landscapes of the two games are shown in \cref{fig:zero_sum_reward,fig:max2_reward}.
More details of the environment as well as the experiments are given in the appendix.

\begin{figure}[!ht]
  \centering
  \begin{subfigure}{0.22\columnwidth}
  \centering
    {\includegraphics[width=1.0\columnwidth]{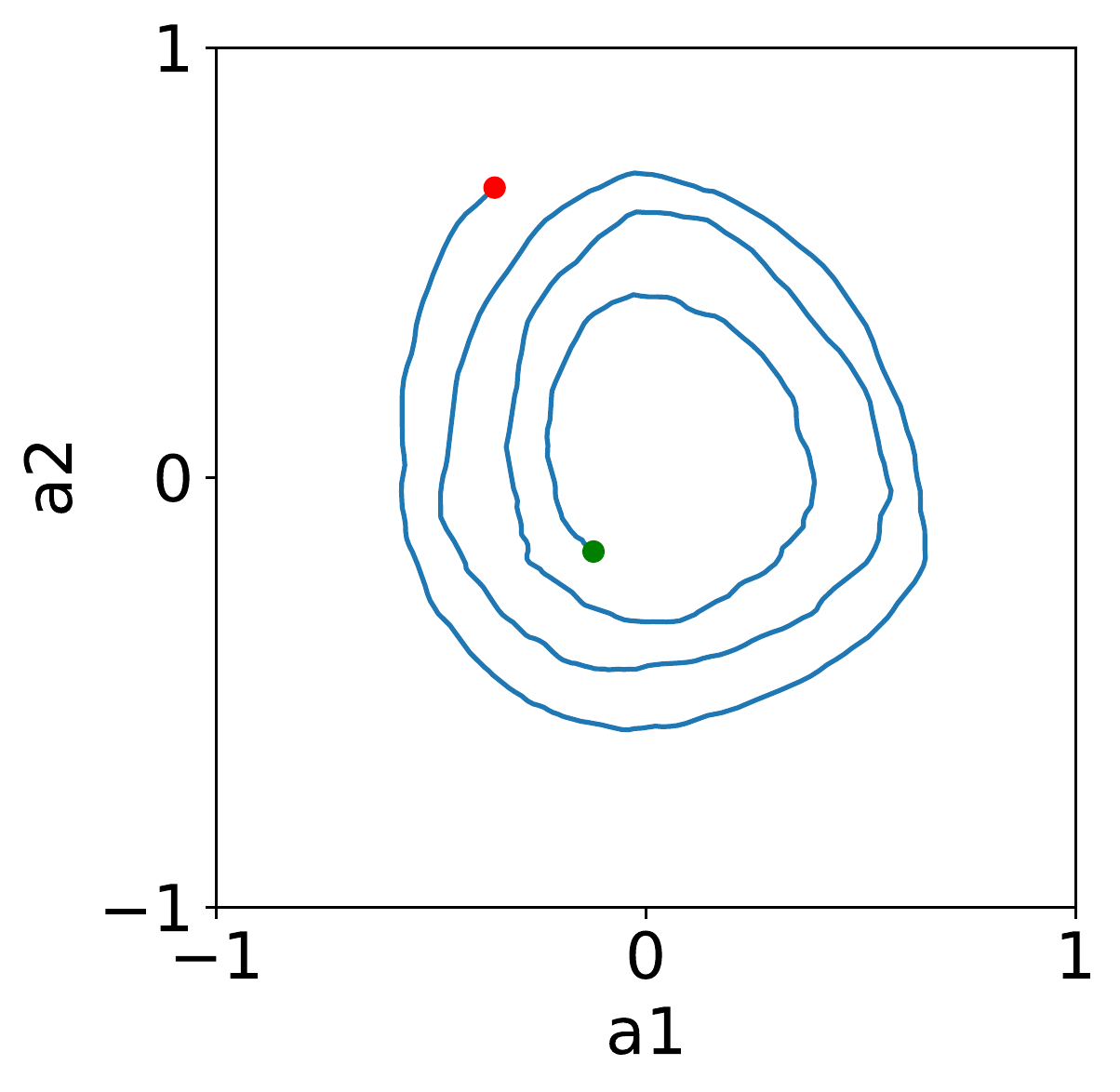}}
    \caption{PPO}\label{fig:zero_sum_ppo}
  \end{subfigure}
  \begin{subfigure}{0.22\columnwidth}
  \centering
    {\includegraphics[width=1.0\columnwidth]{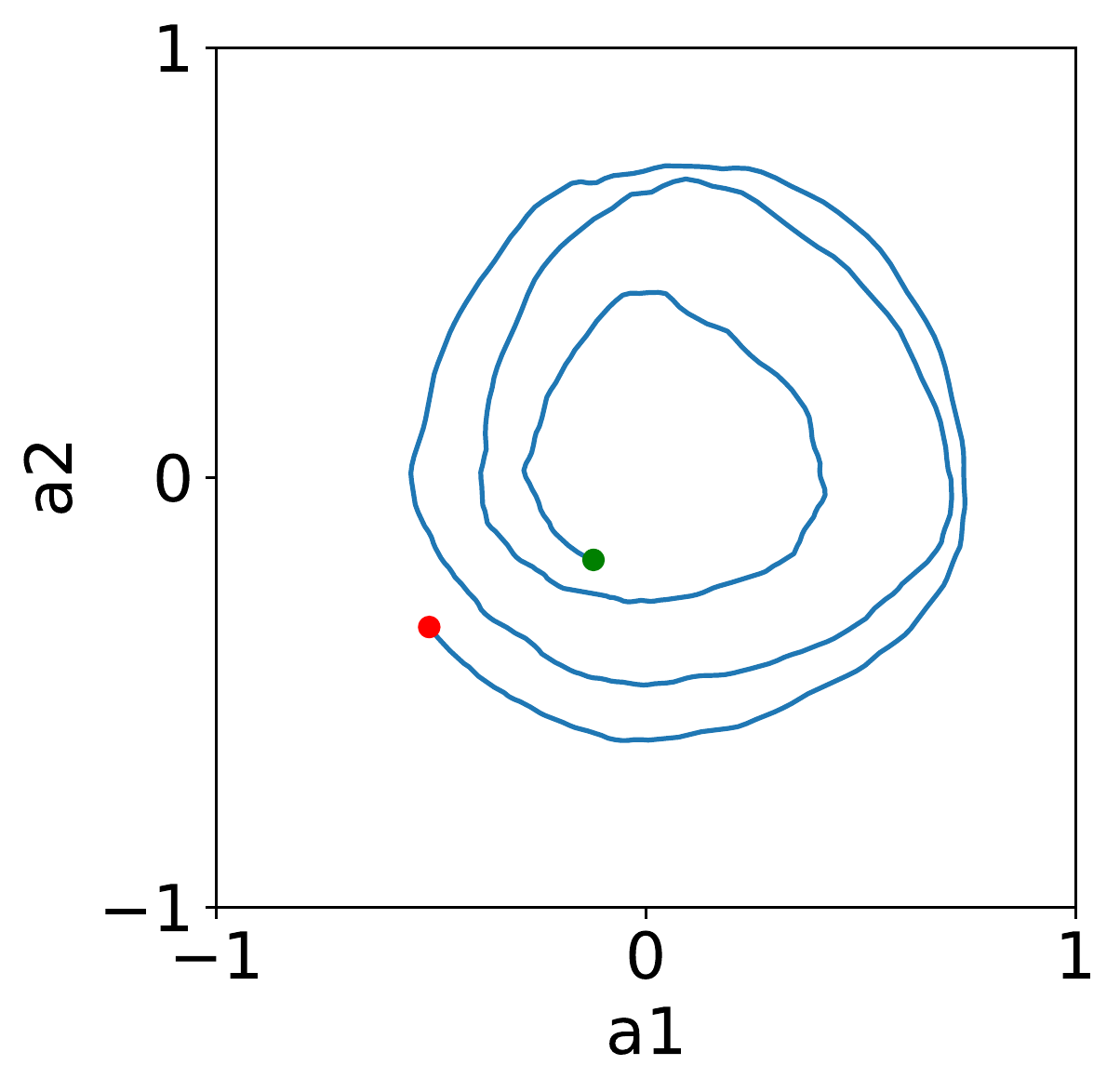}}
    \caption{COMA}\label{fig:zero_sum_coma}
  \end{subfigure}
  \begin{subfigure}{0.22\columnwidth}
  \centering
    {\includegraphics[width=1.0\columnwidth]{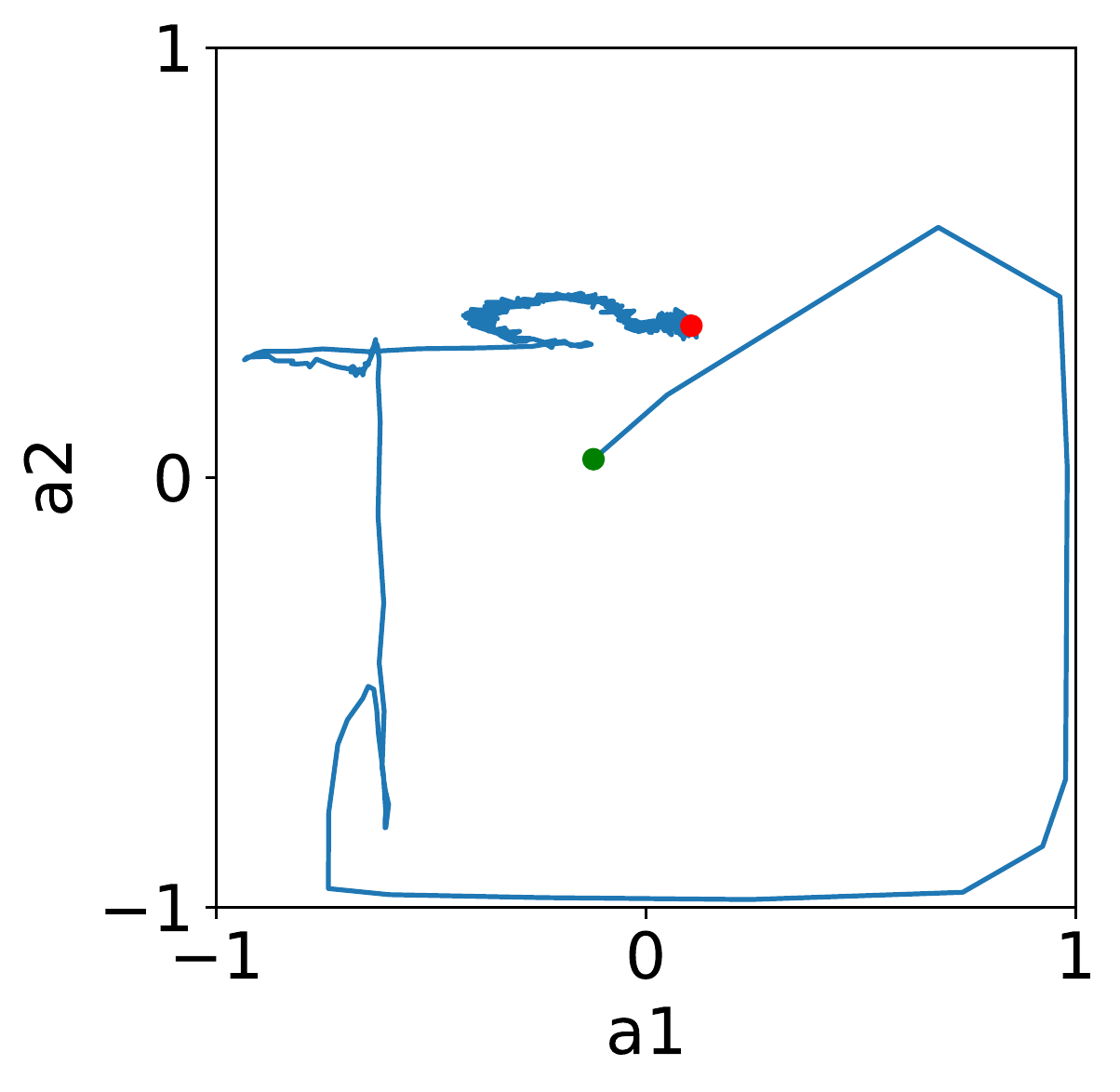}}
    \caption{SAC}\label{fig:zero_sum_sac}
  \end{subfigure}
  \begin{subfigure}{0.22\columnwidth}
  \centering
    {\includegraphics[width=1.0\columnwidth]{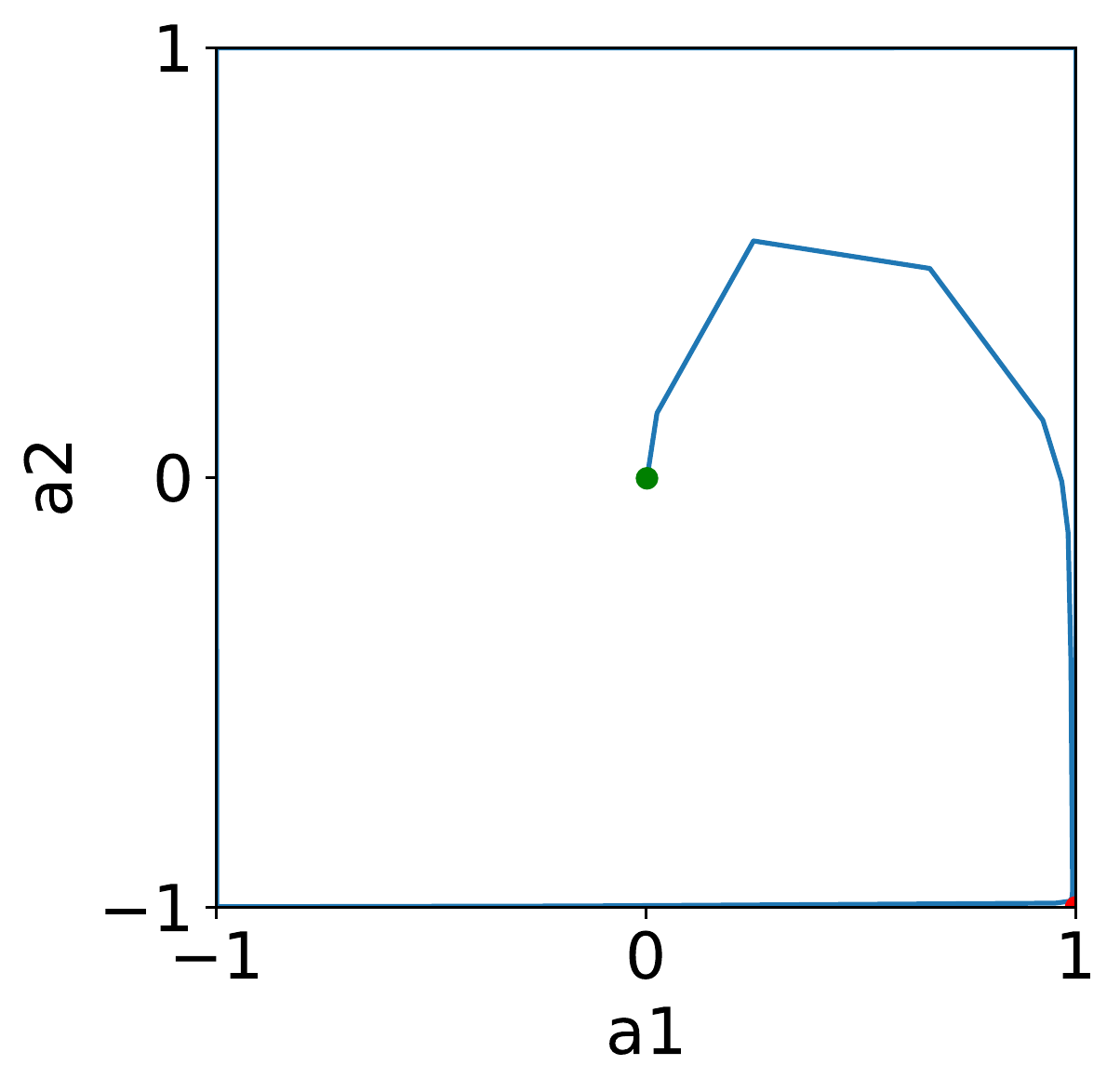}}
    \caption{MADDPG}\label{fig:zero_sum_maddpg}
  \end{subfigure}
  \begin{subfigure}{0.22\columnwidth}
  \centering
    {\includegraphics[width=1.0\columnwidth]{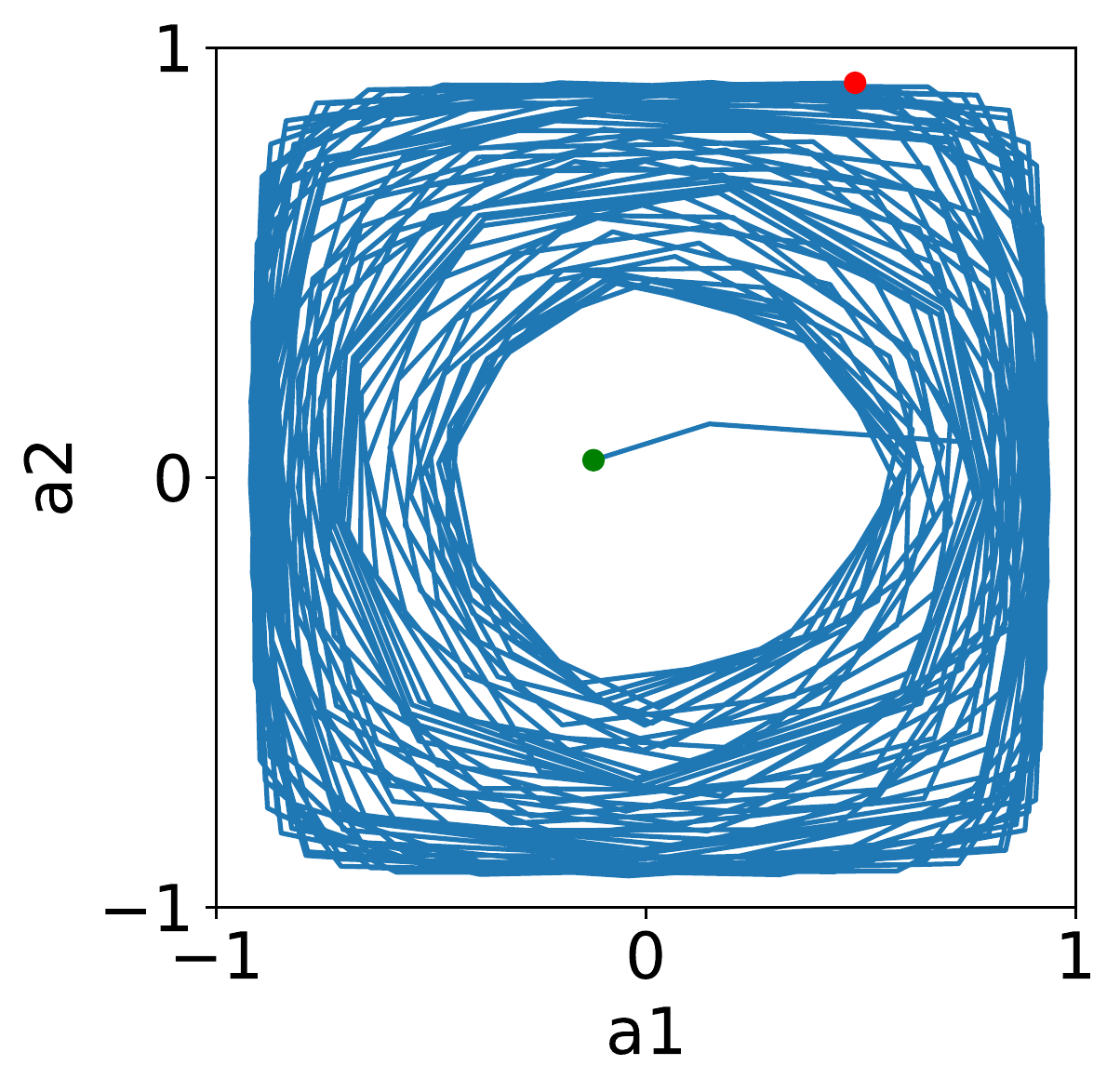}}
    \caption{MASAC}\label{fig:zero_sum_masac}
  \end{subfigure}
  \begin{subfigure}{0.22\columnwidth}
  \centering
    {\includegraphics[width=1.0\columnwidth]{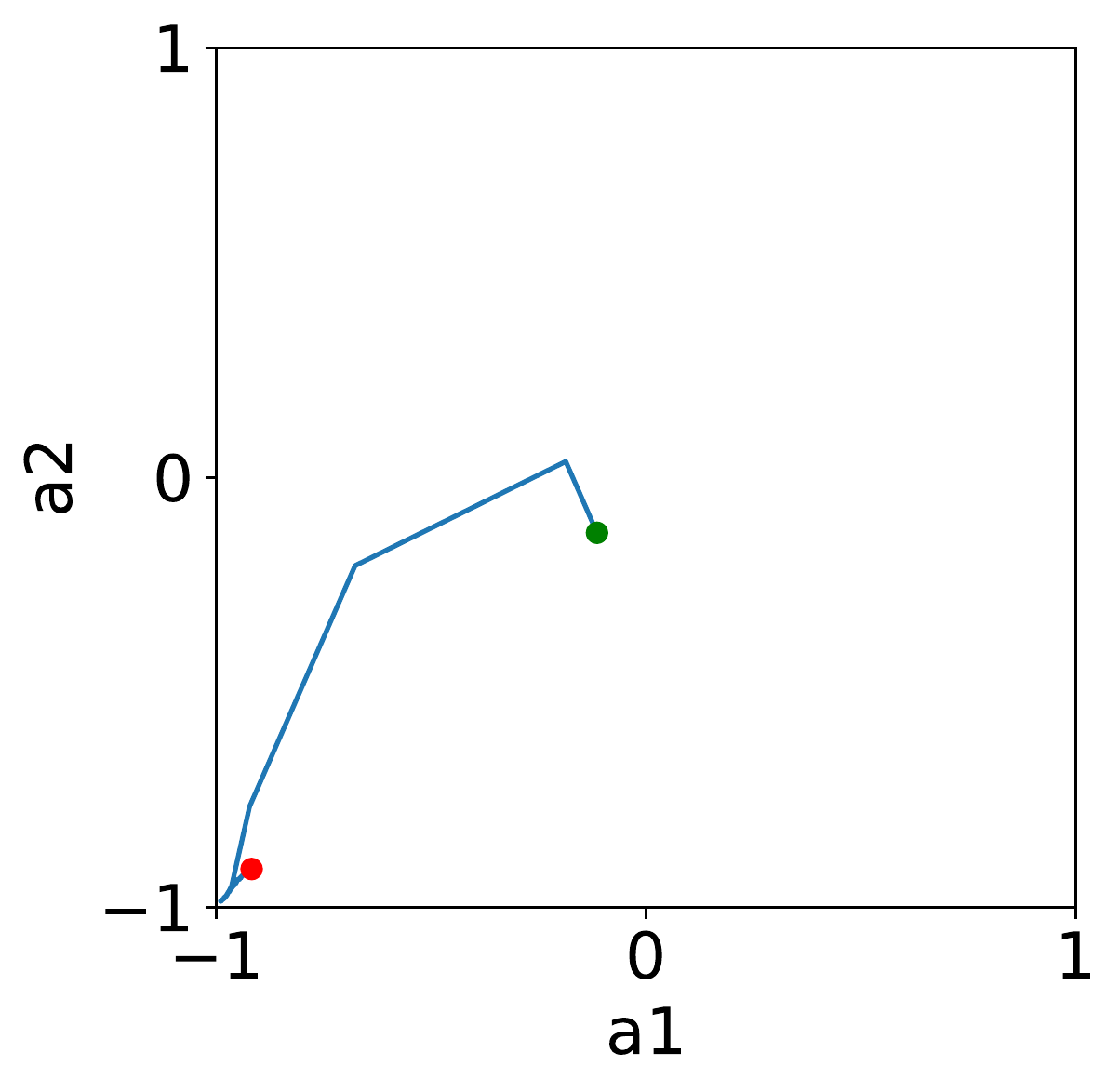}}
    \caption{PR2}\label{fig:zero_sum_pr2}
  \end{subfigure}
  \begin{subfigure}{0.22\columnwidth}
  \centering
    {\includegraphics[width=1.0\columnwidth]{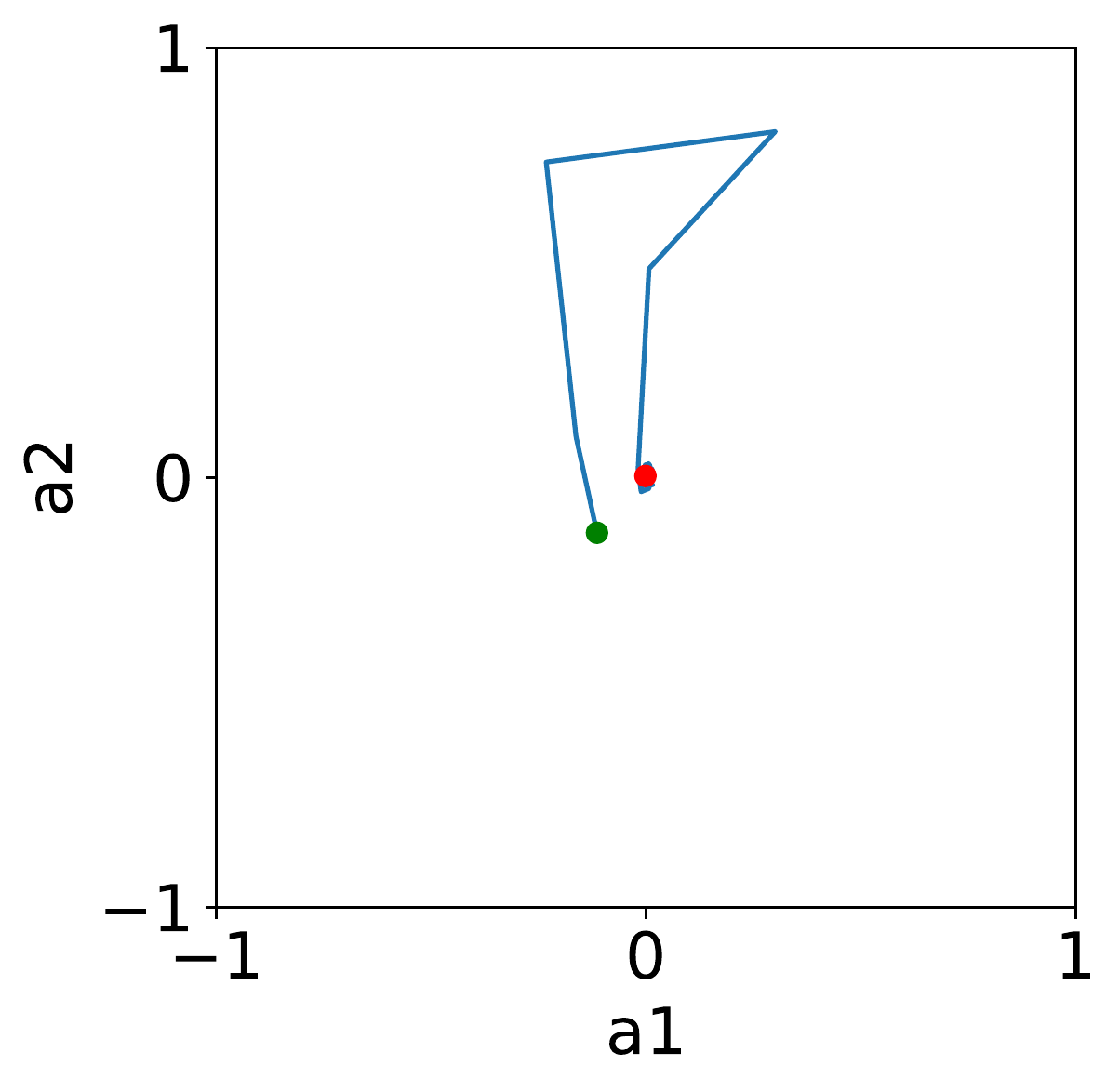}}
    \caption{R2G}\label{fig:zero_sum_r2g}
  \end{subfigure}
  \vspace{-0.2cm}
  \caption{Zero Sum: Trajectories of most likely actions for different algorithms over 1000 iterations.}
  \vspace{-0.3cm}
  \label{fig:zero_sum}
\end{figure}

\begin{figure}[!ht]
  \centering
  \begin{subfigure}{0.45\columnwidth}
  \centering
    {\includegraphics[width=0.48\columnwidth]{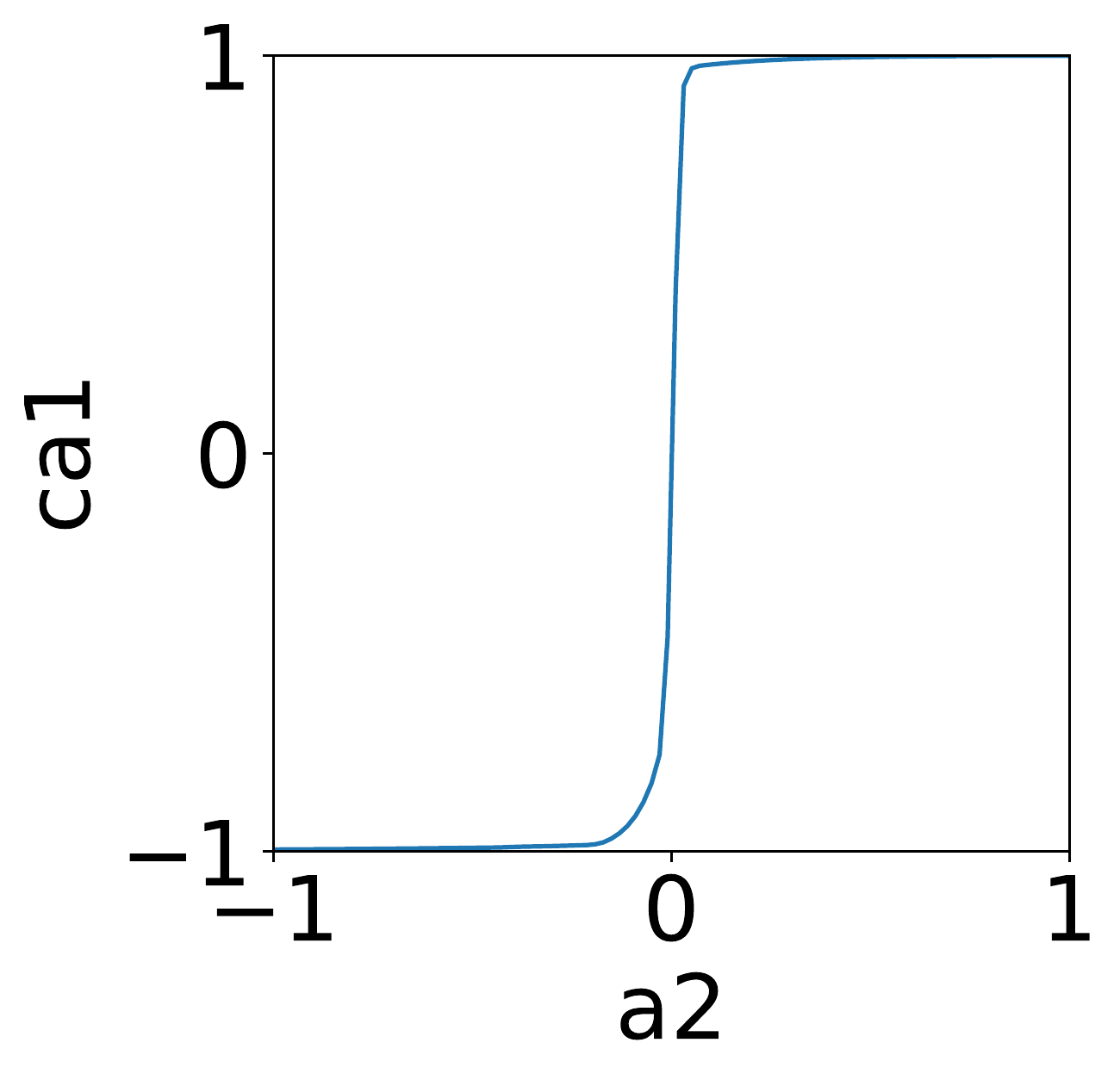}}
    {\includegraphics[width=0.48\columnwidth]{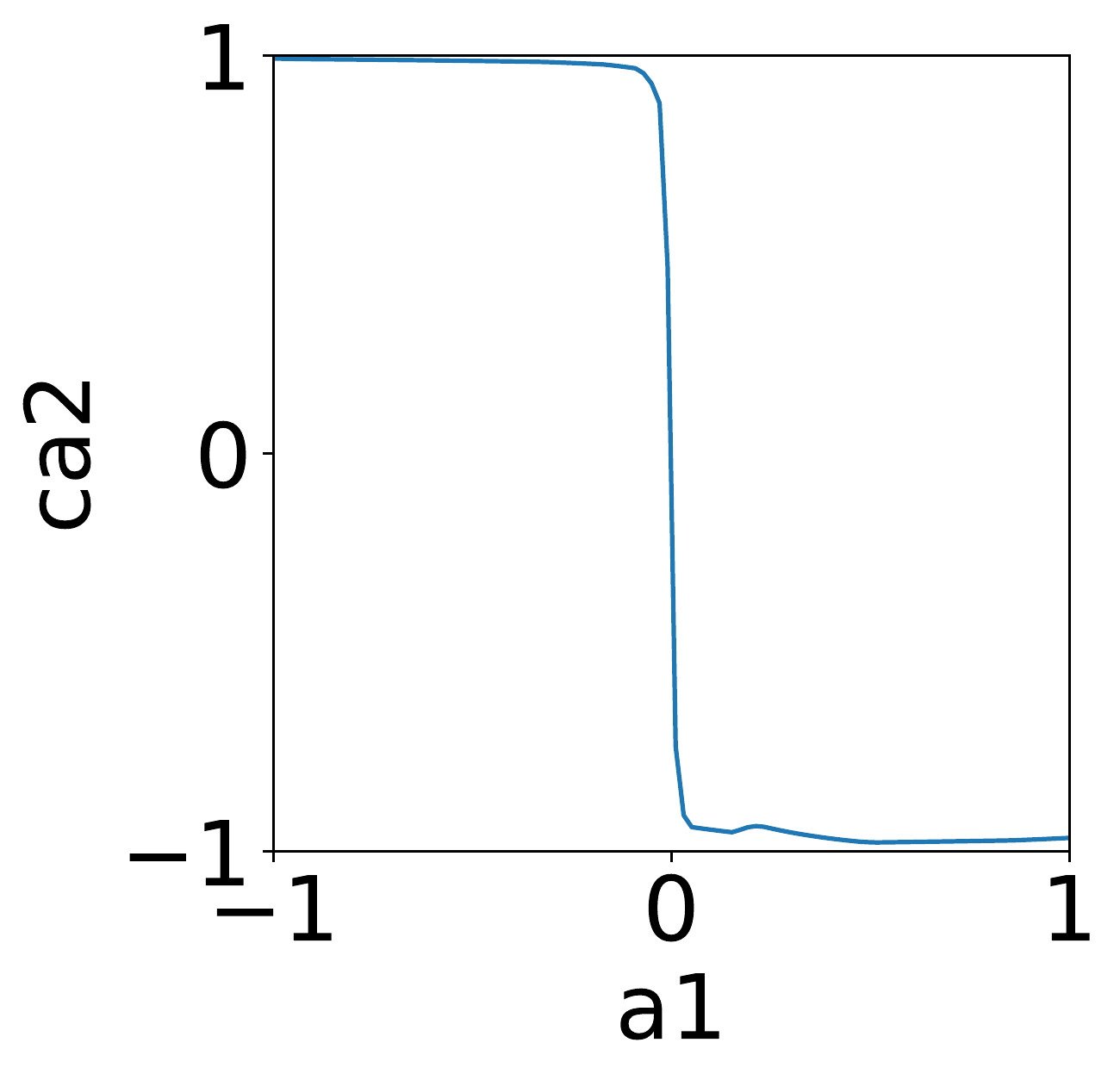}}
    \vspace{-0.25cm}
    \caption{Zero Sum}\label{fig:zero_sum_ca}
  \end{subfigure}
  \begin{subfigure}{0.45\columnwidth}
  \centering
    {\includegraphics[width=0.48\columnwidth]{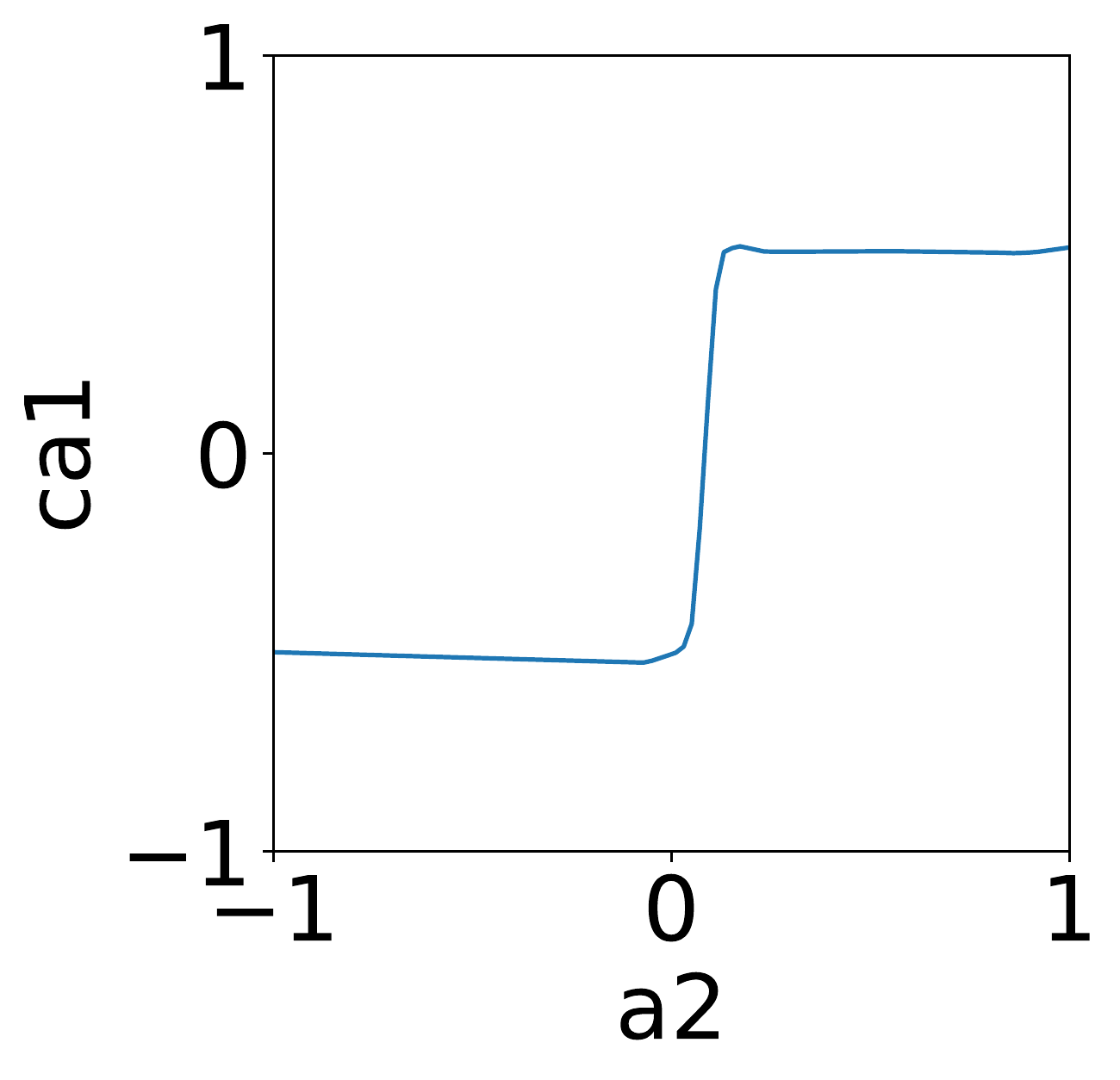}}
    {\includegraphics[width=0.48\columnwidth]{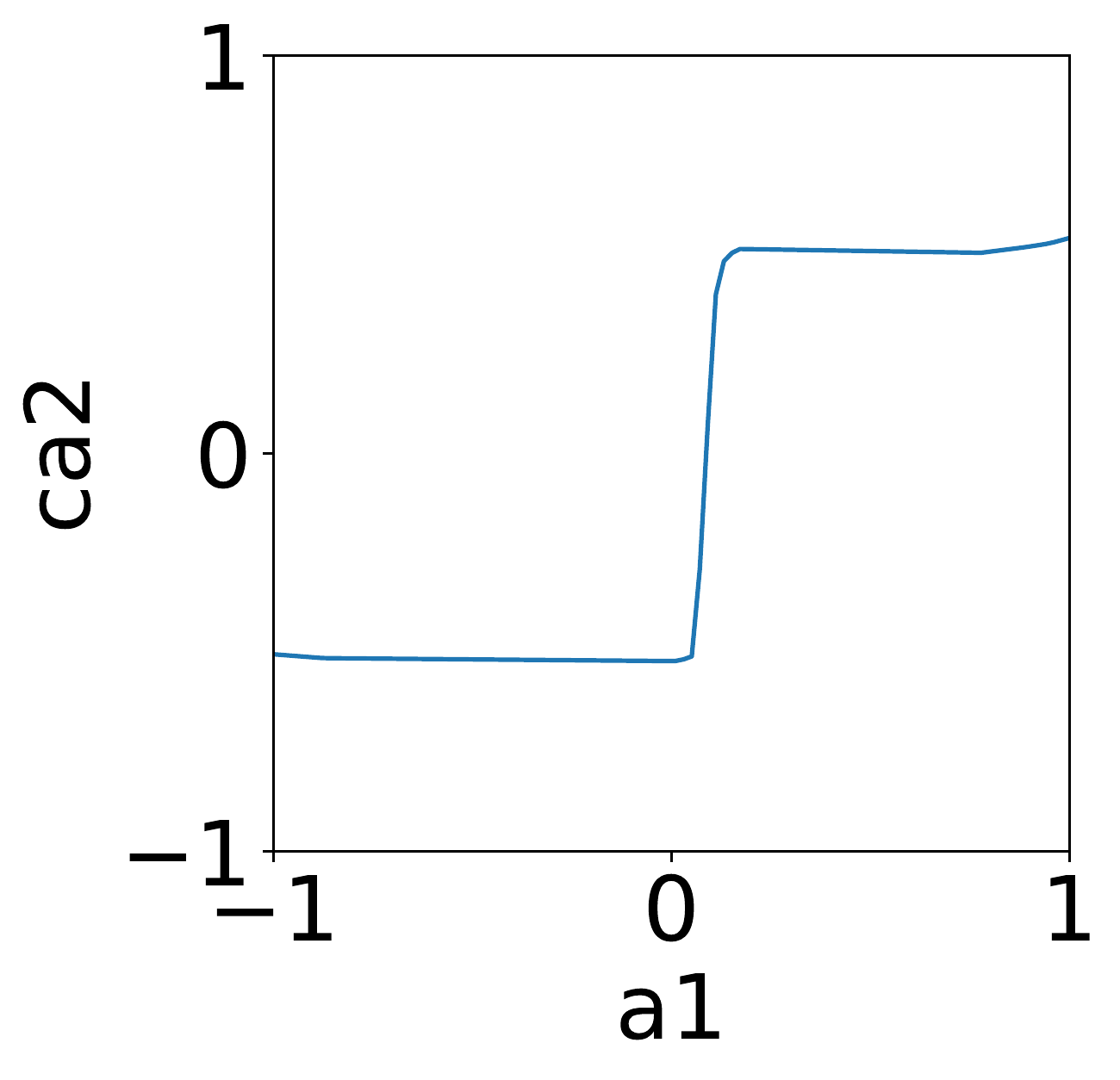}}
    \vspace{-0.25cm}
    \caption{Max of Two}\label{fig:max2_ca}
  \end{subfigure}
  \vspace{-0.2cm}
  \caption{Differential Game: Central actors of R2G. Left: Most likely output of $\pi^1_c$ with respect to $a^2$; Right: Most likely output of $\pi^2_c$ with respect to $a^1$. The central actors successfully learn the optimal response in all scenarios.}
  \vspace{-0.3cm}
  \label{fig:ca}
\end{figure}

For Zero Sum, the best action for agent 1 given $a^2$ is $a^{1,*}=1 \cdot \sign(a^2)$, and the best action for agent 2 given $a^1$ is $a^{2,*}=-1 \cdot \sign(a^1)$. 
Thus, if both agents only consider their opponent's current strategy, they will have an oscillatory learning by alternating between $1$ and $-1$. Such behavior indeed appears in the training of PPO, COMA, and MASAC as shown in \cref{fig:zero_sum_ppo,fig:zero_sum_coma,fig:zero_sum_masac}. 
However, if the agent is aware of the potential response of the opponent to its own action change, their actions should converge to the (0,0) point, as any action deviates from 0 would get a lower reward when the opponent takes response. The central actor in R2G successfully captures this potential response as shown in \cref{fig:zero_sum_ca}. As a result, the R2G agents successfully converge to (0,0) after a short exploration as shown in \cref{fig:zero_sum_r2g}.
The action trajectory of SAC, MADDPG, and PR2 is quite chaotic due to changing opponent's behavior, the non-optimal opponent's action distribution in buffer, and the violation of cooperative assumption.

\begin{figure}[!htbp]
  \centering
  \begin{subfigure}{0.22\columnwidth}
  \centering
    {\includegraphics[width=1.0\columnwidth]{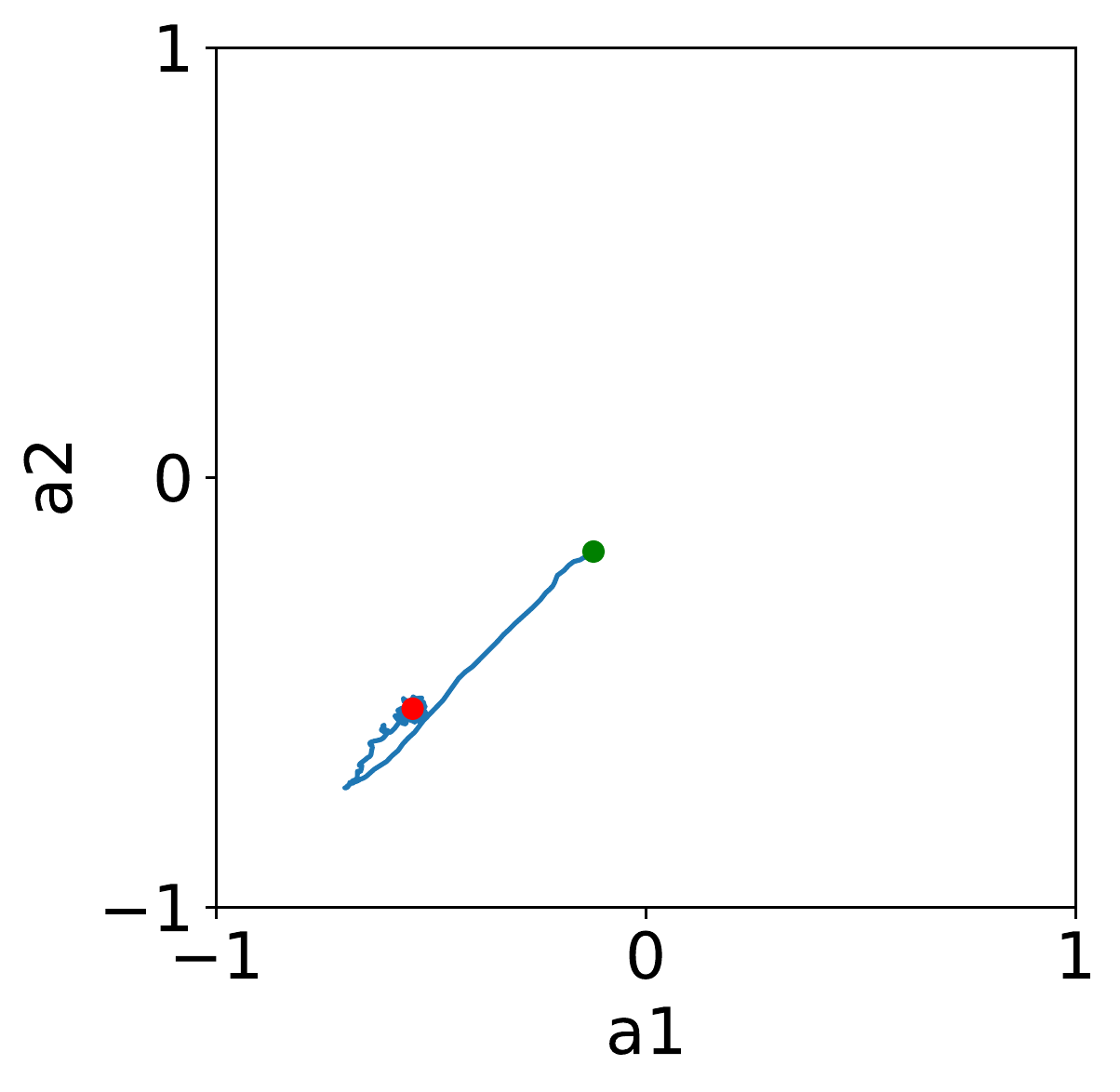}}
    \caption{PPO}\label{fig:max2_ppo}
  \end{subfigure}
  \begin{subfigure}{0.22\columnwidth}
  \centering
    {\includegraphics[width=1.0\columnwidth]{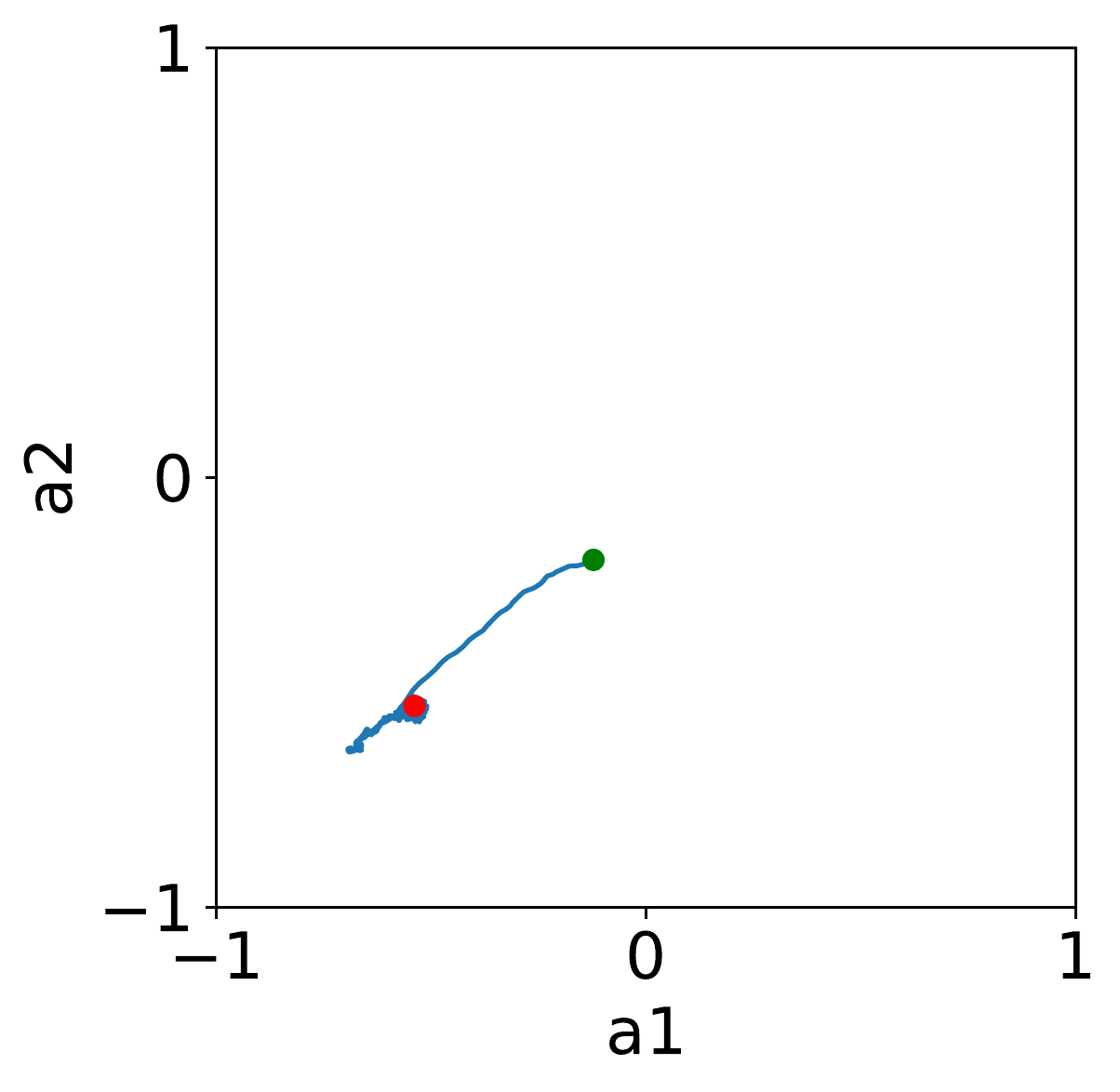}}
    \caption{COMA}\label{fig:max2_coma}
  \end{subfigure}
  \begin{subfigure}{0.22\columnwidth}
  \centering
    {\includegraphics[width=1.0\columnwidth]{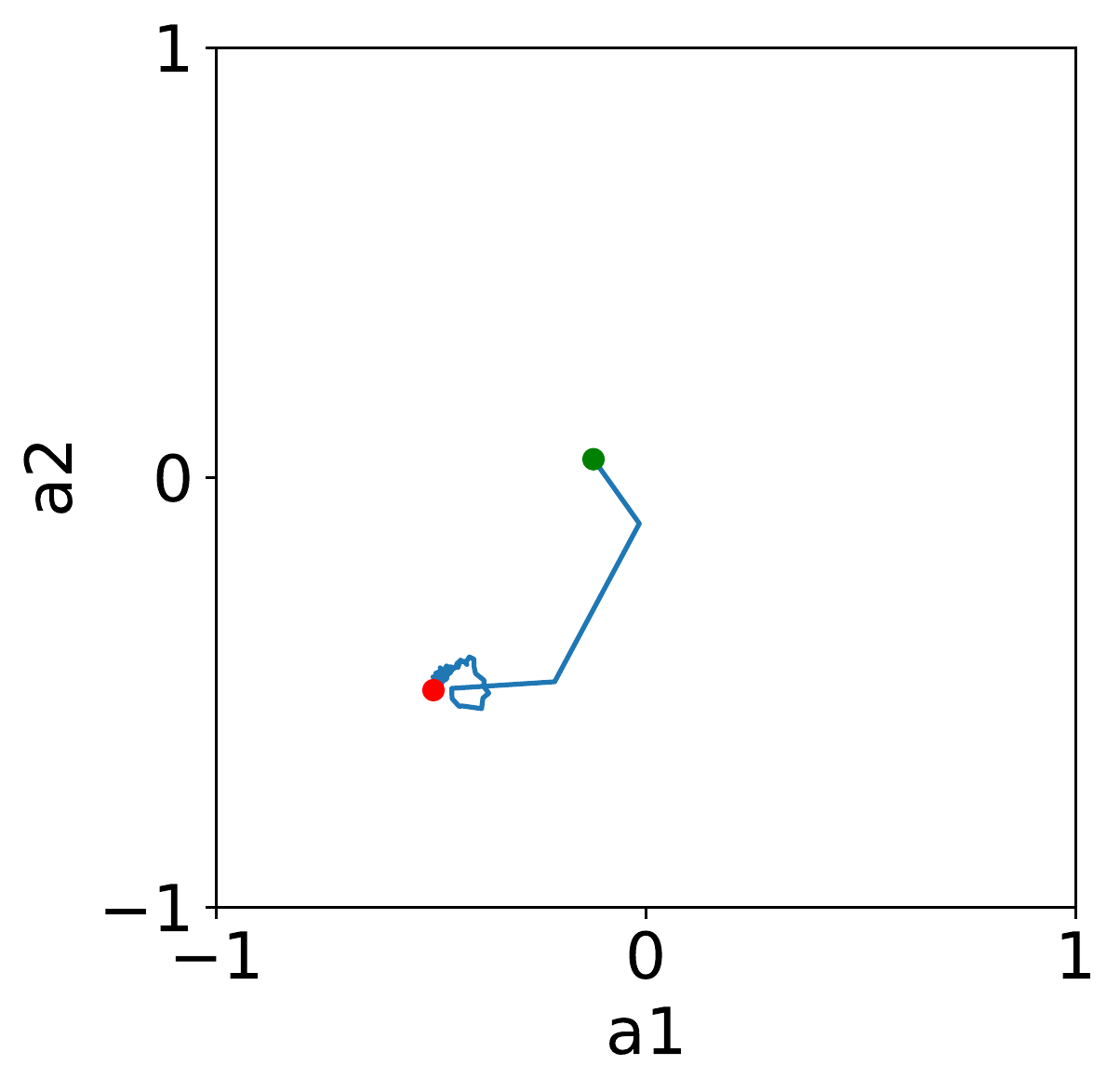}}
    \caption{SAC}\label{fig:max2_sac}
  \end{subfigure}
  \begin{subfigure}{0.22\columnwidth}
  \centering
    {\includegraphics[width=1.0\columnwidth]{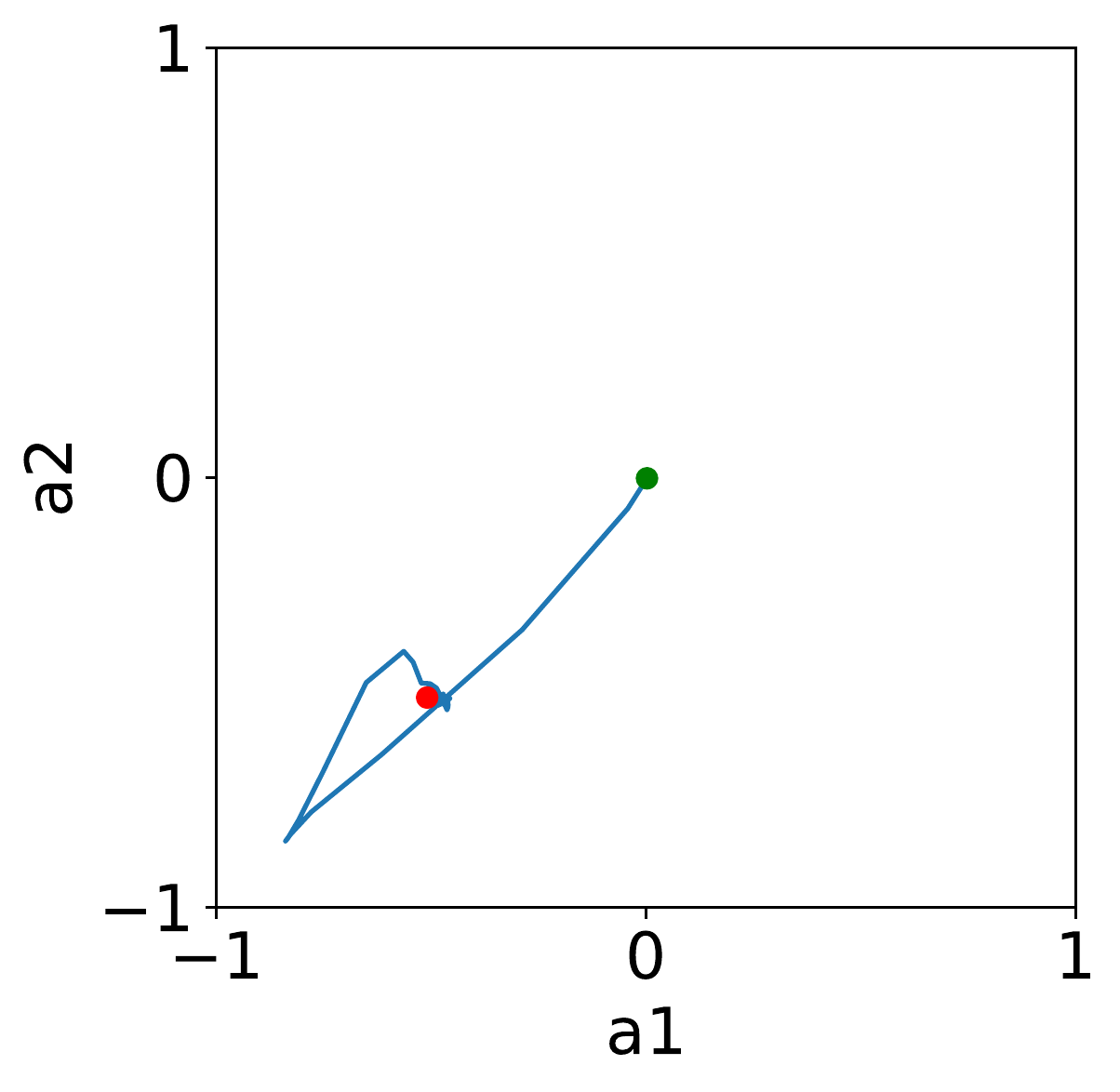}}
    \caption{MADDPG}\label{fig:max2_maddpg}
  \end{subfigure}
  \begin{subfigure}{0.22\columnwidth}
  \centering
    {\includegraphics[width=1.0\columnwidth]{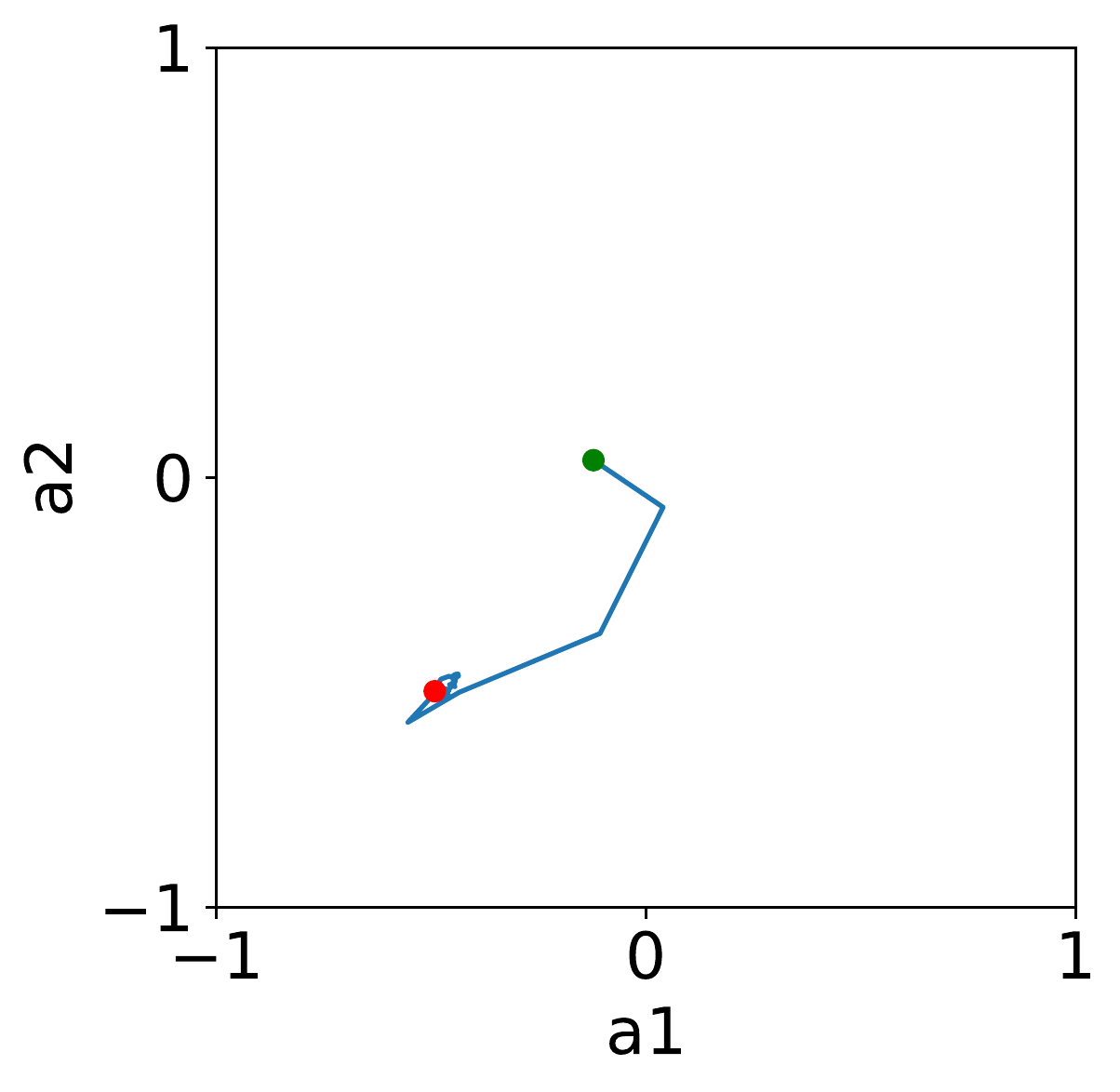}}
    \caption{MASAC}\label{fig:max2_masac}
  \end{subfigure}
  \begin{subfigure}{0.22\columnwidth}
  \centering
    {\includegraphics[width=1.0\columnwidth]{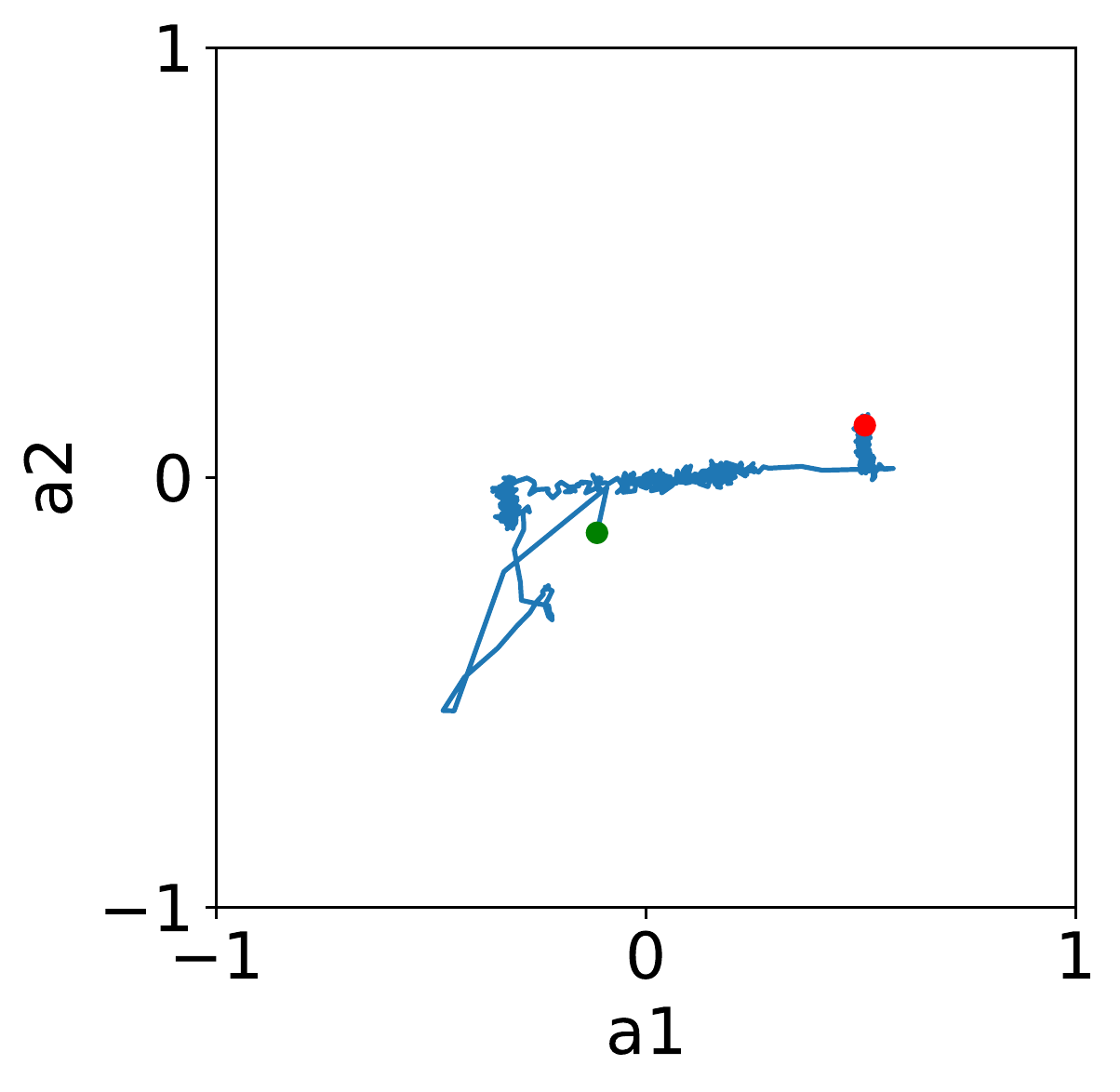}}
    \caption{PR2}\label{fig:max2_pr2}
  \end{subfigure}
  \begin{subfigure}{0.22\columnwidth}
  \centering
    {\includegraphics[width=1.0\columnwidth]{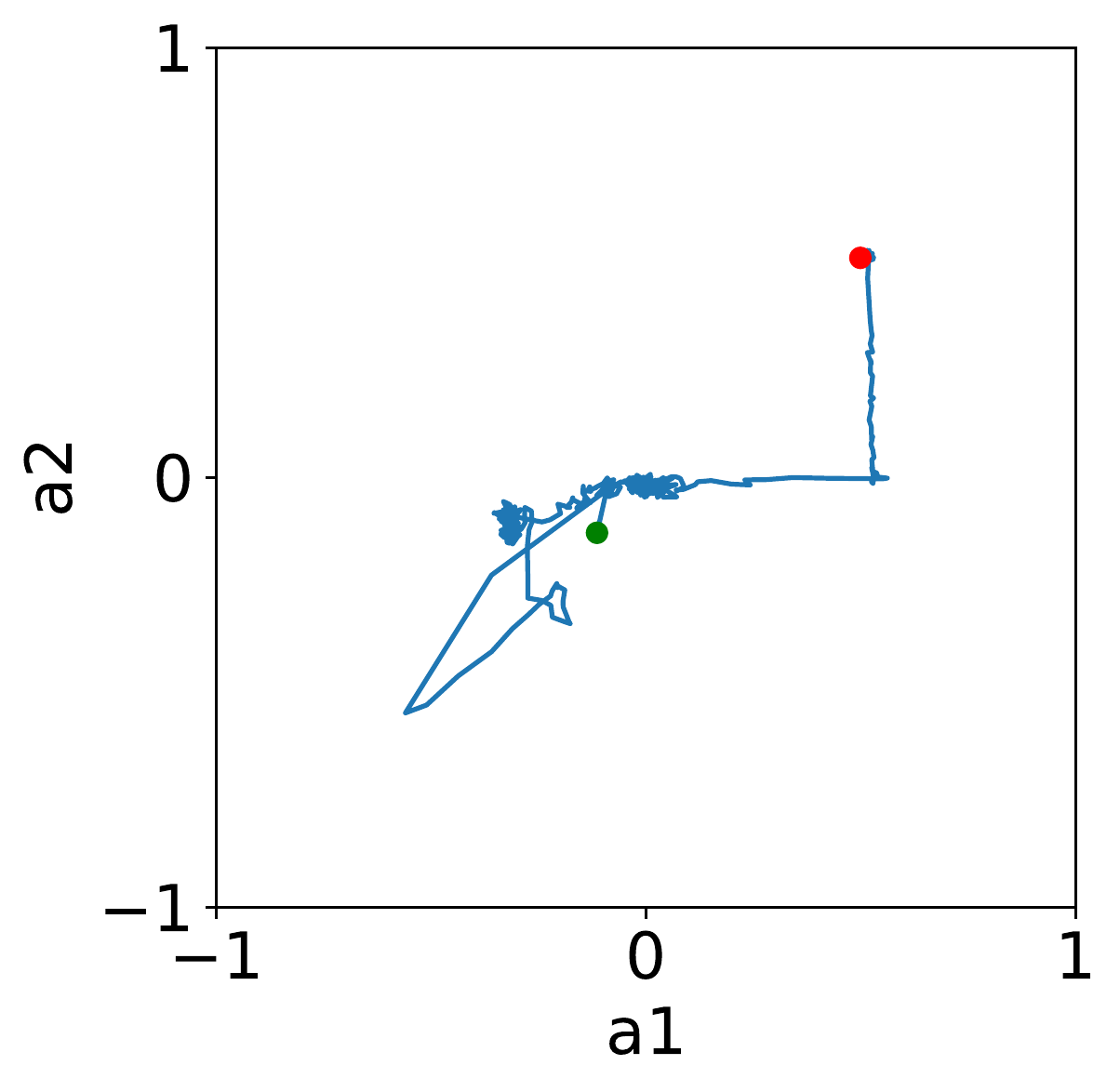}}
    \caption{R2G}\label{fig:max2_r2g}
  \end{subfigure}
  \vspace{-0.2cm}
  \caption{Max of Two: Trajectories of most likely actions for different algorithms over 1000 iterations.}
  \vspace{-0.3cm}
  \label{fig:max2}
\end{figure}

For Max of Two, as a cooperative game, there is a local optimum at $(-0.5,-0.5)$ and a global optimum at $(0.5,0.5)$. 
Due to the shape of the reward landscape, as the initial action distributions of agents are in the vicinity of $(0,0)$, even with sufficient exploration, each agent's reward estimation at \num{-0.5} would be higher than that at \num{0.5} given the distribution of its opponent's action. Thus, the learning agents are easily attracted into the local optimum. 
As shown in \cref{fig:max2_ppo,fig:max2_coma,fig:max2_sac,fig:max2_maddpg,fig:max2_masac}, all methods except for PR2 and R2G suffers from such relative over-generalization problem. 
However, if the agent could predict the cooperative response of the opponent, e.g., when the agent acts at the global optimum, the opponent would also act at the global optimum, its value estimation of the global optimum action would be higher than the local optimum action. 
Thus, both agents would converge to the global optimum. The central actor in R2G successfully captures this potential response as shown in \cref{fig:max2_ca}, and the R2G agents successfully converge to $(0.5,0.5)$ as shown in \cref{fig:max2_r2g}. Similar behavior could also be observed in the training of PR2 as shown in \cref{fig:max2_pr2}.

\begin{figure}[!ht]
  \centering
  \begin{subfigure}{0.24\columnwidth}
  \centering
    {\includegraphics[width=0.75\columnwidth]{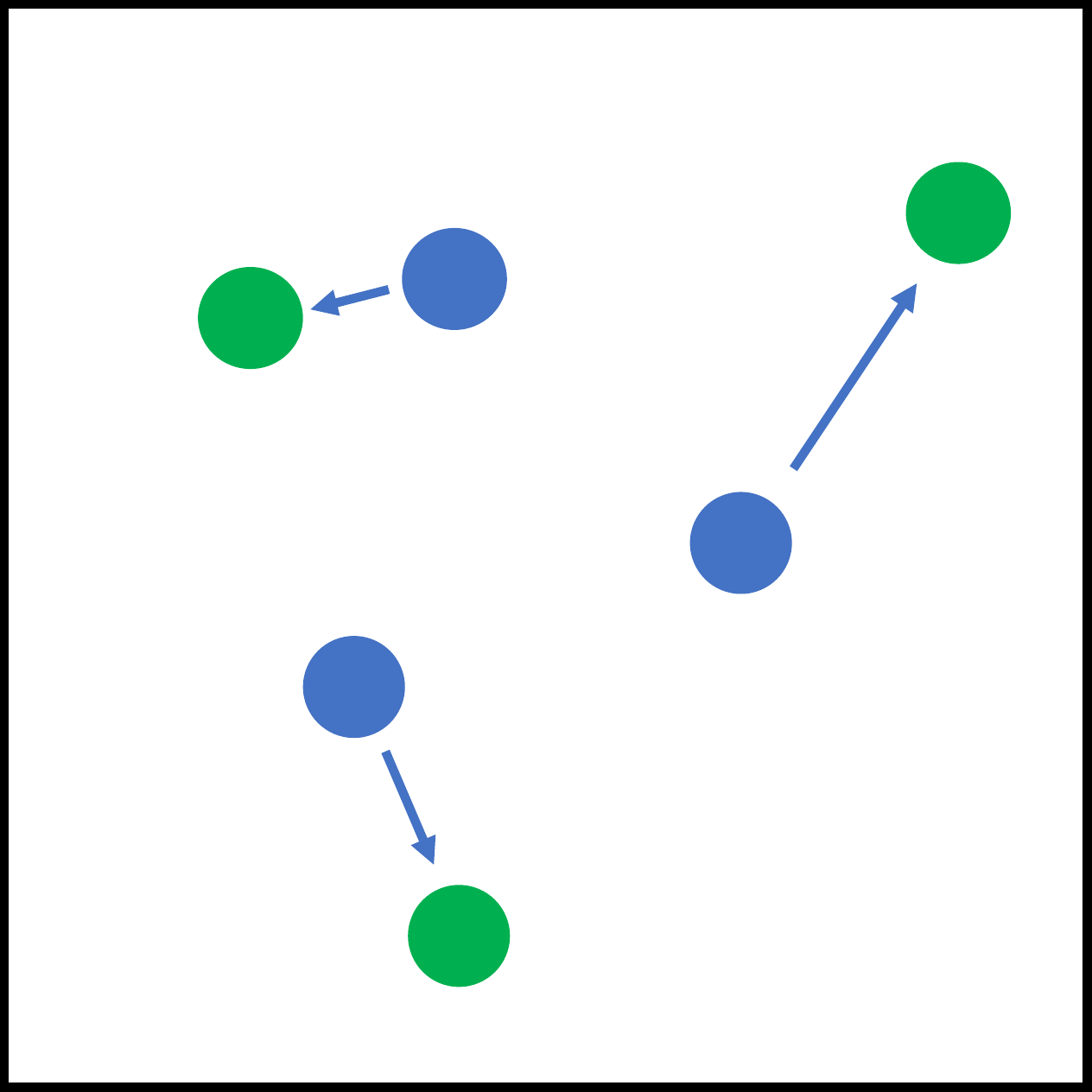}}
    \caption{Cooperative \\\hspace*{0.4cm} Navigation}\label{fig:spread}
  \end{subfigure}
  \begin{subfigure}{0.24\columnwidth}
  \centering
    {\includegraphics[width=0.75\columnwidth]{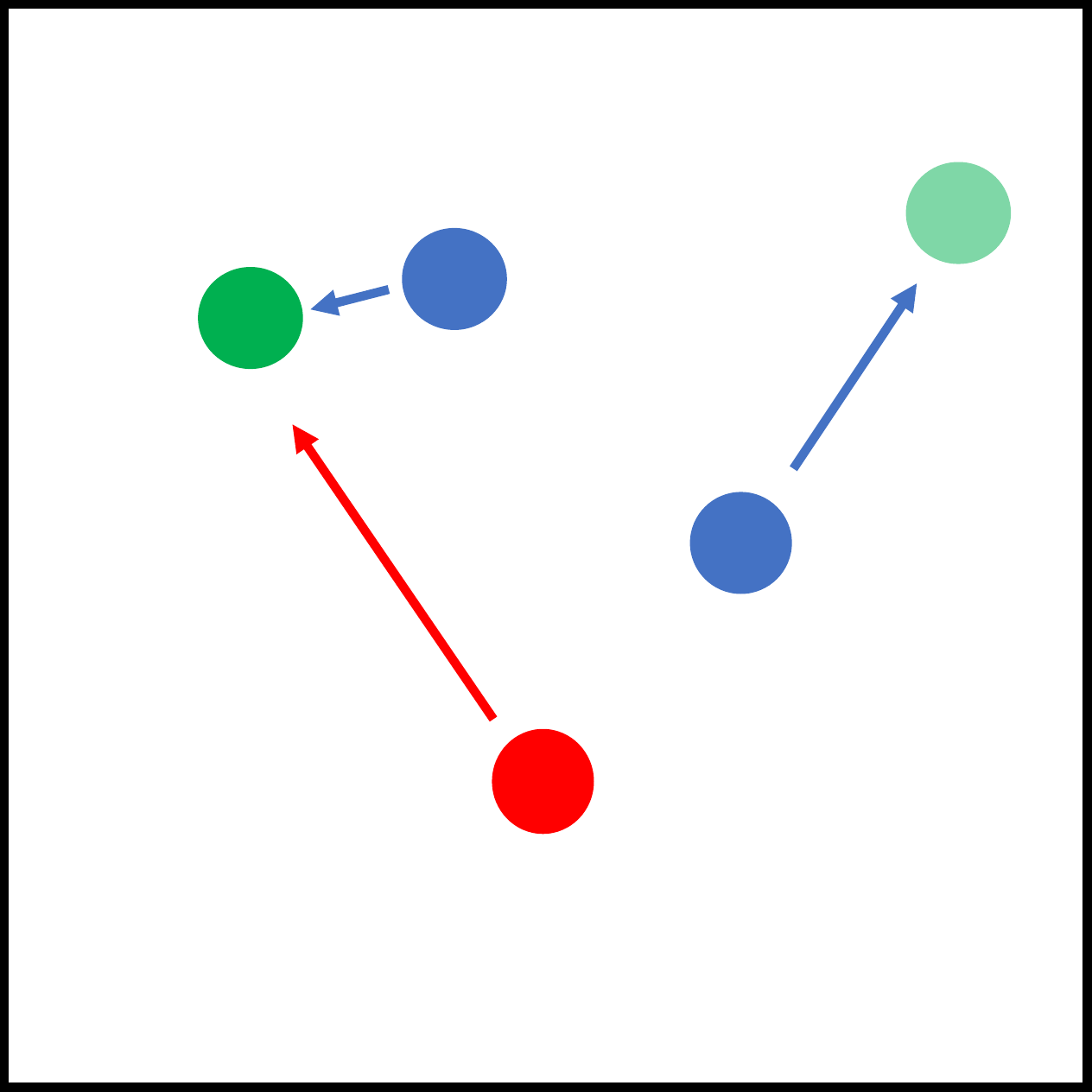}}
    \caption{Physical \\\hspace*{0.4cm} Deception}\label{fig:adversary}
  \end{subfigure}
  \begin{subfigure}{0.24\columnwidth}
  \centering
    {\includegraphics[width=0.75\columnwidth]{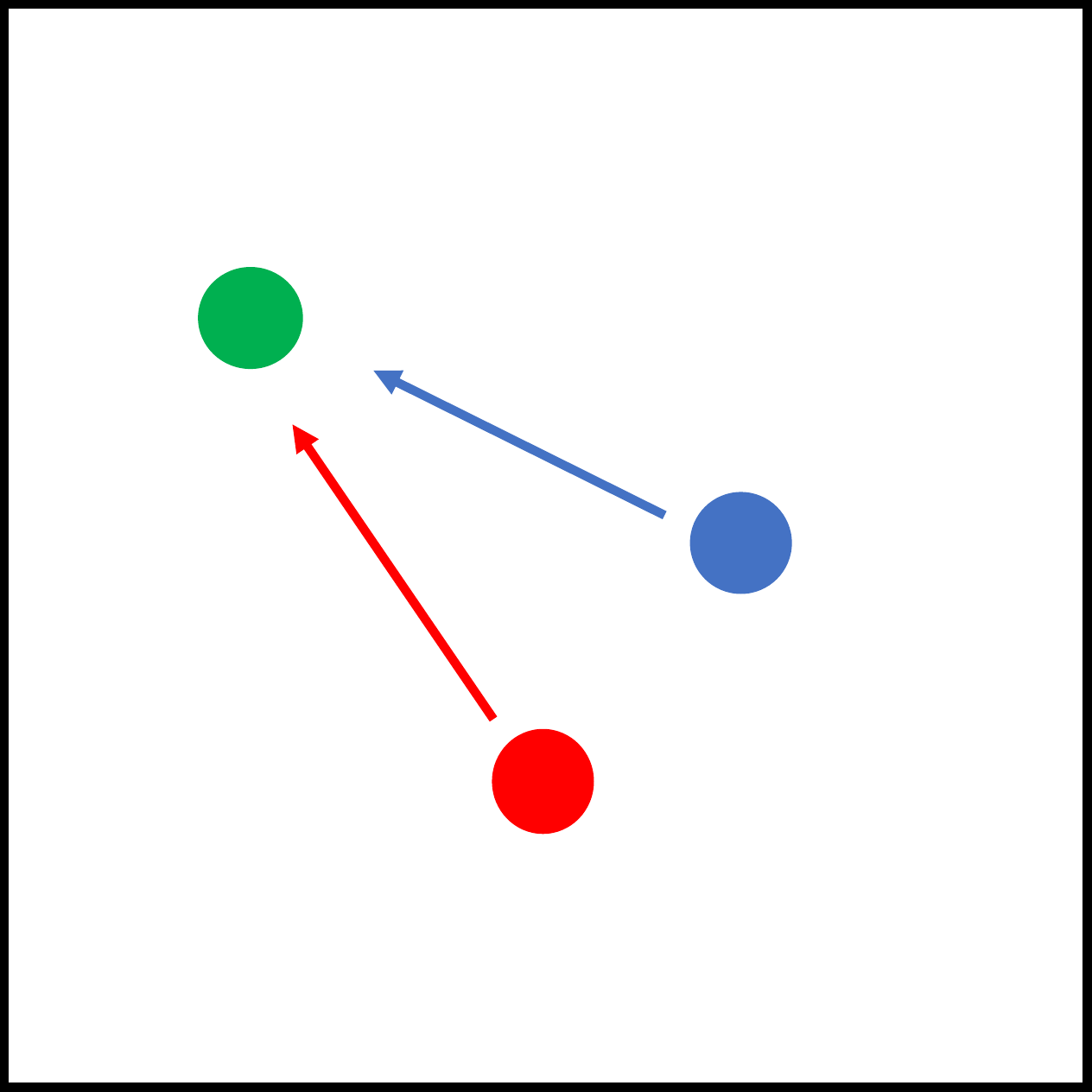}}
    \caption{Keep  \\\hspace*{0.4cm} Away}\label{fig:push}
  \end{subfigure}
  \begin{subfigure}{0.24\columnwidth}
  \centering
    {\includegraphics[width=0.75\columnwidth]{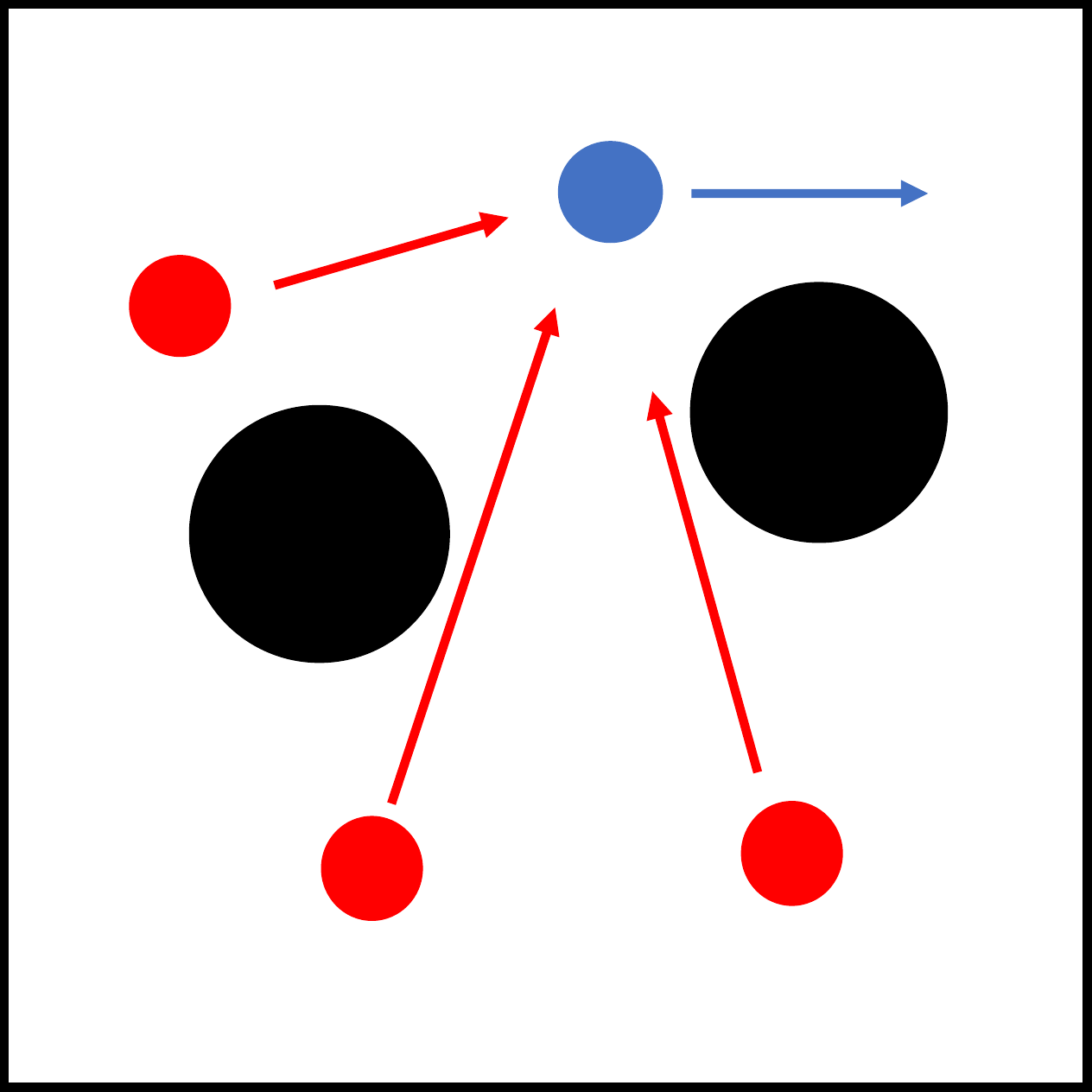}}
    \caption{Predator  \\\hspace*{0.4cm} Prey}\label{fig:tag}
  \end{subfigure}
  \caption{\footnotesize{Particle World: Illustration of scenarios. The blue and red dots are good agents and adversaries. The green dots indicate the landmarks. In Physical Deception, a shadow green dot indicates the fake target landmark. The black dots are static obstacles. Arrows indicate the desired motion of agents.}}
  \vspace{-0.2cm}
  \label{fig:particle}
\end{figure}

\begin{figure}[!ht]
  \centering
  \includegraphics[width=0.9\columnwidth]{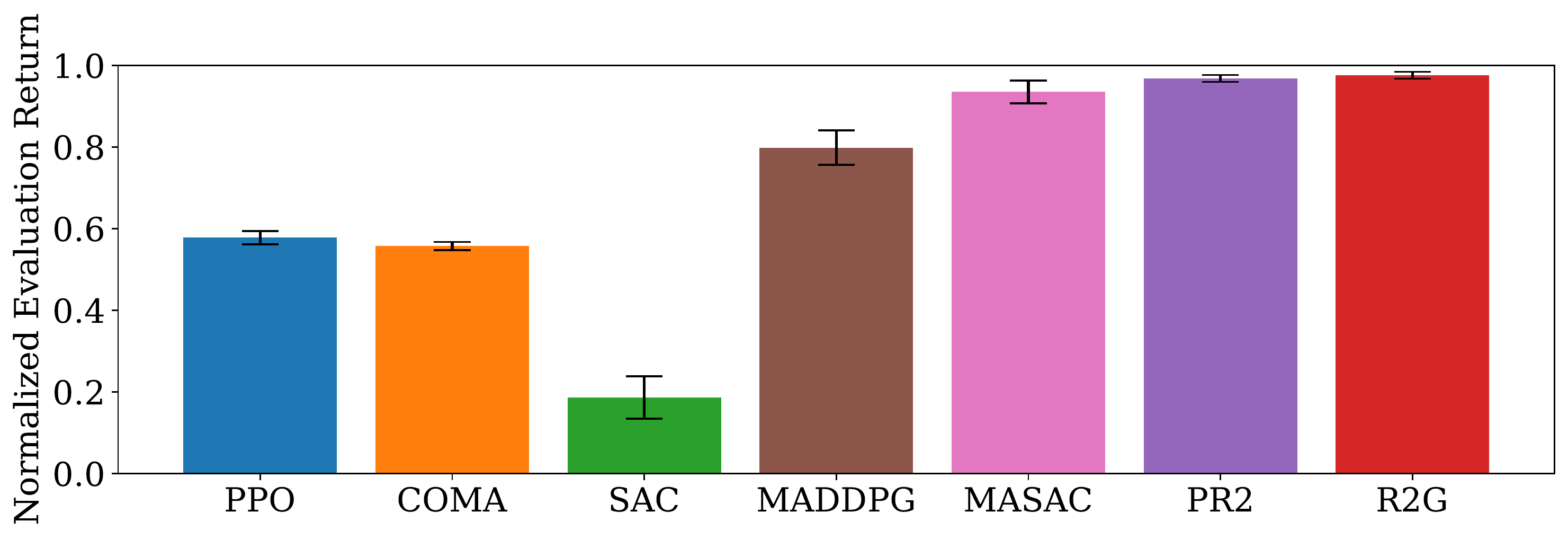}
  \caption{Cooperative Navigation: Normalized evaluation return of different algorithms.}
  \vspace{-0.5cm}
  \label{fig:spread_return}
\end{figure}

\begin{figure*}[!ht]
  \centering
  \begin{subfigure}{2.1\columnwidth}
  \centering
    {\includegraphics[width=0.34\columnwidth]{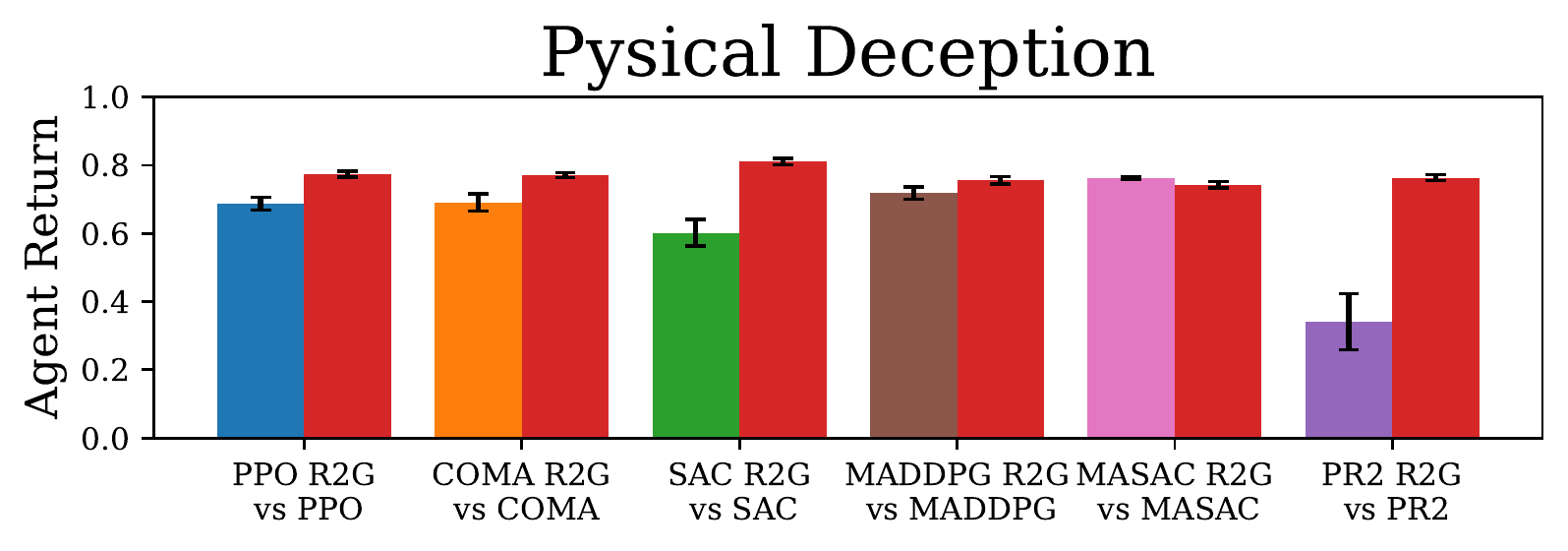}}
    {\includegraphics[width=0.32\columnwidth]{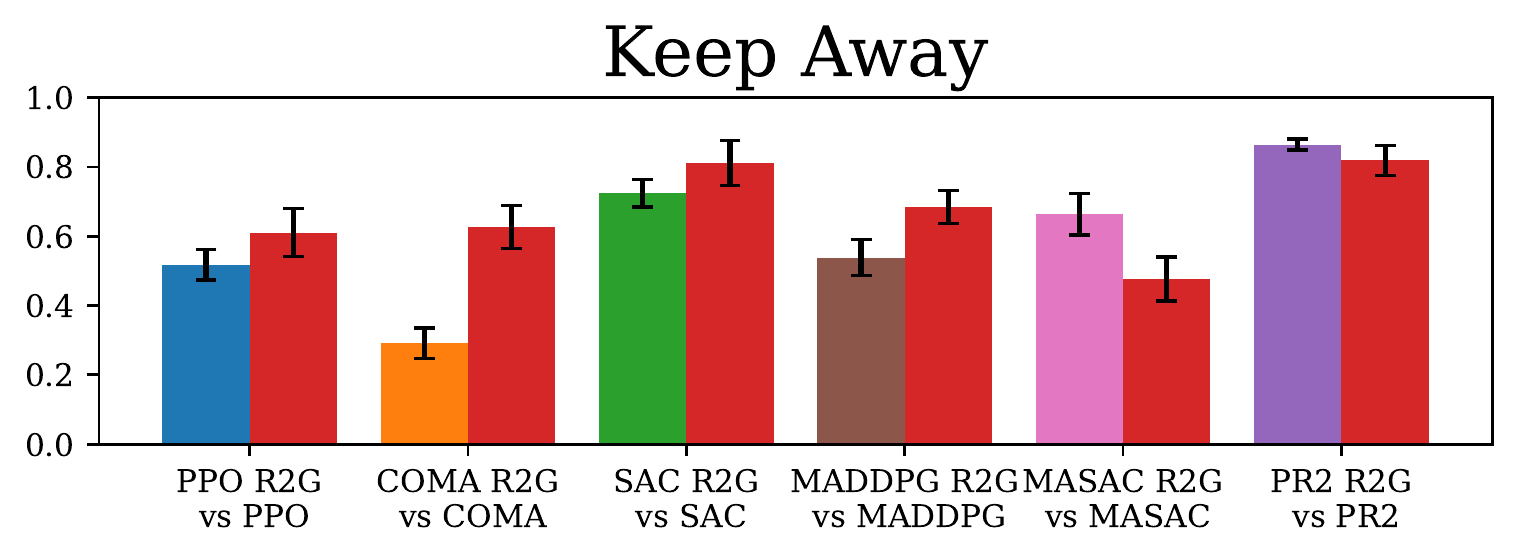}}
    {\includegraphics[width=0.32\columnwidth]{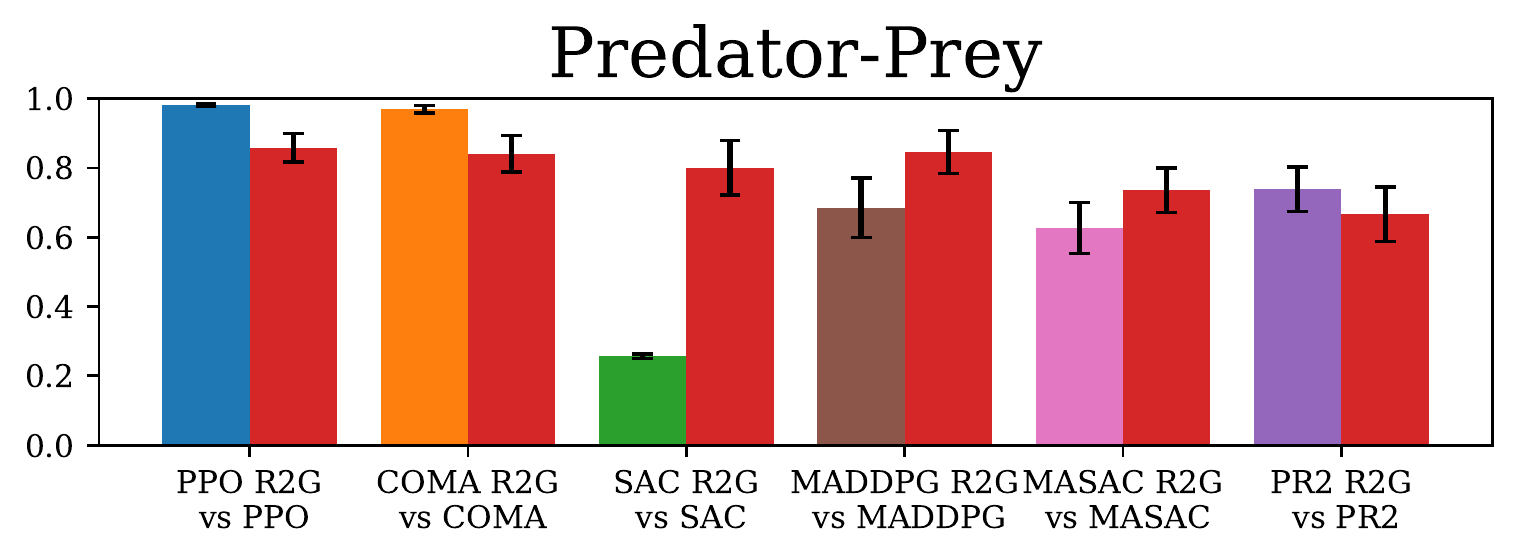}}
    \caption{Average agent return playing against the same baseline adversaries}
    \label{fig:particle_ag_baseline}
  \end{subfigure}
  \begin{subfigure}{2.1\columnwidth}
  \centering
    {\includegraphics[width=0.34\columnwidth]{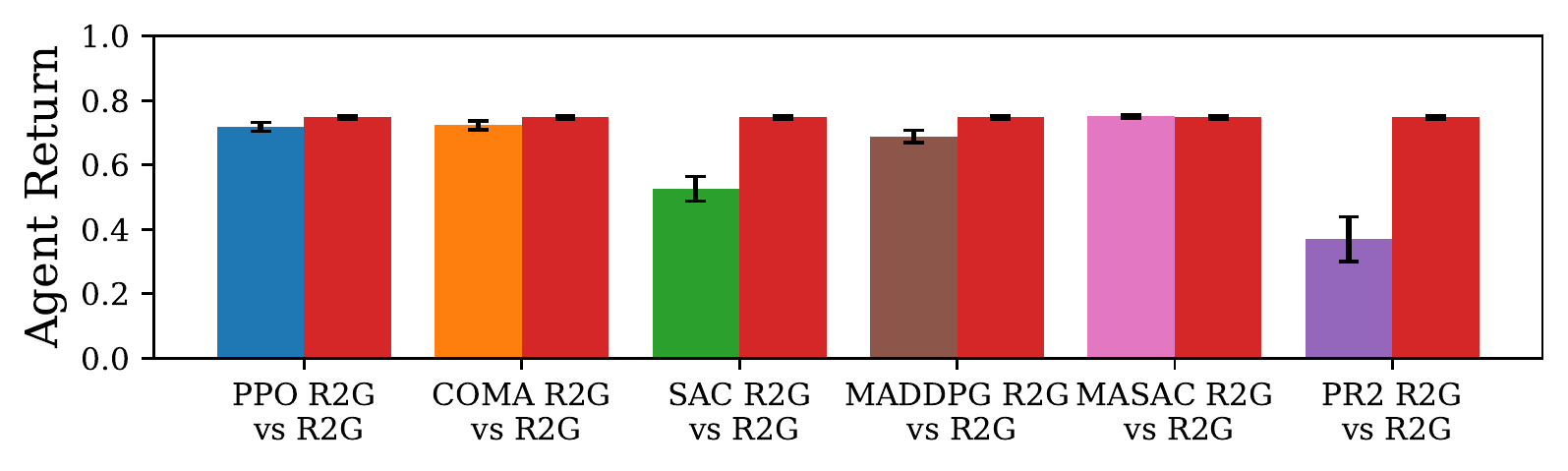}}
    {\includegraphics[width=0.32\columnwidth]{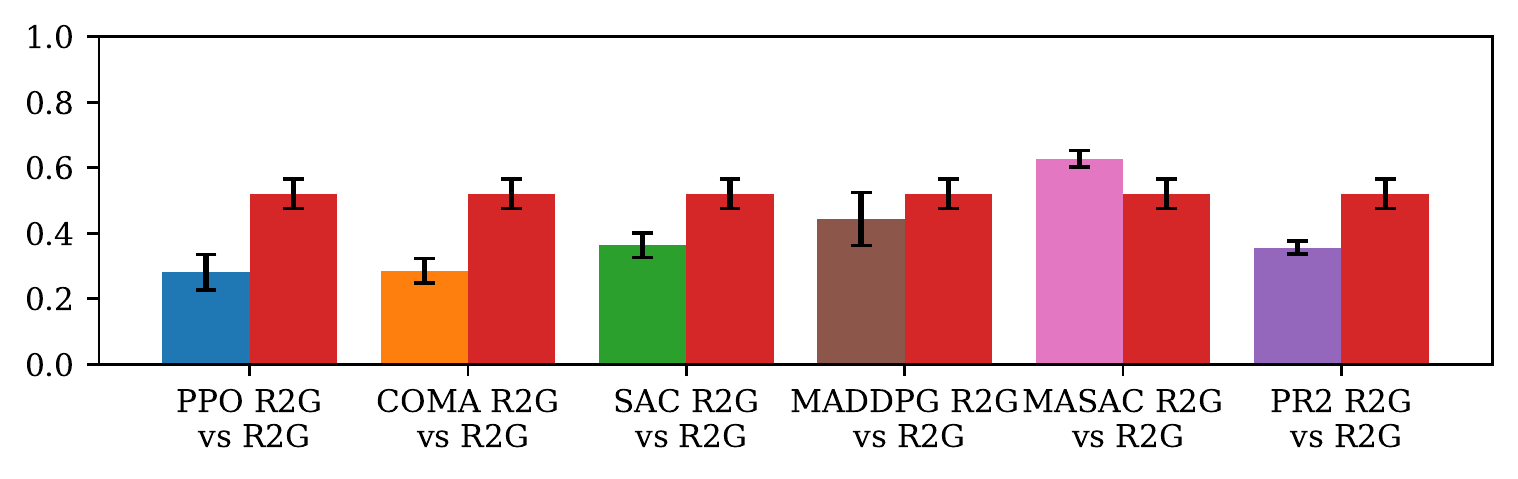}}
    {\includegraphics[width=0.32\columnwidth]{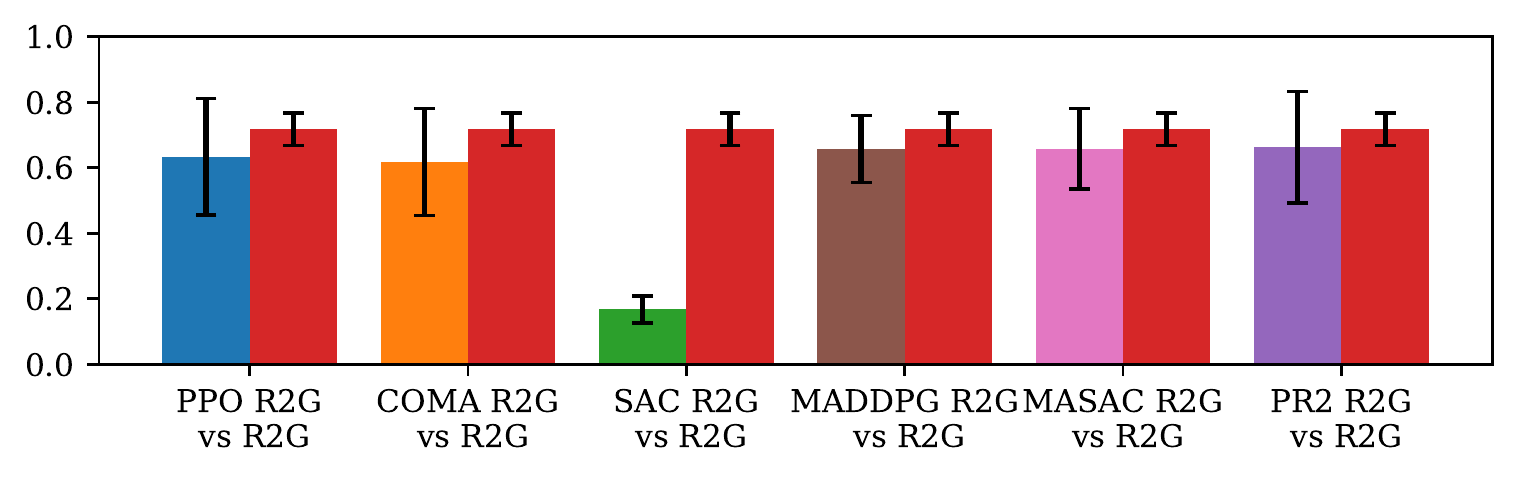}}
    \caption{Average agent return playing against the same R2G adversaries}
    \label{fig:particle_ag_r2g}
  \end{subfigure}
  \begin{subfigure}{2.1\columnwidth}
  \centering
    {\includegraphics[width=0.34\columnwidth]{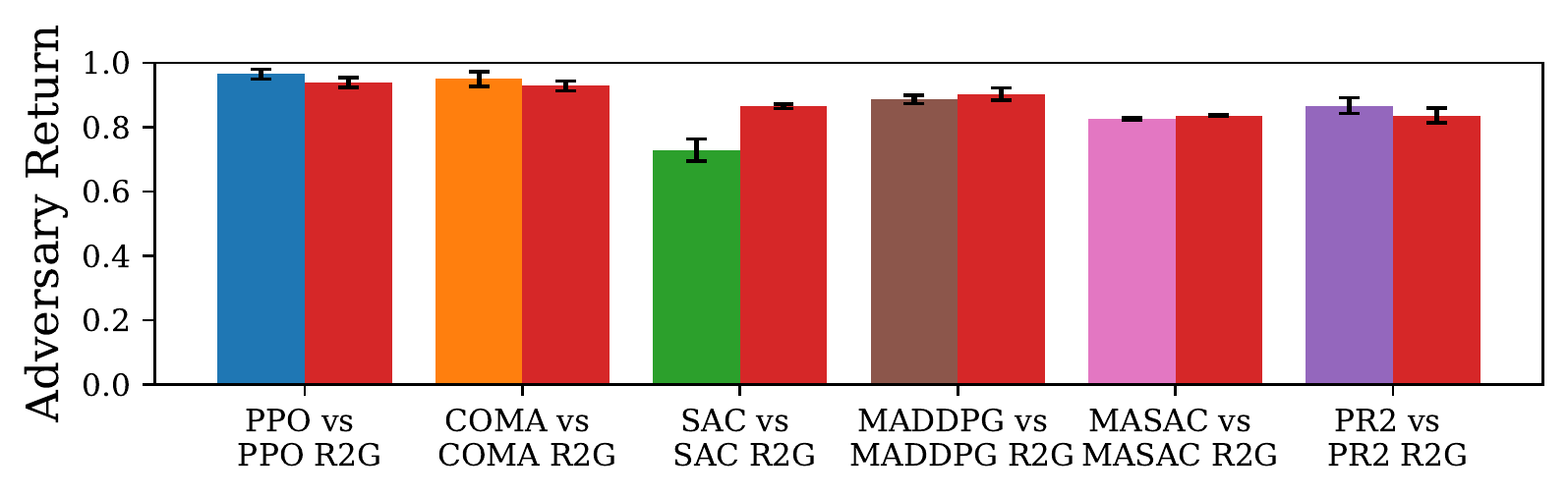}}
    {\includegraphics[width=0.32\columnwidth]{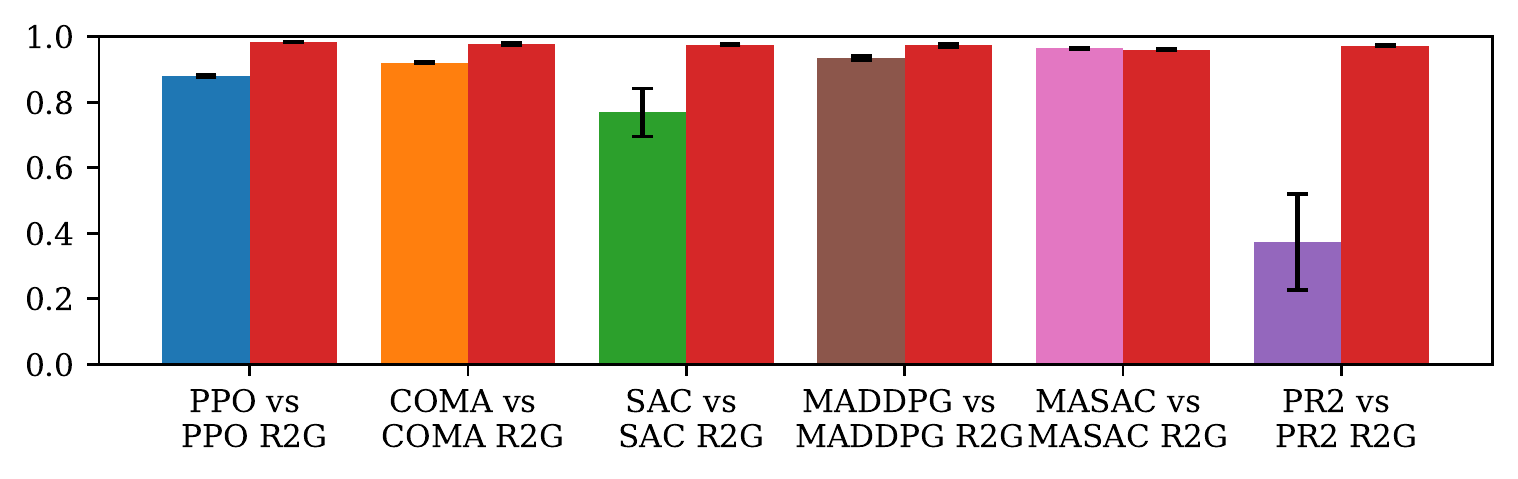}}
    {\includegraphics[width=0.32\columnwidth]{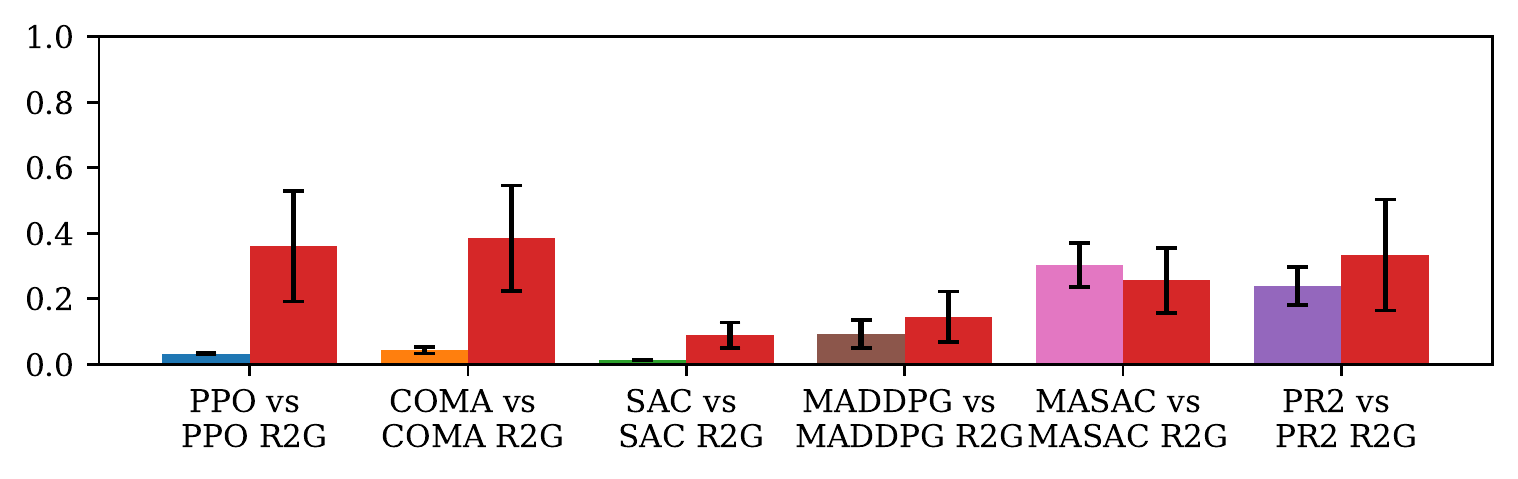}}
    \caption{Average adversary return playing against the same baseline agents}
    \label{fig:particle_adv_baseline}
  \end{subfigure}
  \begin{subfigure}{2.1\columnwidth}
  \centering
    {\includegraphics[width=0.34\columnwidth]{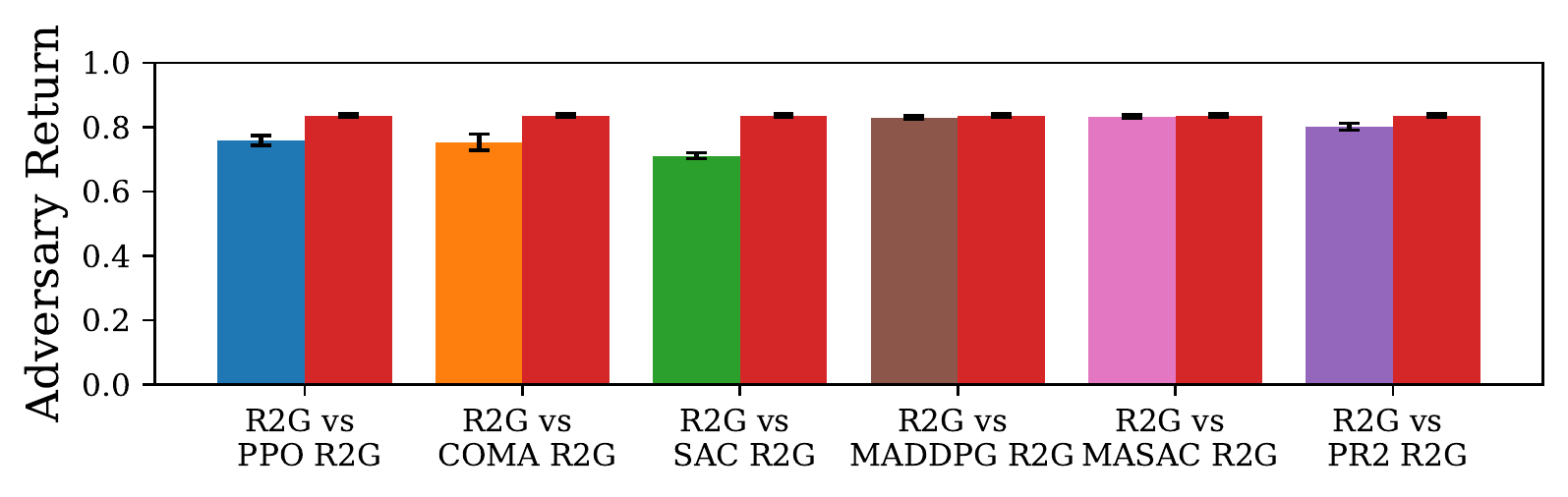}}
    {\includegraphics[width=0.32\columnwidth]{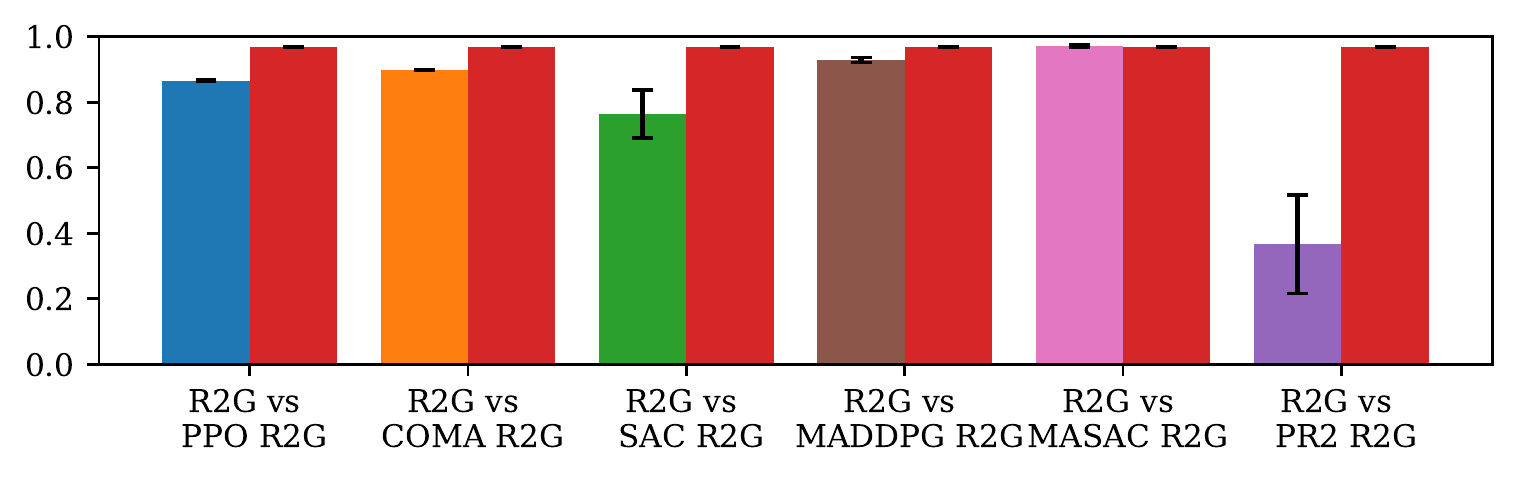}}
    {\includegraphics[width=0.32\columnwidth]{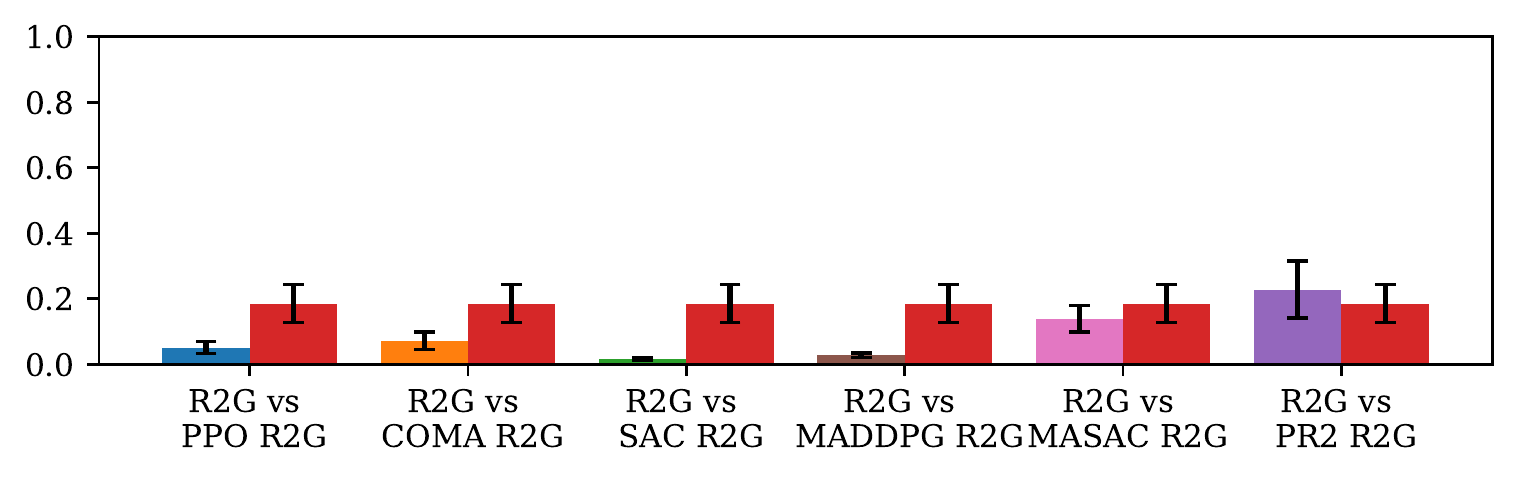}}
    \caption{Average adversary return playing against the same R2G agents}
    \label{fig:particle_adv_r2g}
  \end{subfigure}
  \caption{Particle World: Pairwise comparison of agent and adversary returns playing against the same opponents. The left bar indicates the baseline return, and the red bar indicates the R2G return.}
  \label{fig:particle_return}
\end{figure*}

\subsection{Particle World}\label{sec:ca}

We further compare R2G against other baselines on several multi-player multi-state Particle World environments~\citep{maddpg} including:

\begin{itemize}
    \item \textbf{Cooperative Navigation}: A cooperative game with 3 agents and 3 landmarks. Agents are rewarded to spread and cover all landmarks and penalized by collisions. 
    \item \textbf{Physical Deception}: A competitive game with 1 adversary, 2 good agents, and 2 landmarks. All agents could observe the positions of other agents and landmarks. There is only one true target landmark which is only known to good agents. The good agents are rewarded on how close any one of them is to the target landmark and penalized if the adversary is close to the target landmark. The adversary is rewarded on how close it is to the target landmark.
    \item \textbf{Keep Away}: A competitive game with 1 adversary, 1 good agent, and 1 landmark. The good agent is rewarded on how close it is to the landmark. The adversary is rewarded if it is close to the landmark, and if the agent is pushed away from the landmark. 
    \item \textbf{Predator-Prey}: A competitive game with 3 adversaries (predators), 1 good agent (prey), and several obstacles. The good agent is faster and is rewarded to run away from the three adversaries. The adversaries are slower and are rewarded to catch the good agent.
\end{itemize}

We first compare the performance of different approaches on the Cooperative Navigation environment by evaluating the test time returns of agents trained together. The returns are averaged over 5 random seeds each with 1000 trajectories, and are reported using Min-Max normalization~\citep{patro2015normalization}. 
As shown in \cref{fig:spread_return}, R2G and PR2 achieve the best performance among all tested approaches.
In this game, a positive return needs each agent cover one of the landmarks. Thus, an agent is only motivated to approach one landmark when it expects the other agents also approaching other landmarks. The recursive reasoning in PR2 and R2G successfully capture this.
The off-policy actor-critic methods (MADDPG, MASAC, PR2, R2G) are generally better than on-policy methods (PPO and COMA), this is likely due to the better data efficiency of actor-critic methods. The independent SAC performs much worse than the others, since the experience stored in the buffer is highly biased due to the learning of other agents.

To compare the performance of R2G on competitive games, we plot the pair-wise returns of different approaches playing against the same opponents. The results are shown in \cref{fig:particle_return}. In each pair, we first evaluate the returns of different good agents playing against the same adversary trained from the baseline (\cref{fig:particle_ag_baseline}) or from R2G (\cref{fig:particle_ag_r2g}). Then we evaluate the returns of different adversaries playing against the same good agent trained from the baseline (\cref{fig:particle_adv_baseline}) or from R2G (\cref{fig:particle_adv_r2g}).

\begin{figure*}[!ht]
  \begin{subfigure}{2.1\columnwidth}
  \centering
    {\includegraphics[width=.34\columnwidth]{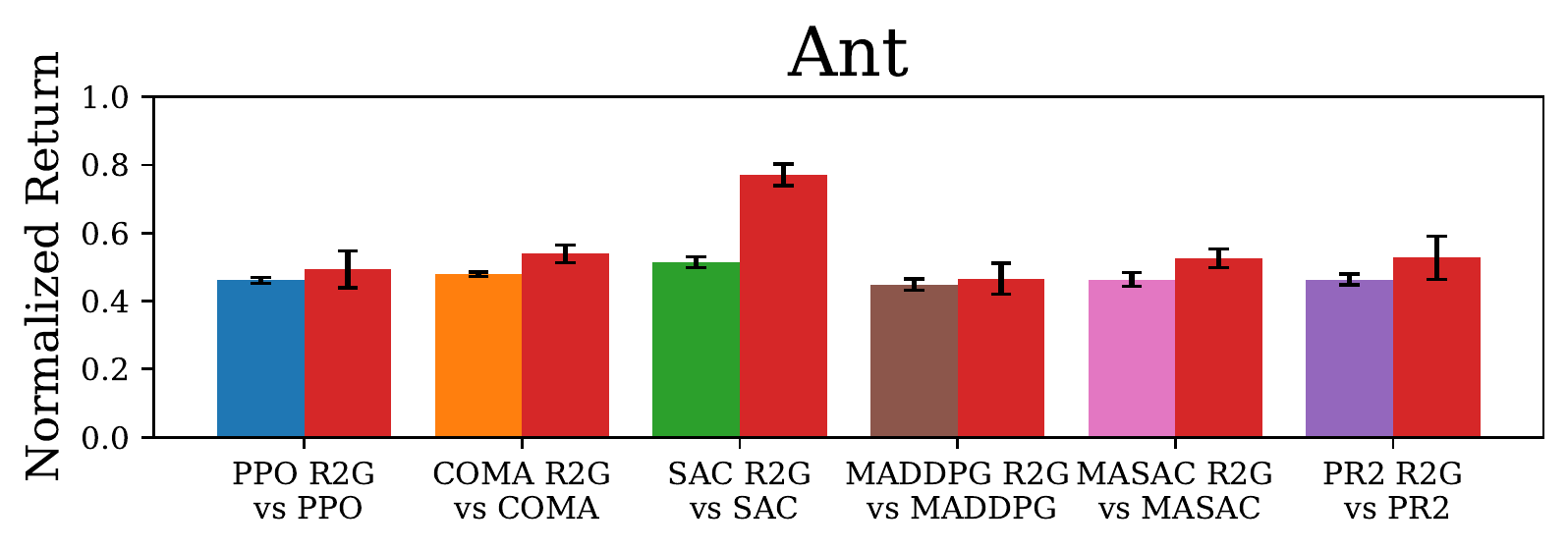}}
    {\includegraphics[width=.32\columnwidth]{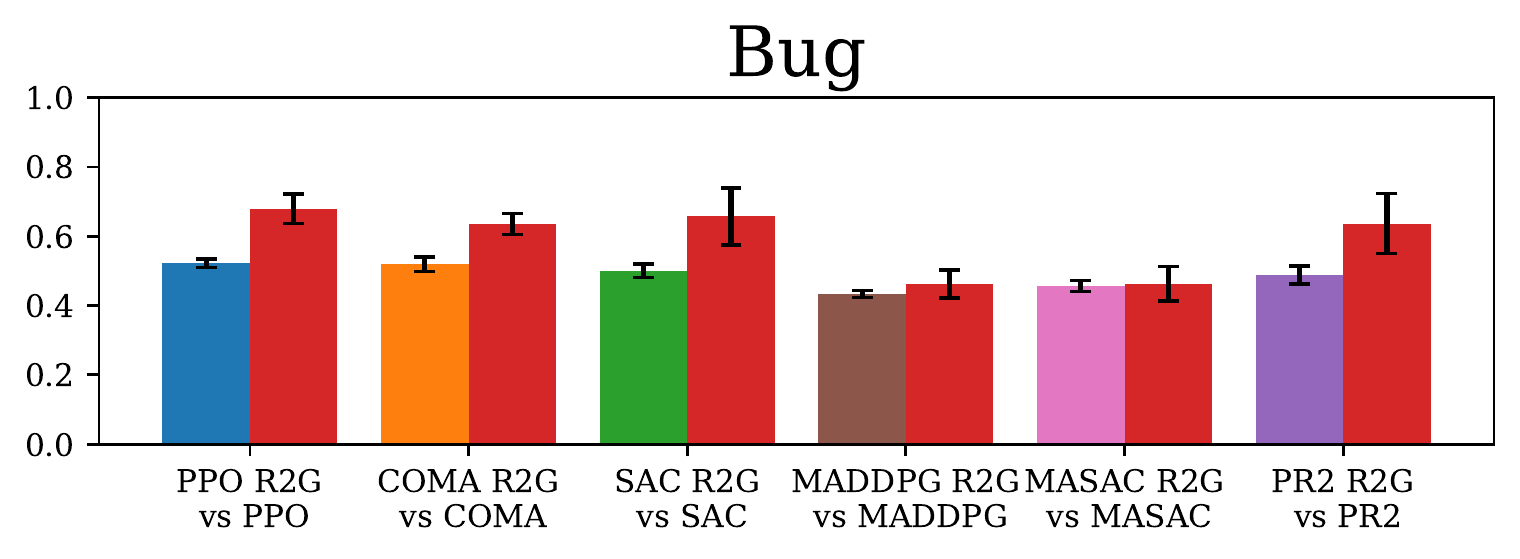}}
    {\includegraphics[width=.32\columnwidth]{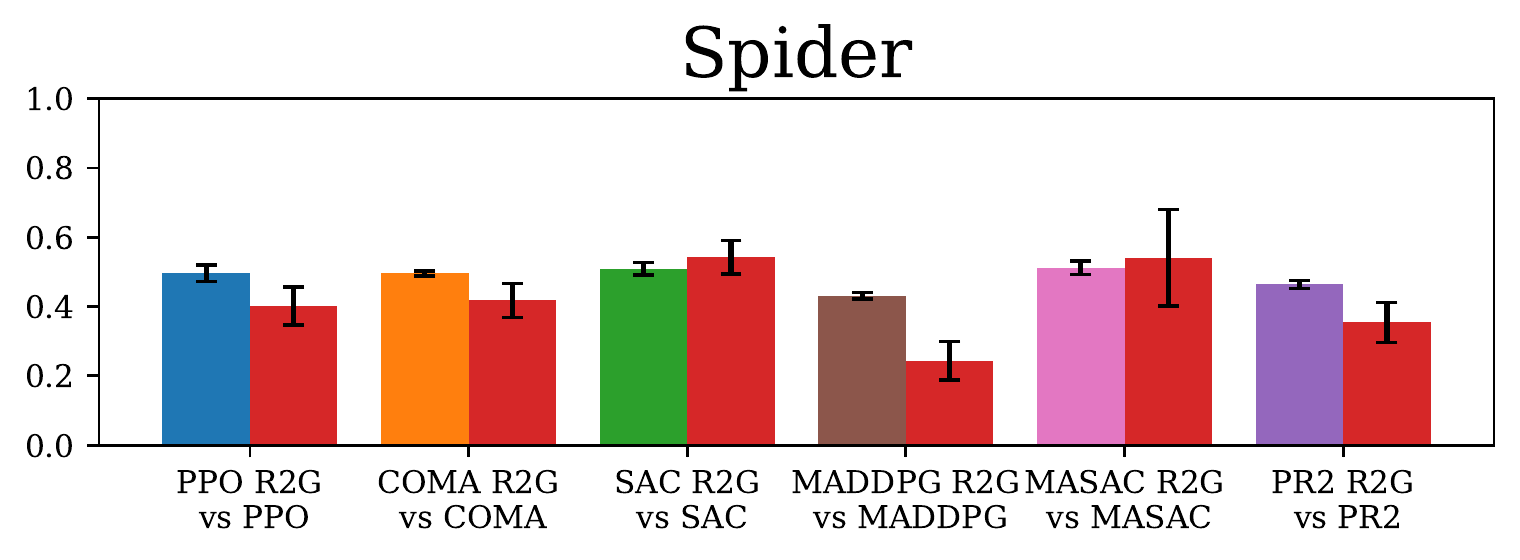}}
    \caption{Average agent return playing against the same baseline opponent}
    \label{fig:robosumo_baseline}
  \end{subfigure}

  \begin{subfigure}{2.1\columnwidth}
  \centering
    {\includegraphics[width=.34\columnwidth]{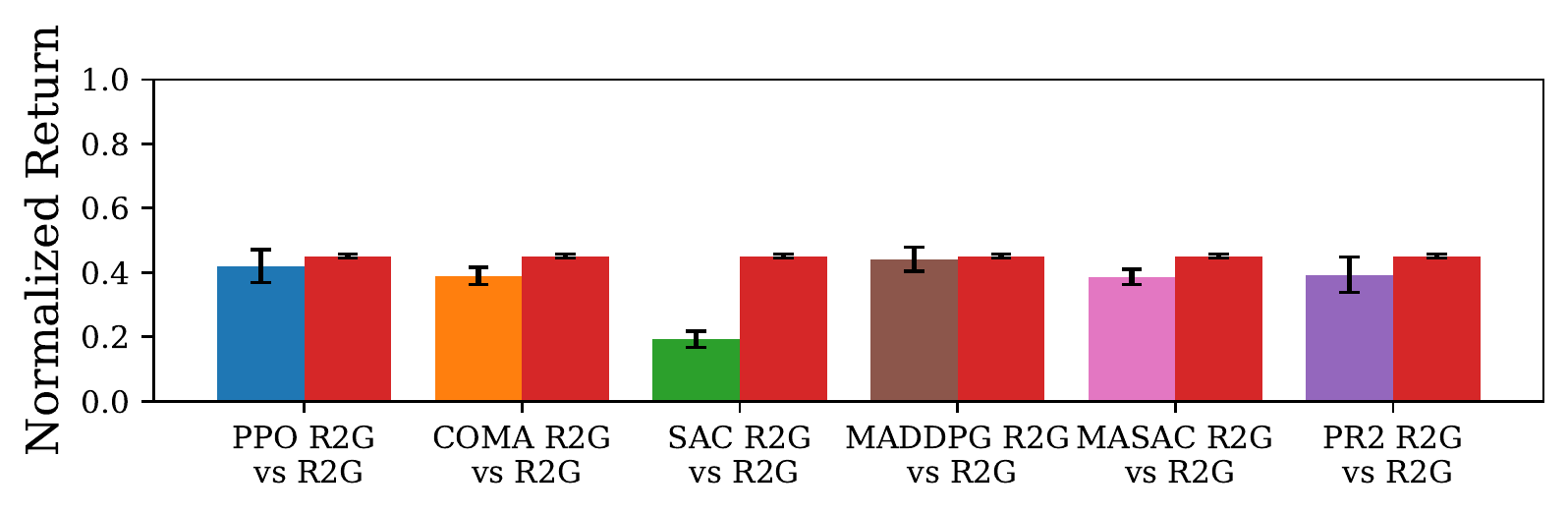}}
    {\includegraphics[width=.32\columnwidth]{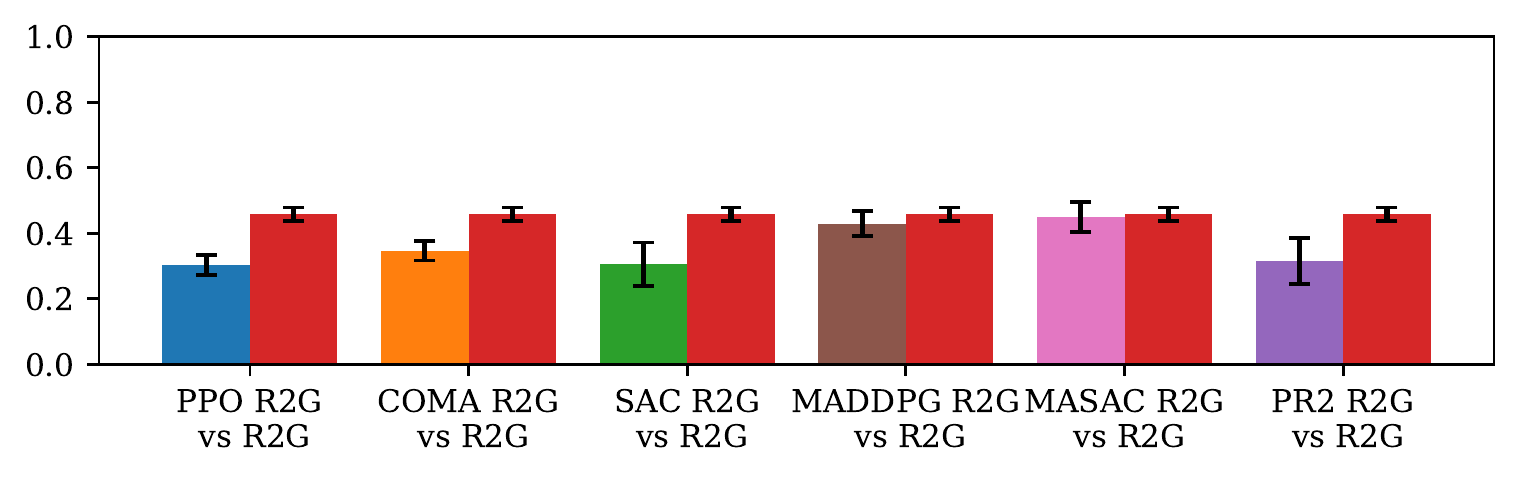}}
    {\includegraphics[width=.32\columnwidth]{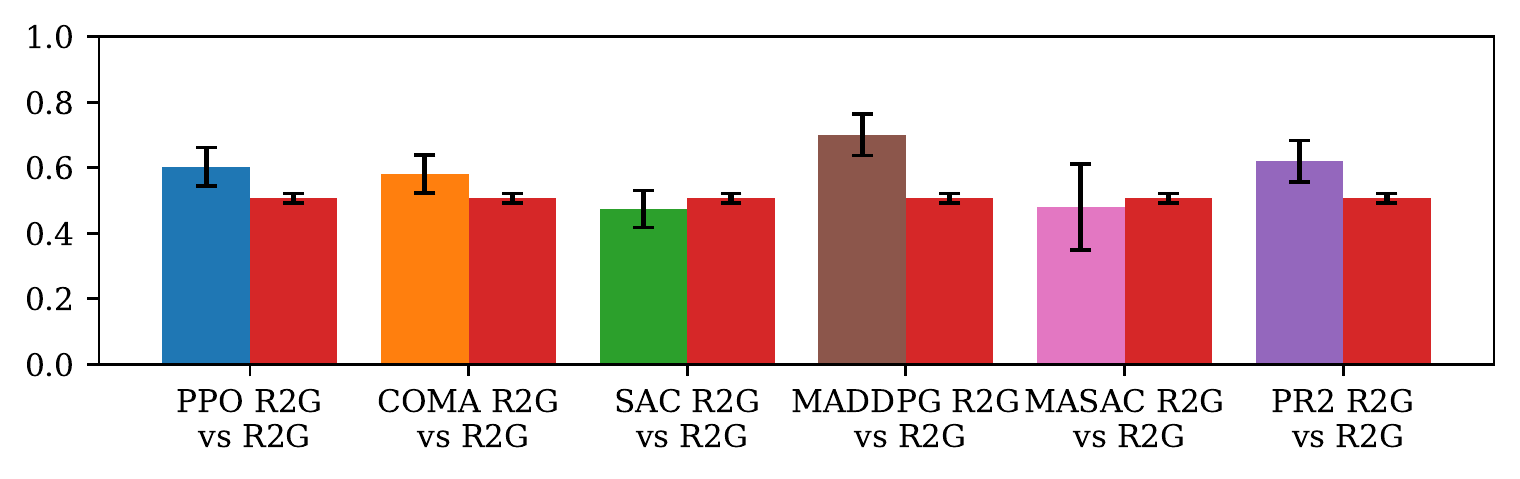}}
    \caption{Average agent return playing against the same R2G opponent}
    \label{fig:robosumo_r2g}
  \end{subfigure}
  \caption{RoboSumo: Pairwise comparison of R2G and the baseline against the same opponent. Left bar: Baseline agents; Right bar: R2G agents.}
  \label{fig:robosumo_return}
\end{figure*}

\begin{figure}[!htbp]
  \centering
  \begin{subfigure}{0.8\columnwidth}
  \centering
    {\includegraphics[width=1.\columnwidth]{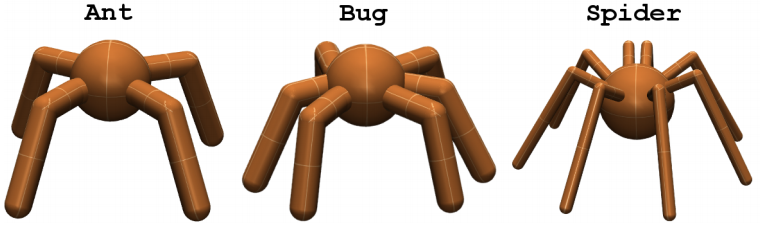}}
  \end{subfigure}
  \hspace{0.8cm}
  \begin{subfigure}{0.8\columnwidth}
    {\includegraphics[width=1.\columnwidth]{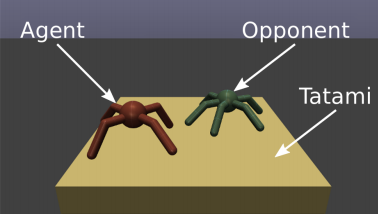}}
  \end{subfigure}
  \caption{RoboSumo: Illustration of the different types of robots (top) and the game environment (bottom).}\label{fig:robosumo}
\end{figure}

In Physical Deception, the R2G agents behave better in most comparisons and only slightly worse in some of the adversary comparisons. For Physical Deception, training good agents' behavior is harder than training adversaries as two good agents need to cooperate to confuse the adversary, where the best response for the adversary is to randomly approach one of the landmarks. Thus, the advantage of R2G on the strategic reasoning is more meaningful when training the good agents.

In Keep Away and Predator-Prey, the R2G agents also show better performance in most comparisons. As there is no clear optimal equilibrium strategies in these games, a dominate strategy is hard to achieve. While there are some comparisons where R2G is slightly worse, the overall winning rate of R2G is still higher than any other baselines.

\subsection{RoboSumo}

In order to test the scalability and performance of R2G on larger state and action spaces, we compare R2G and other baselines on the RoboSumo~\citep{robosumo} simulator, which has much larger state and action spaces with state dimensions ranging from 120 to 208, and action dimensions ranging from 8 to 16. 
RoboSumo contains a set of two-player, competitive environments where two robots are competing for pushing the opponent out of a square platform. In our experiments, three types of robots, Ant, Bug, and Spider, are trained and tested by playing against the same type of the opponent robot. 

The tested robots as well as the competing platform is shown in \cref{fig:robosumo}. 
Since the game is symmetric, we choose the agent which performs the best among the two agents trained under the same method, and use it to compare with the best agent of other methods. The results are shown in \cref{fig:robosumo_return}.
The pairwise comparison results indicate that the R2G agent performs better than all other baselines on the Ant and Bug robots, and most of the baselines except MADDPG and PR2 on the Spider robot.
The performance degeneration in Spider is likely due to a larger state and action space, where learning the basic robot motions is itself difficult given limited samples, and the advantage of strategic reasoning for R2G may not contribute as much.

\section{Conclusion}
In this paper, we proposed Recursive Reasoning Graph (R2G), a multi-agent reinforcement learning (MARL) framework that explicitly incorporates the recursive reasoning process into the training process. 
Based on existing multi-agent actor-critic algorithms, we augment each learning agent with a central actor component to model its conditional response given the strategies of its opponents. 
We demonstrated how to efficiently train the central actors using experiences stored in the replay buffer. 
Treating these central actors as nodes and their responses as messages, a recursive reasoning graph was built to efficiently calculate the optimal response of each agent given opponents actions. 
Thus, the outputs of the graph provided higher level recursion actions  which is used for training the individual policies.
In our experiments, R2G was able to converge to the ideal performance while other baselines could not in two Differential Games, which demonstrated how R2G addresses relative overgeneralization and oscillatory learning problems.
The performance of R2G on multi-player multi-state Particle World and RoboSumo environments was also significantly better than baselines. 

Adopting a graph structure for interactions between agents also brings the possibility of using graph neural networks~\cite{graphsage,gat} for dynamic interaction structure inferring, and efficient parameter sharing between agents.

\bibliography{references} 

\begin{thebibliography}{29}
\providecommand{\natexlab}[1]{#1}

\bibitem[{Al-Shedivat et~al.(2018)Al-Shedivat, Bansal, Burda, Sutskever,
  Mordatch, and Abbeel}]{robosumo}
Al-Shedivat, M.; Bansal, T.; Burda, Y.; Sutskever, I.; Mordatch, I.; and
  Abbeel, P. 2018.
\newblock Continuous adaptation via meta-learning in nonstationary and
  competitive environments.
\newblock In \emph{International Conference on Learning Representations}.

\bibitem[{Albrecht and Stone(2018)}]{albrecht2018autonomous}
Albrecht, S.~V.; and Stone, P. 2018.
\newblock Autonomous agents modelling other agents: A comprehensive survey and
  open problems.
\newblock \emph{Artificial Intelligence}, 258: 66--95.

\bibitem[{Foerster et~al.(2018)Foerster, Farquhar, Afouras, Nardelli, and
  Whiteson}]{coma}
Foerster, J.~N.; Farquhar, G.; Afouras, T.; Nardelli, N.; and Whiteson, S.
  2018.
\newblock Counterfactual multi-agent policy gradients.
\newblock In \emph{AAAI Conference on Artificial Intelligence (AAAI)}.

\bibitem[{Gupta, Egorov, and Kochenderfer(2017)}]{gupta2017cooperative}
Gupta, J.~K.; Egorov, M.; and Kochenderfer, M. 2017.
\newblock Cooperative multi-agent control using deep reinforcement learning.
\newblock In \emph{International Conference on Autonomous Agents and Multiagent
  Systems (AAMAS)}, 66--83. Springer.

\bibitem[{Haarnoja et~al.(2018)Haarnoja, Zhou, Hartikainen, Tucker, Ha, Tan,
  Kumar, Zhu, Gupta, Abbeel et~al.}]{sac}
Haarnoja, T.; Zhou, A.; Hartikainen, K.; Tucker, G.; Ha, S.; Tan, J.; Kumar,
  V.; Zhu, H.; Gupta, A.; Abbeel, P.; et~al. 2018.
\newblock Soft actor-critic algorithms and applications.
\newblock \emph{arXiv preprint arXiv:1812.05905}.

\bibitem[{Hamilton, Ying, and Leskovec(2017)}]{graphsage}
Hamilton, W.; Ying, Z.; and Leskovec, J. 2017.
\newblock Inductive representation learning on large graphs.
\newblock In \emph{Advances in Neural Information Processing Systems (NIPS)},
  1024--1034.

\bibitem[{He et~al.(2016)He, Boyd-Graber, Kwok, and
  Daum{\'e}~III}]{he2016opponent}
He, H.; Boyd-Graber, J.; Kwok, K.; and Daum{\'e}~III, H. 2016.
\newblock Opponent modeling in deep reinforcement learning.
\newblock In \emph{International Conference on Machine Learning (ICML)},
  1804--1813.

\bibitem[{Hessel et~al.(2018)Hessel, Modayil, Van~Hasselt, Schaul, Ostrovski,
  Dabney, Horgan, Piot, Azar, and Silver}]{rainbow}
Hessel, M.; Modayil, J.; Van~Hasselt, H.; Schaul, T.; Ostrovski, G.; Dabney,
  W.; Horgan, D.; Piot, B.; Azar, M.; and Silver, D. 2018.
\newblock Rainbow: Combining improvements in deep reinforcement learning.
\newblock In \emph{AAAI Conference on Artificial Intelligence (AAAI)},
  volume~32.

\bibitem[{Johannink et~al.(2019)Johannink, Bahl, Nair, Luo, Kumar, Loskyll,
  Ojea, Solowjow, and Levine}]{johannink2019residual}
Johannink, T.; Bahl, S.; Nair, A.; Luo, J.; Kumar, A.; Loskyll, M.; Ojea,
  J.~A.; Solowjow, E.; and Levine, S. 2019.
\newblock Residual reinforcement learning for robot control.
\newblock In \emph{IEEE International Conference on Robotics and Automation
  (ICRA)}, 6023--6029.

\bibitem[{Li et~al.(2019)Li, Wu, Cui, Dong, Fang, and Russell}]{m3ddpg}
Li, S.; Wu, Y.; Cui, X.; Dong, H.; Fang, F.; and Russell, S. 2019.
\newblock Robust multi-agent reinforcement learning via minimax deep
  deterministic policy gradient.
\newblock In \emph{AAAI Conference on Artificial Intelligence (AAAI)}.

\bibitem[{Lillicrap et~al.(2015)Lillicrap, Hunt, Pritzel, Heess, Erez, Tassa,
  Silver, and Wierstra}]{ddpg}
Lillicrap, T.~P.; Hunt, J.~J.; Pritzel, A.; Heess, N.; Erez, T.; Tassa, Y.;
  Silver, D.; and Wierstra, D. 2015.
\newblock Continuous control with deep reinforcement learning.
\newblock \emph{arXiv preprint arXiv:1509.02971}.

\bibitem[{Littman(1994)}]{MG}
Littman, M. 1994.
\newblock Markov games as a framework for multi-agent reinforcement learning.
\newblock In \emph{International Conference on Machine Learning (ICML)}.

\bibitem[{Lowe et~al.(2017)Lowe, Wu, Tamar, Harb, Abbeel, and
  Mordatch}]{maddpg}
Lowe, R.; Wu, Y.; Tamar, A.; Harb, J.; Abbeel, P.; and Mordatch, I. 2017.
\newblock Multi-agent actor-critic for mixed cooperative-competitive
  environments.
\newblock In \emph{Advances in Neural Information Processing Systems
  (NeurIPS)}, 6379--6390.

\bibitem[{Mohseni-Kabir, Isele, and Fujimura(2019)}]{mohseni2019interaction}
Mohseni-Kabir, A.; Isele, D.; and Fujimura, K. 2019.
\newblock Interaction-aware multi-agent reinforcement learning for mobile
  agents with individual goals.
\newblock In \emph{International Conference on Robotics and Automation (ICRA)},
  3370--3376. IEEE.

\bibitem[{Patro and Sahu(2015)}]{patro2015normalization}
Patro, S.; and Sahu, K.~K. 2015.
\newblock Normalization: A preprocessing stage.
\newblock \emph{arXiv preprint arXiv:1503.06462}.

\bibitem[{Rashid et~al.(2018)Rashid, Samvelyan, Schroeder, Farquhar, Foerster,
  and Whiteson}]{qmix}
Rashid, T.; Samvelyan, M.; Schroeder, C.; Farquhar, G.; Foerster, J.; and
  Whiteson, S. 2018.
\newblock Qmix: Monotonic value function factorisation for deep multi-agent
  reinforcement learning.
\newblock In \emph{International Conference on Machine Learning (ICML)},
  4295--4304.

\bibitem[{Schulman et~al.(2017)Schulman, Wolski, Dhariwal, Radford, and
  Klimov}]{ppo}
Schulman, J.; Wolski, F.; Dhariwal, P.; Radford, A.; and Klimov, O. 2017.
\newblock Proximal policy optimization algorithms.
\newblock \emph{arXiv preprint arXiv:1707.06347}.

\bibitem[{Shen and How(2021)}]{shen2021robust}
Shen, M.; and How, J.~P. 2021.
\newblock Robust Opponent Modeling via Adversarial Ensemble Reinforcement
  Learning.
\newblock In \emph{Proceedings of the International Conference on Automated
  Planning and Scheduling}, volume~31, 578--587.

\bibitem[{Sunehag et~al.(2018)Sunehag, Lever, Gruslys, Czarnecki, Zambaldi,
  Jaderberg, Lanctot, Sonnerat, Leibo, Tuyls et~al.}]{vdn}
Sunehag, P.; Lever, G.; Gruslys, A.; Czarnecki, W.~M.; Zambaldi, V.~F.;
  Jaderberg, M.; Lanctot, M.; Sonnerat, N.; Leibo, J.~Z.; Tuyls, K.; et~al.
  2018.
\newblock Value-Decomposition Networks For Cooperative Multi-Agent Learning
  Based On Team Reward.
\newblock In \emph{International Conference on Autonomous Agents and Multiagent
  Systems (AAMAS)}.

\bibitem[{Tan(1993)}]{indq}
Tan, M. 1993.
\newblock Multi-Agent Reinforcement Learning: Independent versus Cooperative
  Agents.
\newblock In \emph{International Conference on Machine Learning (ICML)},
  330--337.

\bibitem[{Veli{\v{c}}kovi{\'c} et~al.(2018)Veli{\v{c}}kovi{\'c}, Cucurull,
  Casanova, Romero, Li{\`o}, and Bengio}]{gat}
Veli{\v{c}}kovi{\'c}, P.; Cucurull, G.; Casanova, A.; Romero, A.; Li{\`o}, P.;
  and Bengio, Y. 2018.
\newblock Graph Attention Networks.
\newblock In \emph{International Conference on Learning Representations}.

\bibitem[{Vinyals et~al.(2019)Vinyals, Babuschkin, Czarnecki, Mathieu, Dudzik,
  Chung, Choi, Powell, Ewalds, Georgiev et~al.}]{vinyals2019grandmaster}
Vinyals, O.; Babuschkin, I.; Czarnecki, W.~M.; Mathieu, M.; Dudzik, A.; Chung,
  J.; Choi, D.~H.; Powell, R.; Ewalds, T.; Georgiev, P.; et~al. 2019.
\newblock Grandmaster level in {S}tar{C}raft {II} using multi-agent
  reinforcement learning.
\newblock \emph{Nature}, 575(7782): 350--354.

\bibitem[{Von Der~Osten, Kirley, and Miller(2017)}]{von2017minds}
Von Der~Osten, F.~B.; Kirley, M.; and Miller, T. 2017.
\newblock The Minds of Many: Opponent Modeling in a Stochastic Game.
\newblock In \emph{International Joint Conference on Artificial Intelligence
  (IJCAI)}, 3845--3851.

\bibitem[{Wei and Luke(2016)}]{wei2016lenient}
Wei, E.; and Luke, S. 2016.
\newblock Lenient learning in independent-learner stochastic cooperative games.
\newblock \emph{Journal of Machine Learning Research}, 17(1): 2914--2955.

\bibitem[{Wei et~al.(2018)Wei, Wicke, Freelan, and Luke}]{masoftq}
Wei, E.; Wicke, D.; Freelan, D.; and Luke, S. 2018.
\newblock Multiagent soft {Q}-learning.
\newblock In \emph{AAAI Spring Symposium Series}.

\bibitem[{Wen et~al.(2020)Wen, Yang, Luo, and Wang}]{gr2}
Wen, Y.; Yang, Y.; Luo, R.; and Wang, J. 2020.
\newblock Modelling Bounded Rationality in Multi-Agent Interactions by
  Generalized Recursive Reasoning.
\newblock In \emph{International Joint Conference on Artificial Intelligence
  (IJCAI)}.

\bibitem[{Wen et~al.(2018)Wen, Yang, Luo, Wang, and Pan}]{pr2}
Wen, Y.; Yang, Y.; Luo, R.; Wang, J.; and Pan, W. 2018.
\newblock Probabilistic Recursive Reasoning for Multi-Agent Reinforcement
  Learning.
\newblock In \emph{International Conference on Learning Representations}.

\bibitem[{Wright and Leyton-Brown(2010)}]{wright2010beyond}
Wright, J.; and Leyton-Brown, K. 2010.
\newblock Beyond equilibrium: Predicting human behavior in normal-form games.
\newblock In \emph{AAAI Conference on Artificial Intelligence (AAAI)}.

\bibitem[{Yang et~al.(2019)Yang, Nakhaei, Isele, Fujimura, and
  Zha}]{yang2019cm3}
Yang, J.; Nakhaei, A.; Isele, D.; Fujimura, K.; and Zha, H. 2019.
\newblock {CM}3: Cooperative Multi-goal Multi-stage Multi-agent Reinforcement
  Learning.
\newblock In \emph{International Conference on Learning Representations}.

\end{thebibliography}

\newpage
\appendix


\section{Experiment Details}
\subsection{Differential Games}
\label{appdix:differential}
The Differential Games contain two single-state two-player games, the Zero Sum and the Max of Two. Each game has a one-hot state space indicating the identity of the player, and a scalar action space of $[-1,1]$. The reward function of each game is:
\begin{itemize}
    \item Zero Sum: $r^1(a^1,a^2)=-r^2(a^1,a^2)=10a^1\cdot 10a^2$.
    \item Max of Two: $r^1(a^1,a^2)=r^2(a^1,a^2)=\max(f_1,f_2)$, where $f_1=0.8\times [-(\frac{a^1+0.5}{0.3})^2-(\frac{a^2+0.5}{0.3})^2]$ and $f_2=1.0\times [-(\frac{a^1-0.5}{0.1})^2-(\frac{a^2-0.5}{0.1})^2]+10$. 
\end{itemize}

We use a multi-layer perceptron (MLP) with 2 hidden layers each with 16 hidden units for all the Q networks, policies, as well as central actors. We use a diagonal Gaussian as the output distribution for policies. Each algorithm is trained with 1000 epochs with 100 exploration steps per epoch and a batch size of 256. We use a learning rate of \num{e-3} for the Q and value network, and \num{e-4} for the policies. 
To encourage exploration, for PPO and COMA, we add an additional entropy loss in the policy loss with a weight of 0.01. 
For MADDPG, we implement the Ornstein-Uhlenbeck process for action noise as described in \citet{ddpg}.
For SAC, MASAC, PR2, and R2G, we implemented the automatic entropy adjustment as introduced in \citet{sac}.
Our implementation is based on the \textit{rlkit}\footnote{https://github.com/vitchyr/rlkit} open source reinforcement learning package. 

\subsection{Particle World}
\label{appdix:particle}
We use similar network structures as well as other hyper-parameters as in the Differential Games with hidden dimensions increased to 64 and number of exploration steps per epoch increased to \num{e3}. Each training is repeated 5 times with different random seeds and the average performance is reported. The maximum trajectory length of the environment is set to 25.

\begin{figure}[!htbp]
  \centering
  \begin{subfigure}{0.7\columnwidth}
  \centering
    {\includegraphics[width=1.\columnwidth]{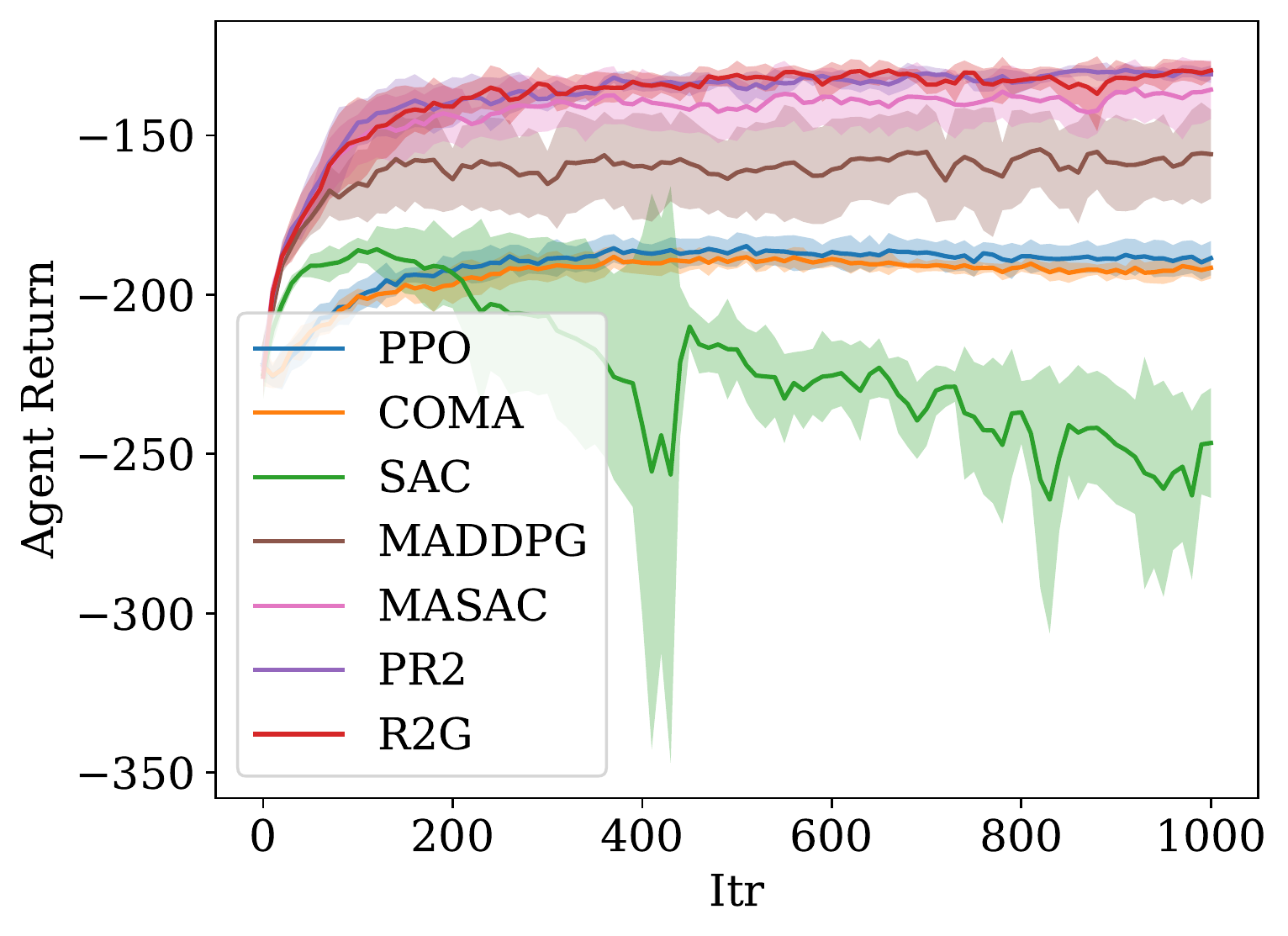}}
    \caption{Cooperative Navigation}\label{fig:spread_curve}
  \end{subfigure}
  \begin{subfigure}{0.7\columnwidth}
  \centering
    {\includegraphics[width=1.\columnwidth]{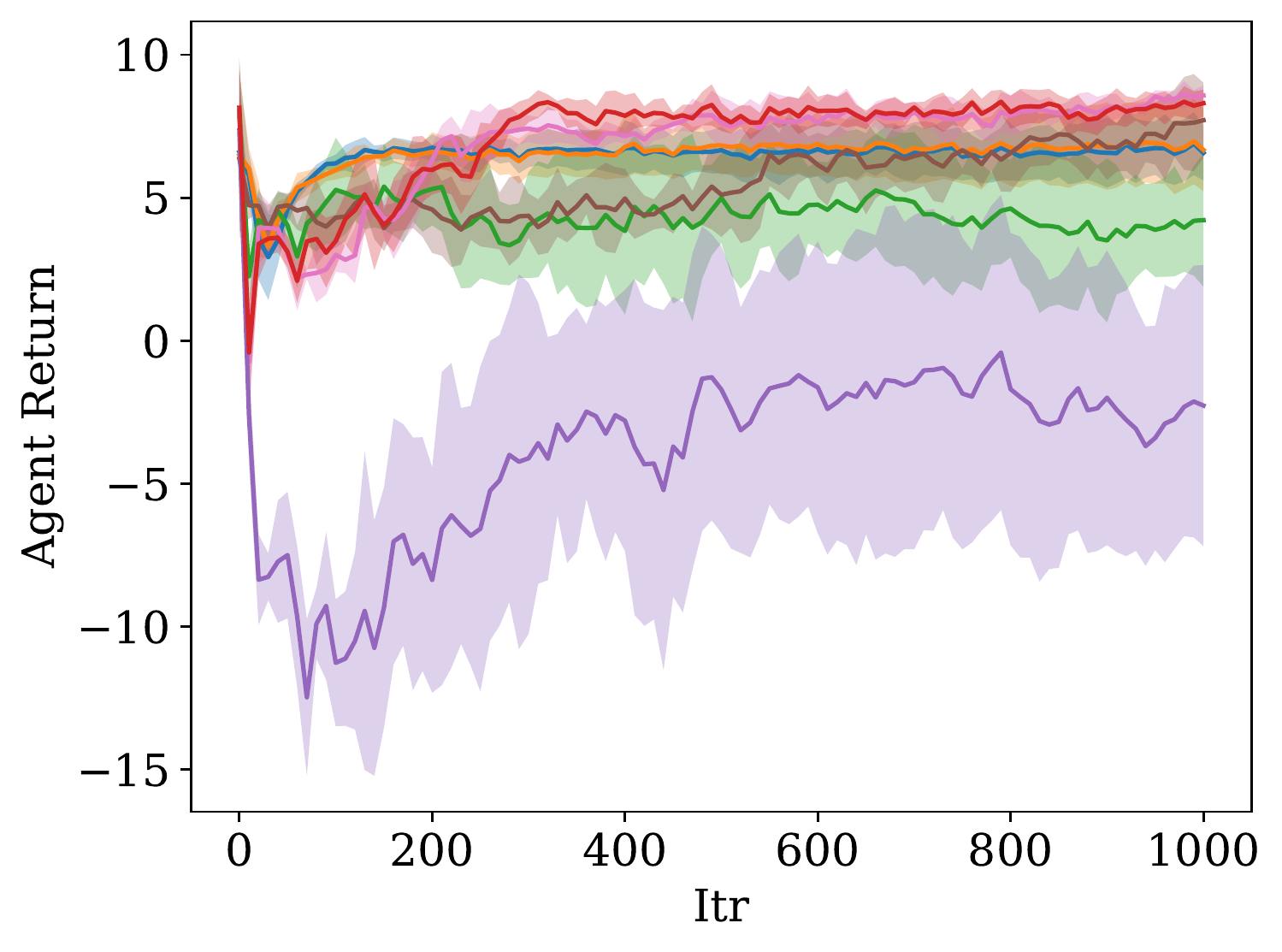}}
    \caption{Physical Deception}\label{fig:adv_curve}
  \end{subfigure}
  \begin{subfigure}{0.7\columnwidth}
  \centering
    {\includegraphics[width=1.\columnwidth]{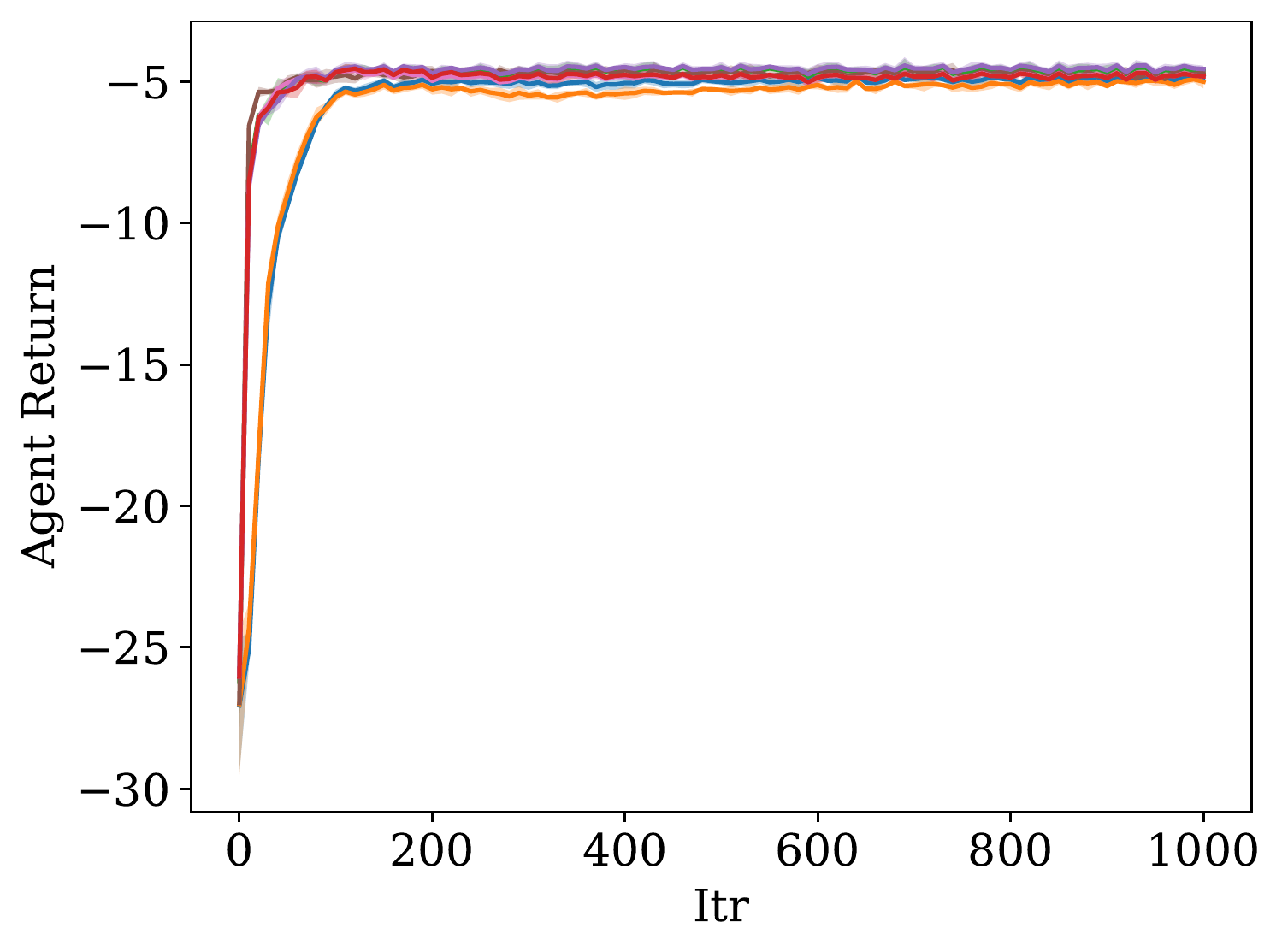}}
    \caption{Keep Away}\label{fig:push_curve}
  \end{subfigure}
  \begin{subfigure}{0.7\columnwidth}
  \centering
    {\includegraphics[width=1.\columnwidth]{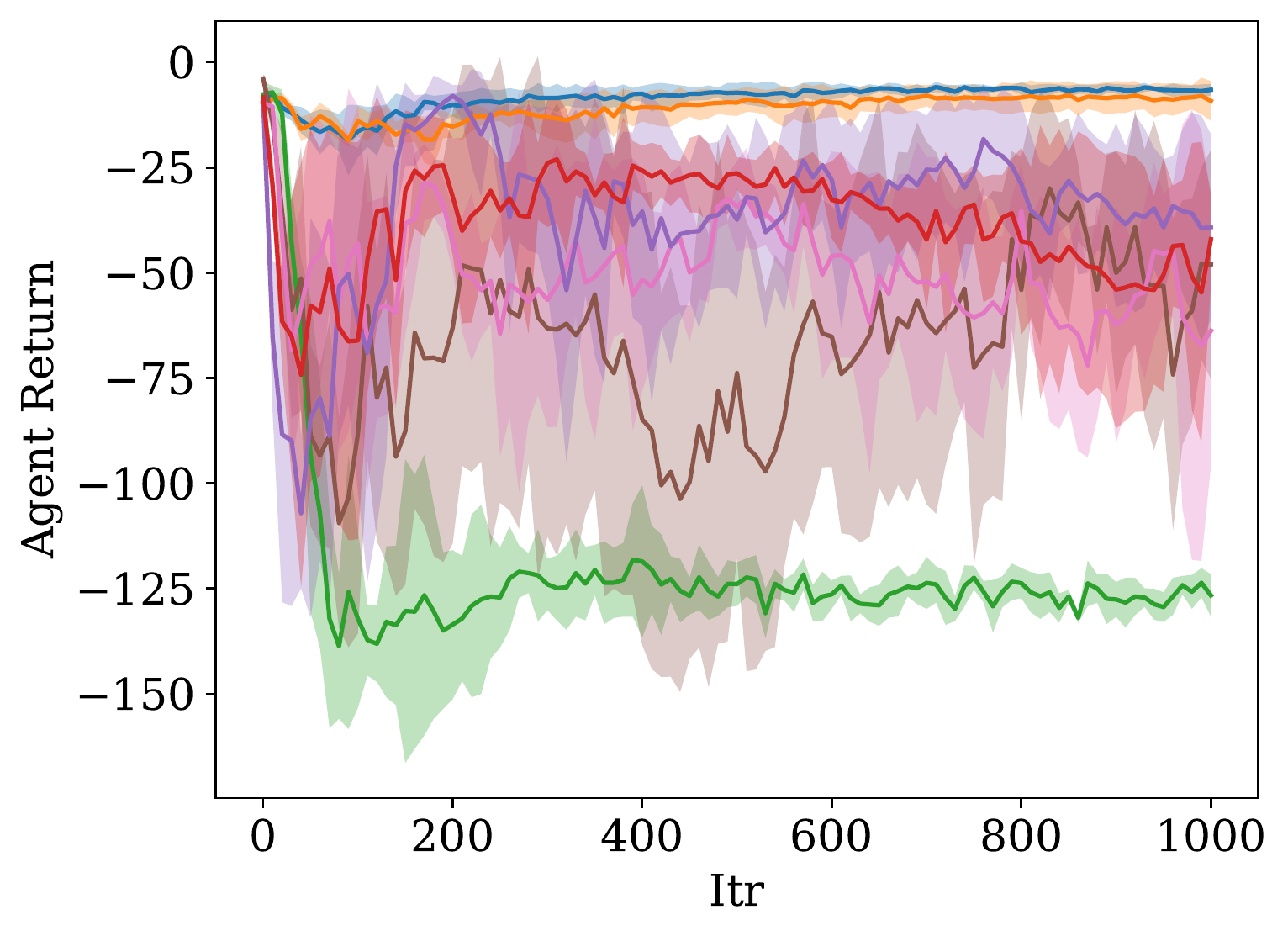}}
    \caption{Predator-Prey}\label{fig:tag_curve}
  \end{subfigure}
  \caption{Particle World: Learning curves of different methods. The agent return is calculated by playing the agent against the opponents that are trained together under the same method, and is averaged by 40 trajectories. This could NOT provide meaningful comparisons for performance of different methods on competitive games.}
  \label{fig:particle_curve}
\end{figure}

\Cref{fig:particle_curve} illustrates the learning curves of different methods. 
For Cooperative Navigation, all agents share the same reward function. The learning curves show a similar comparison results  as in \cref{fig:spread_return}.
For Physical Deception, Keep Away, and Predator-Prey, the adversaries rewards are omitted since these games are nearly zero-sum.
Notice that since the evaluation return is computed by playing the agent against the opponents trained together under the same method, this could NOT provide a meaningful comparison of different methods on competitive games. 

\subsection{RoboSumo}
\label{appdix:robosumo}
As the environment becomes more complex, we use larger networks with 3 hidden layers and 64 hidden units. We increase the number of exploration steps per epoch as well as the total epoch number to \num{2e3}. The maximum trajectory length of the environment is set to 100. We use a scaling factor of \num{0.01} on the reward to stabilize the training.

\begin{figure}[!htbp]
  \centering
  \begin{subfigure}{0.7\columnwidth}
  \centering
    {\includegraphics[width=1.\columnwidth]{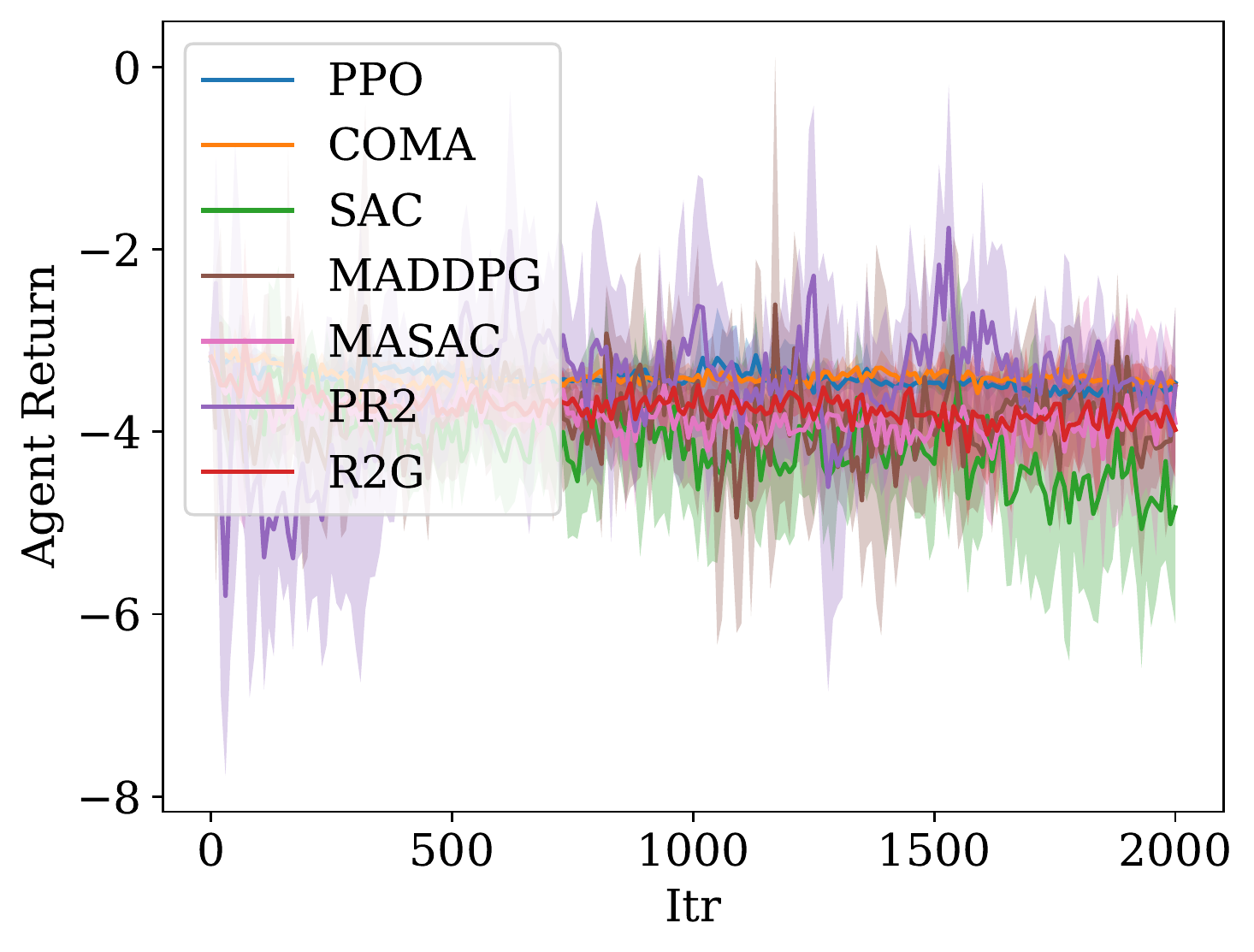}}
    \caption{Spider}\label{fig:ant_curve}
  \end{subfigure}
  \begin{subfigure}{0.7\columnwidth}
  \centering
    {\includegraphics[width=1.\columnwidth]{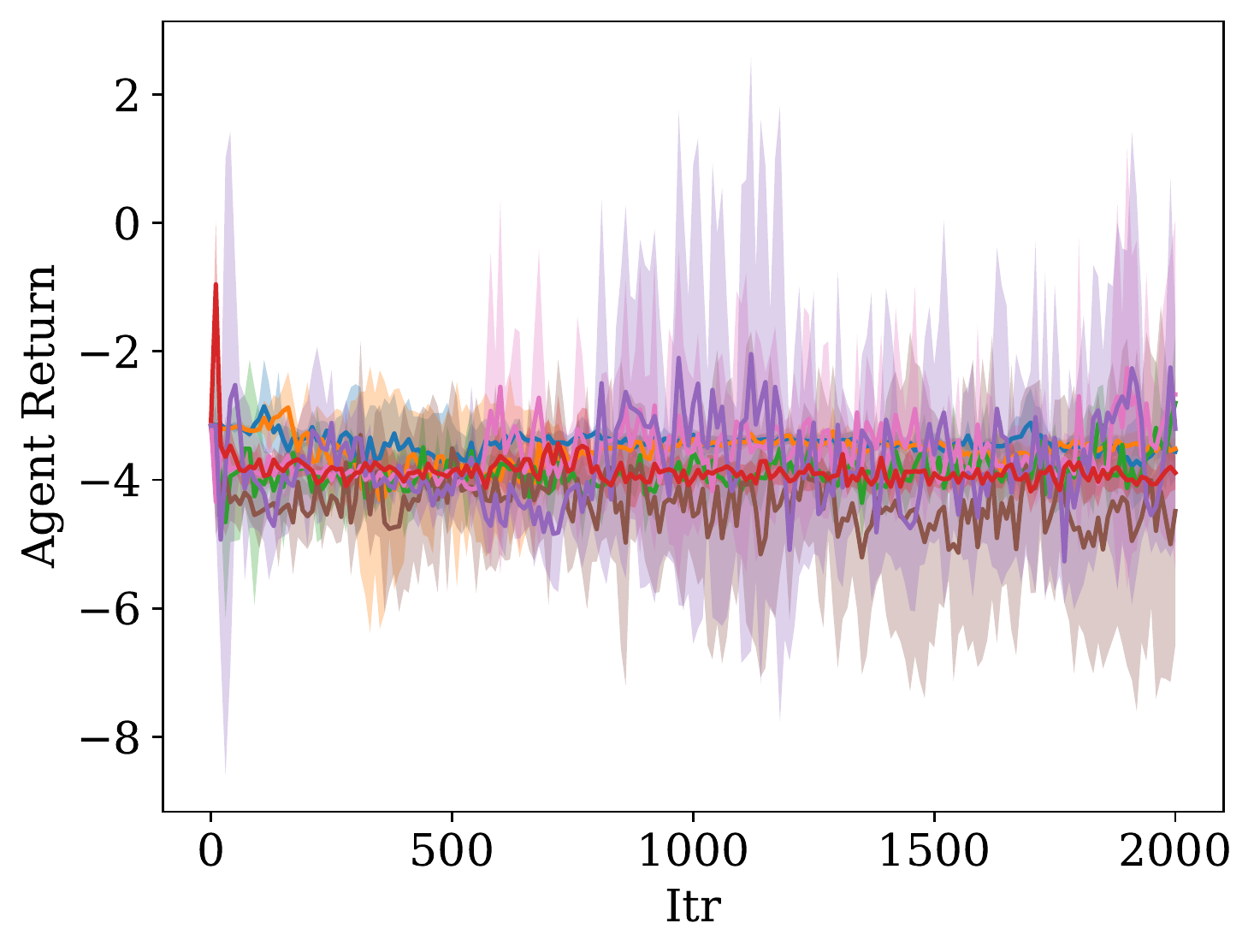}}
    \caption{Bug}\label{fig:bug_curve}
  \end{subfigure}
  \begin{subfigure}{0.7\columnwidth}
  \centering
    {\includegraphics[width=1.\columnwidth]{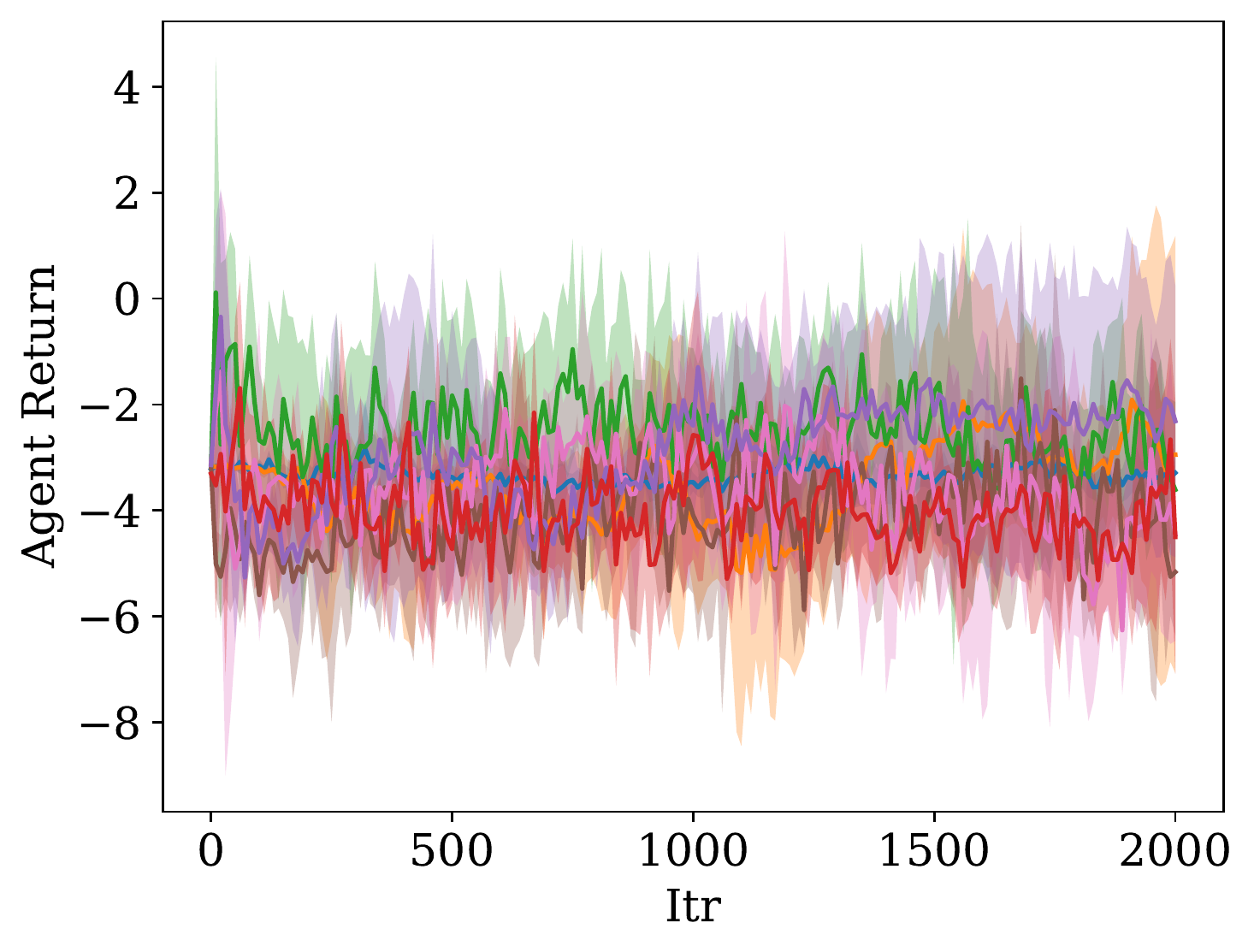}}
    \caption{Spider}\label{fig:spider_curve}
  \end{subfigure}
  \caption{RoboSumo: Learning curves of different methods. The agent return is calculated by playing the agent against the opponents that are trained together under the same method, and is averaged by 20 trajectories. This could NOT provide meaningful comparisons for performance of different methods.}
  \label{fig:robosumo_curve}
\end{figure}

\Cref{fig:robosumo_curve} illustrates the learning curves of different methods. 
Similarly, since the evaluation return is computed by playing the agent against the opponents trained together under the same method, this could NOT provide a meaningful comparison of different methods. 

\section{Ablation Study: Different Recursion Levels}
\label{appdix:k}
The R2G performance shown in previous results are trained with a recursion level of 1. In this section, we provide the comparison results of using different recursion levels. With the help of the central actors and recursive reasoning graphs, the com[utation complexity sclaes linearly with the level of recursions, $k$. Notice that, the R2G with level-0 recursion is the same as MASAC. 

\begin{figure}
  \centering
  \includegraphics[width=0.9\columnwidth]{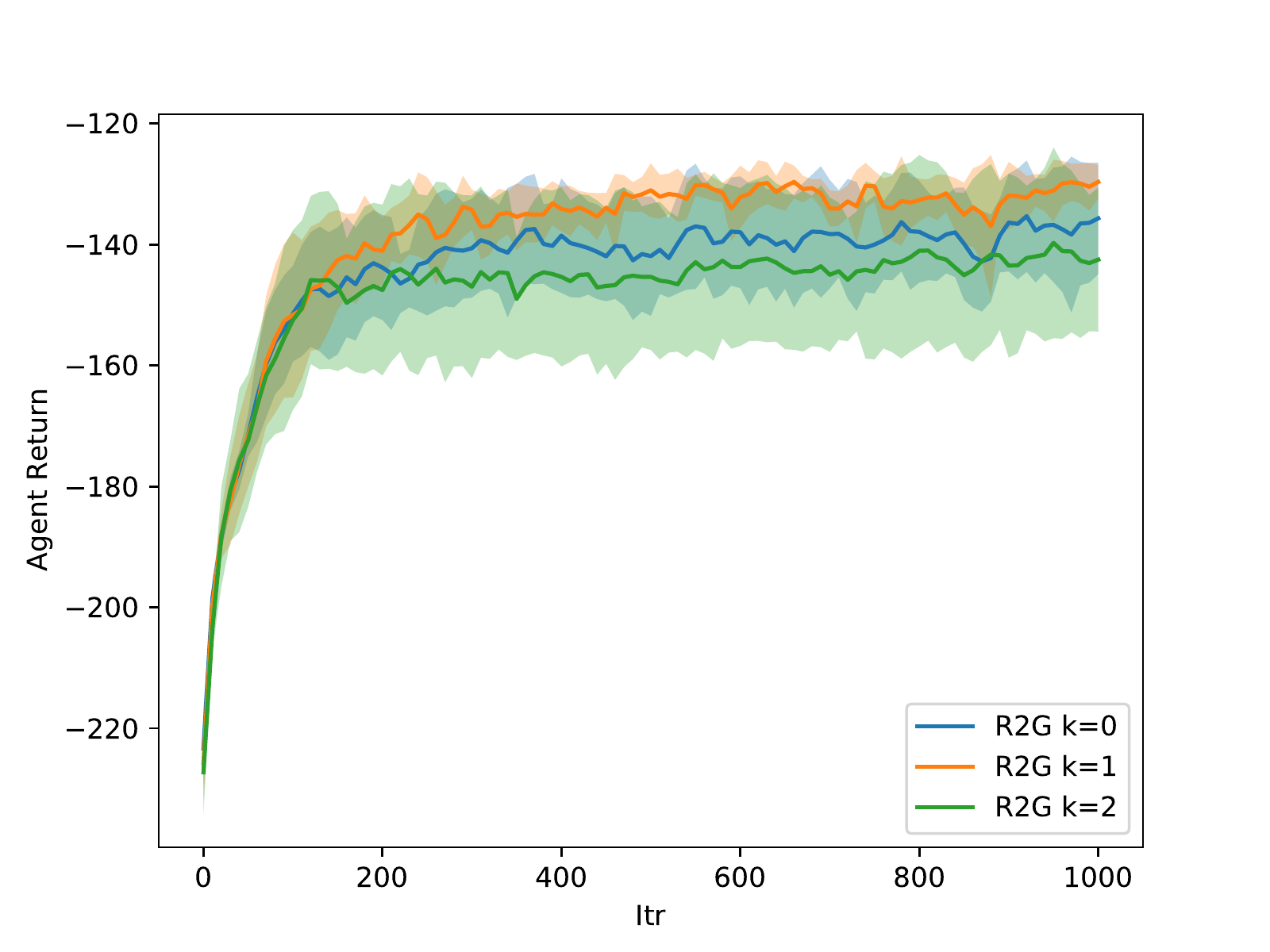}
  \caption{Cooperative Navigation: Comparison of different recursion levels of R2G.}
  \label{fig:spread_k}
\end{figure}

\Cref{fig:spread_k} shows the learning curve of R2G with $k=0,1,2$ in the Cooperative Navigation environment. In this environment, the level-1 recursions behaves slightly better than level-0, which is again slightly better than level-2. 

\begin{figure*}[!htbp]
  \centering
  \begin{subfigure}{0.55\columnwidth}
  \centering
    {\includegraphics[width=1.\columnwidth]{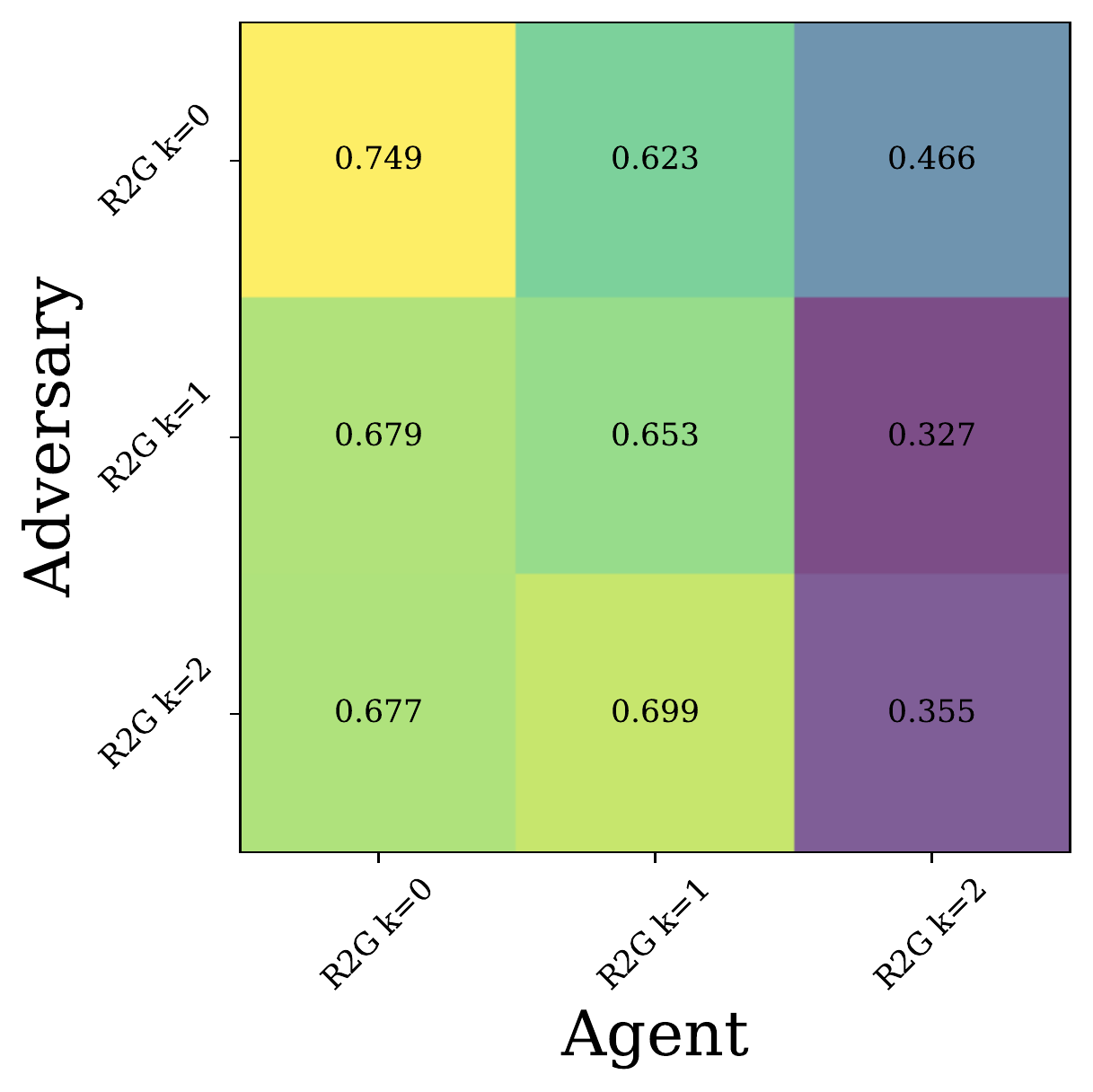}}
    {\includegraphics[width=1.\columnwidth]{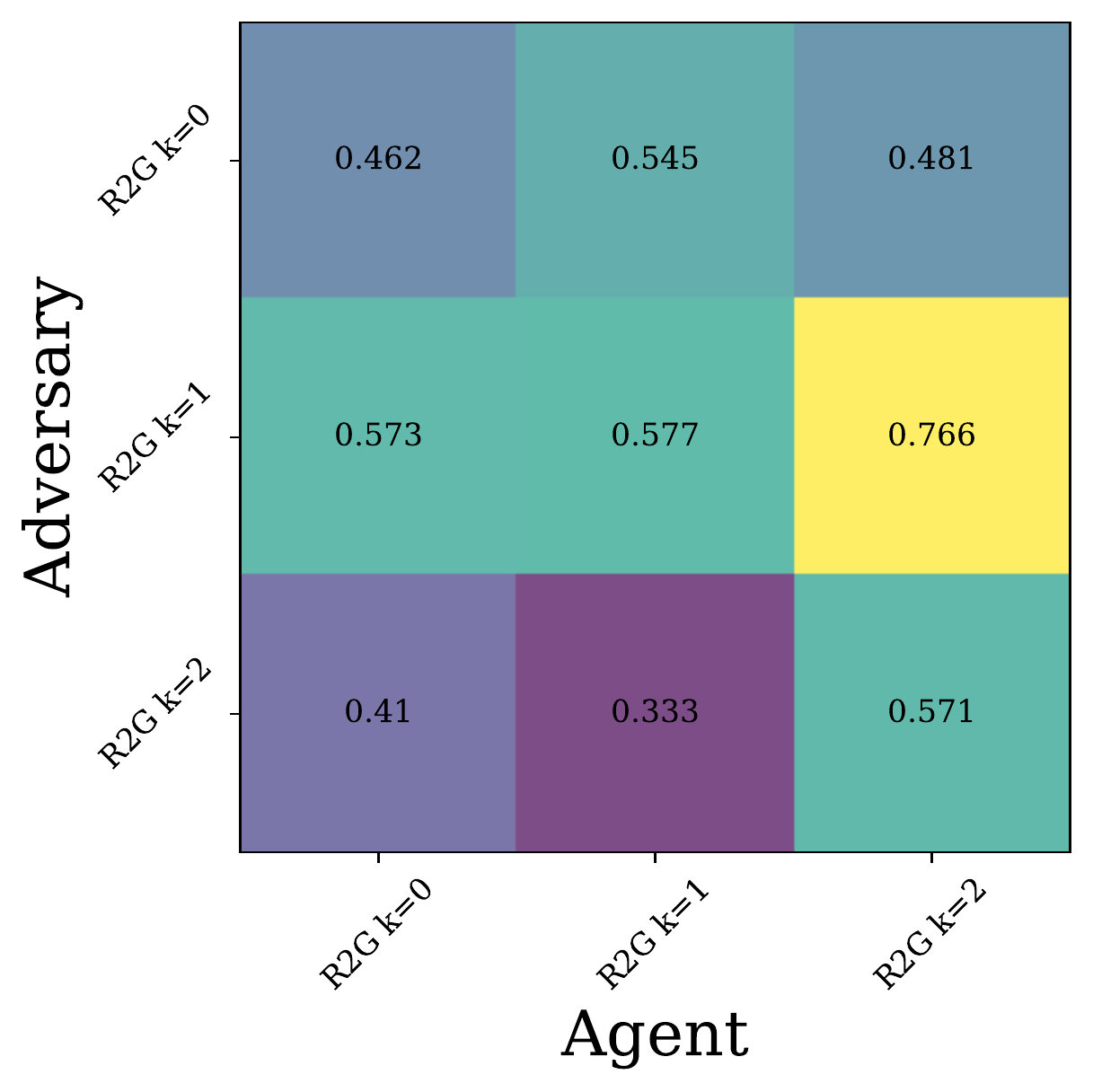}}
    \caption{Physical  \\\hspace*{0.4cm}
    Deception}\label{fig:adversary_k}
  \end{subfigure}
  \begin{subfigure}{0.55\columnwidth}
  \centering
    {\includegraphics[width=1.\columnwidth]{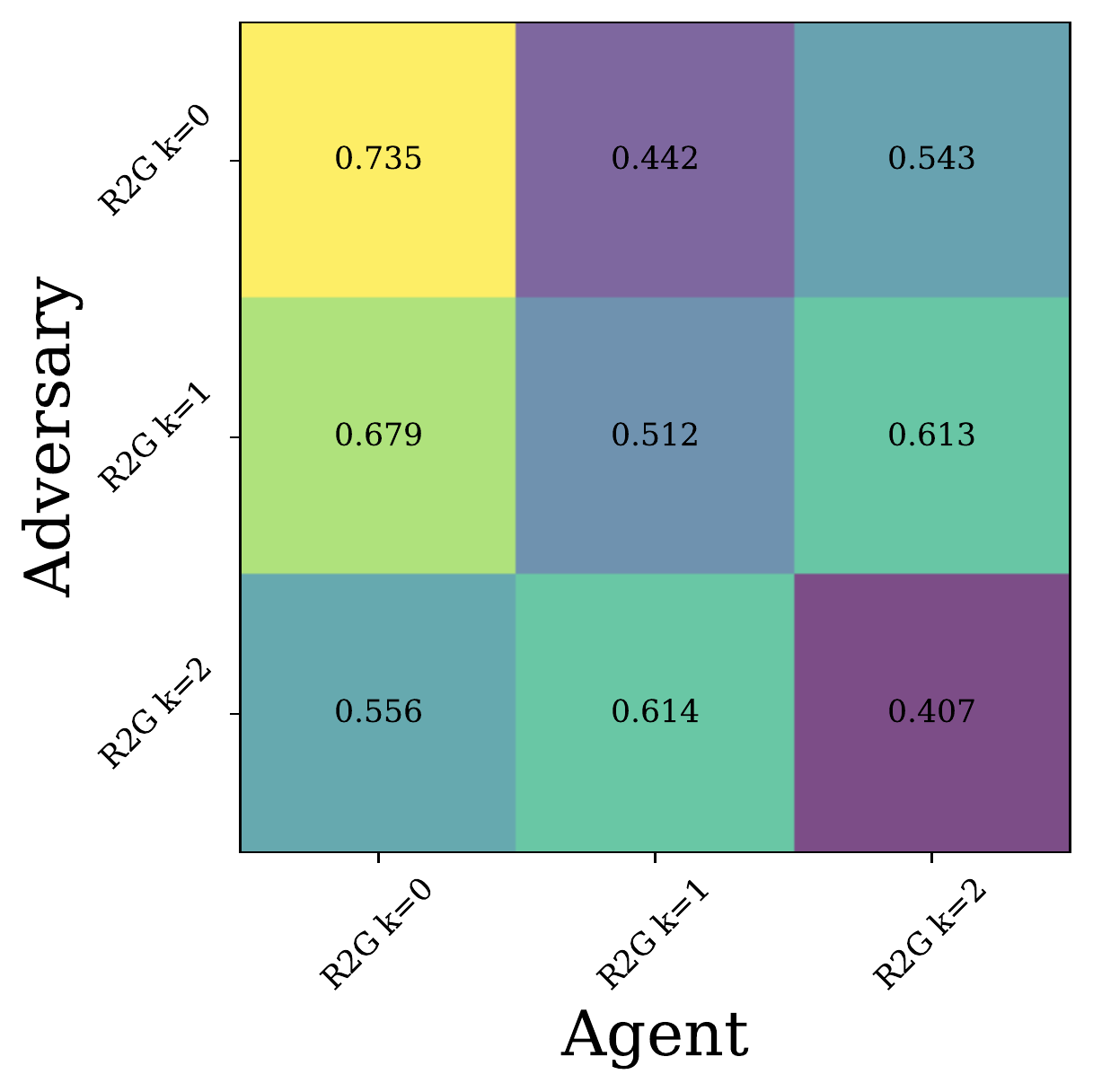}}
    {\includegraphics[width=1.\columnwidth]{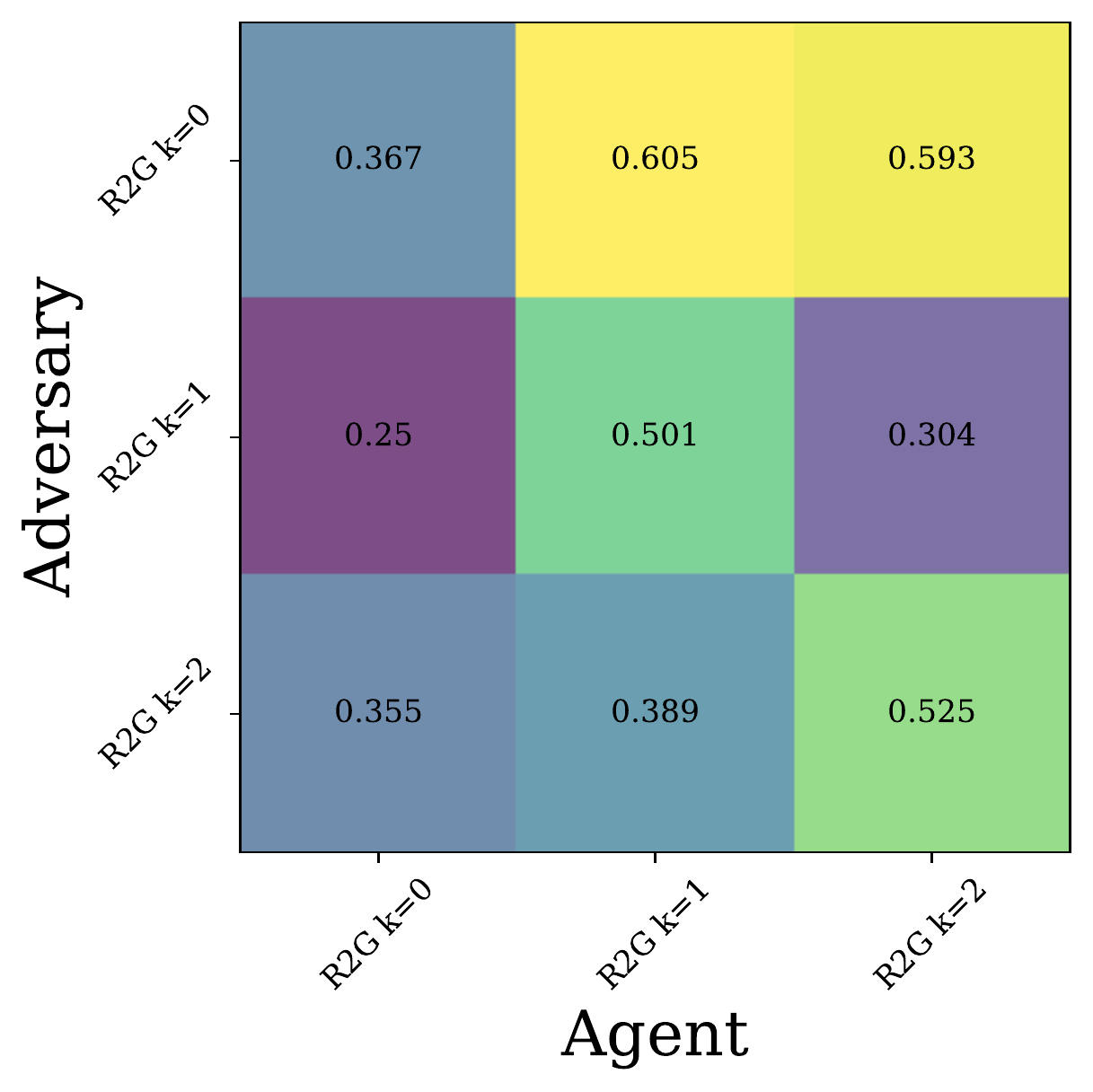}}
    \caption{Keep  \\\hspace*{0.4cm}
    Away}\label{fig:push_k}
  \end{subfigure}
  \begin{subfigure}{0.55\columnwidth}
  \centering
    {\includegraphics[width=1.\columnwidth]{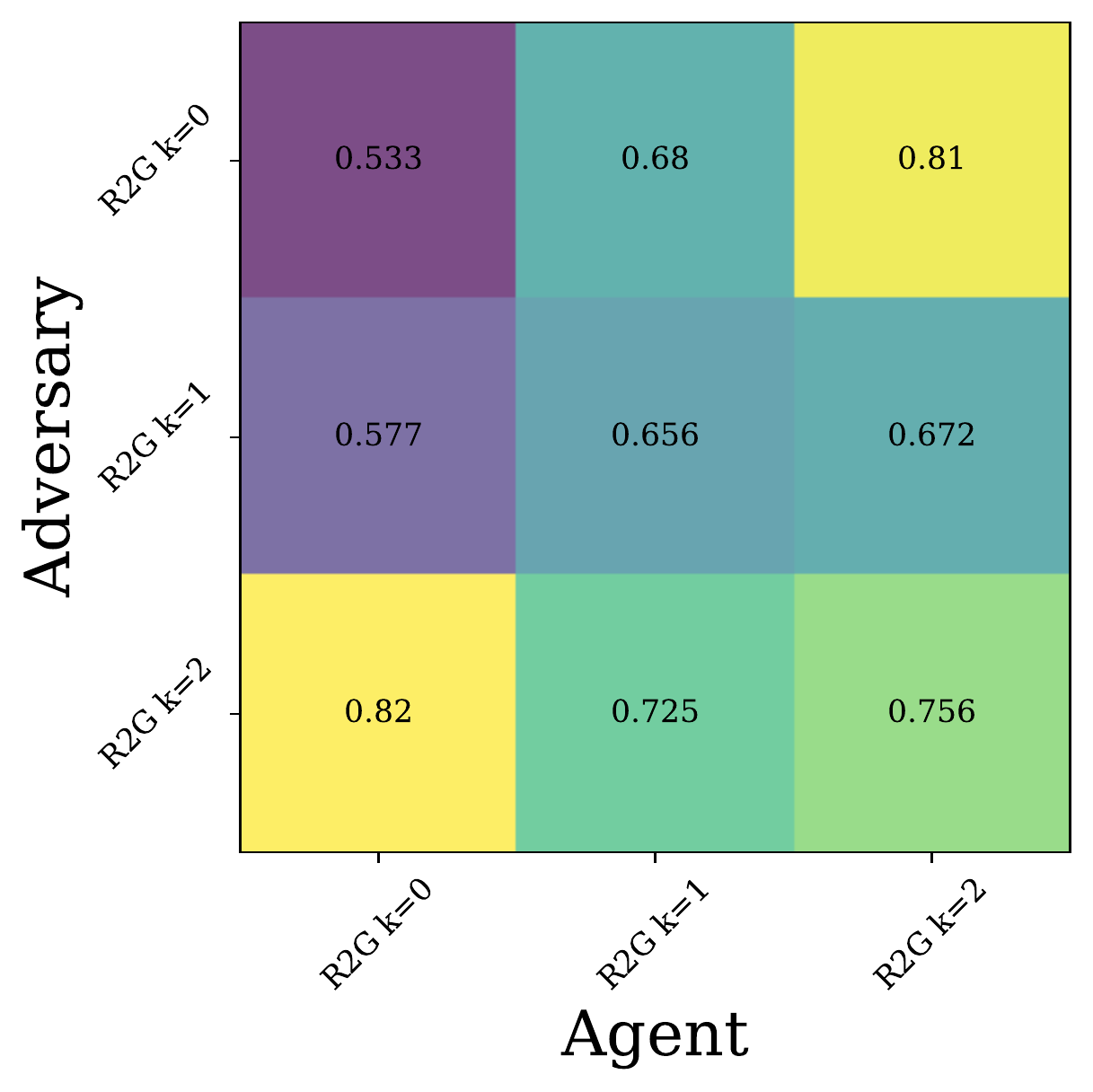}}
    {\includegraphics[width=1.\columnwidth]{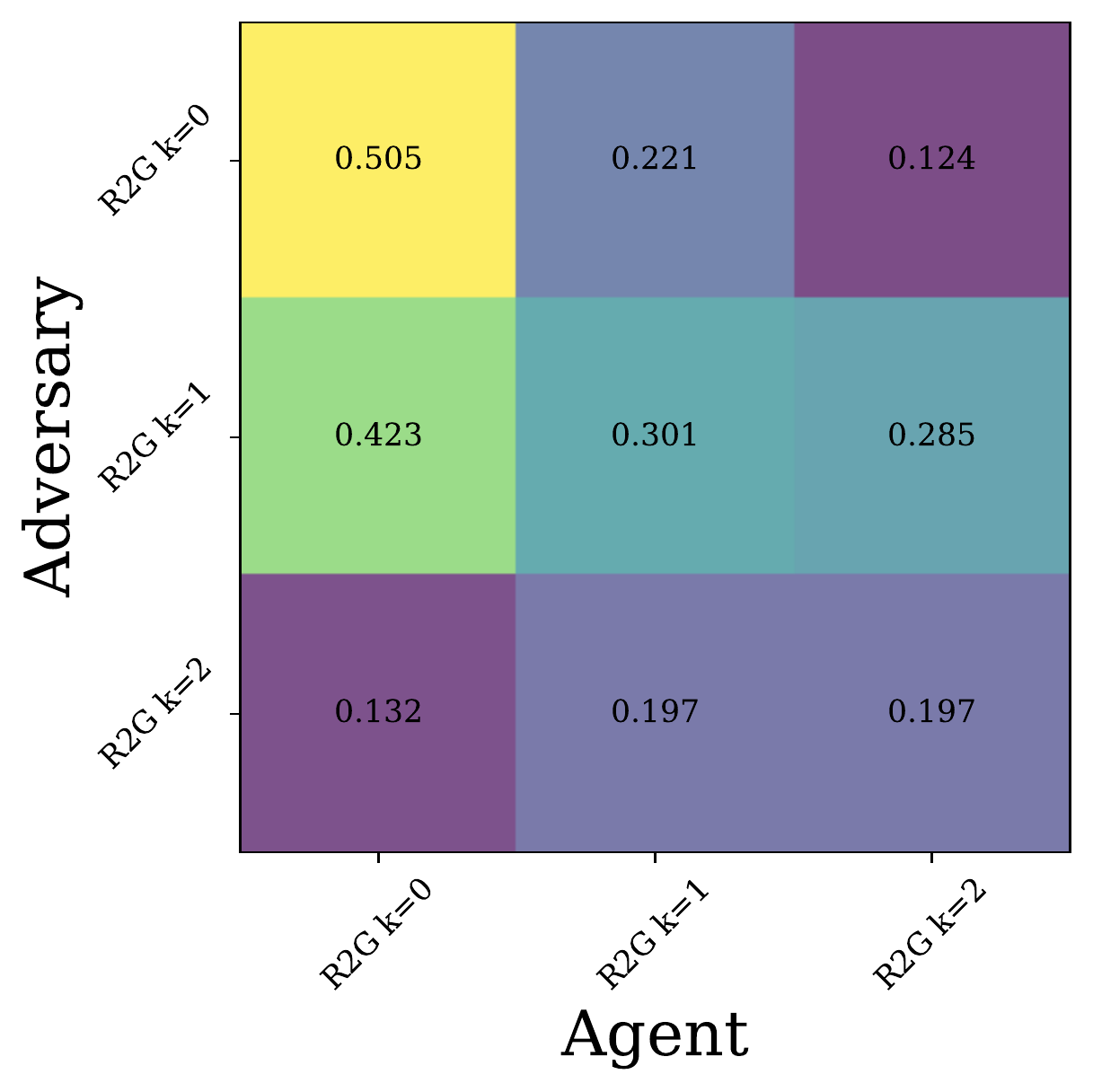}}
    \caption{Predator \\\hspace*{0.4cm}
    Prey}\label{fig:tag_k}
  \end{subfigure}
  \caption{Particle World: Comparison of different recursion levels of R2G in competitive games. Top row: agent returns. Bottom row: adversary returns.}
  \label{fig:particle_k}
\end{figure*}

\Cref{fig:particle_k} shows the return matrices between R2G with different levels on the three competitive particle environments. The row of a matrix indicates the same adversary playing against different agents, and the column of a matrix indicates the same agent playing against different adversaries. In general, the level-0 and level-1 recursions behave better than level-2 recursions, while none of them is dominating.

In general, we expect R2G with level-1 recursion performs well enough for most games for several reasons: 
First, level-1 recursion resembles the solution concept of evaluating the ego action by assuming others are optimally responding to it, which is similar as the minimax optimization in two-player zero-sum games and joint maximization in cooperative games;
Second, while at each training loop we only do 1-level recursion, over multiple training loops, agents' base policies could be regarded as higher level strategies from previous loops.
The same reasoning could be applied to higher levels of recursion, where the level of recursions could be regarded as how many steps we would like to ``look ahead" in the level-0 learning dynamics.
However, using higher levels of recursion would increase the variance of this ``looking ahead" due to sampling from stochastic policies, and it does not give guarantees on improving the learning dynamics.
In fact, there does not exist a dominating recursion level that is better than others in all games, and we focus on the level-1 recursion due to the above reasons.

\section{Convergence of R2G Value Iteration}
Here we provide the convergence proof of value iteration for level-1 R2G in cooperative games, following a similar approach as in single agent SAC~\cite{sac} and PR2~\cite{pr2}.
Assuming that at each iteration, $\pi^i_c$ is trained to optimize \cref{eq:Jpic}, which gives us $\pi^i_c(s,a^{-i})=\arg\max_{a^i}Q^i(s,a^i,a^{-i})$.
Using the fact that $r^i=r^j$ and thus $Q^i=Q^j$ in cooperative games, the value iteration operator, $\Gamma^{\pi_i}$, for $Q^i$ derived from \cref{eq:JQ} with $k=1$ could be written as 
\begin{equation}
 \begin{split}
  \Gamma^{\pi_i} Q^i(s,a^i,a^{-i}) = &r^i(s,a^i,a^{-i})+\\
  &\gamma \mathbb{E}_{s',a'^i\sim T,\pi^i} \max_{a'^{-i}}Q^i(s',a'^i,a'^{-i}) \\
  &-\alpha^i\log \pi^i(a'^i|s')   
 \end{split}
\end{equation}
Define the norm on Q-values as $\|Q^i_1-Q^i_2\| \triangleq \max_{s,a^i,a^{-i}}|Q^i_1(s,a^i,a^{-i})-Q^i_2(s,a^i,a^{-i})|$, and let $\epsilon=\|Q^i_1-Q^i_2\|$.
Then we have $|\max_{a'^{-i}}Q_1^i(s',a'^i,a'^{-i})-\max_{a'^{-i}}Q_2^i(s',a'^i,a'^{-i})| \leq \epsilon$, and thus 
\begin{equation}
    \begin{split}
        \|\Gamma^{\pi_i} Q^i_1-\Gamma^{\pi_i} Q^i_2\| &=
        \|\gamma\mathbb{E}\max_{a'^{-i}}Q_1^i(s',a'^i,a'^{-i})\\
        &\:\:\:\:\:\:\:\:\:
        -\max_{a'^{-i}}Q_2^i(s',a'^i,a'^{-i})\|\\
        &\leq\|\gamma\mathbb{E}\epsilon\|
        =\gamma\|Q^i_1-Q^i_2\|
    \end{split}
\end{equation}
Thus $\Gamma^{\pi_i}$ is a contraction mapping.

\section{Limitations}
\label{appdix:limit}
While R2G has shown great performance on most of the environments tested in the paper, there are certain game structures where R2G does not help. For example, in a two-player cooperative game with two optimal points $(a^{1,*},a^{2,*})$ and $(a^{1,**},a^{2,**})$, where $r(a^{1,*},a^{2,*})=r(a^{1,**},a^{2,**})\geq r(a^1,a^2),\forall a^1,a^2 \in A$, the two R2G agents may converge to different optimal points and the actual performance is then suboptimal. E.g., the player 1 may converge to play $a^{1,*}$ where $a^{2,(1)}=a^{2,*}=\pi^2_c(a^{1,*})$. However, player 2 may converge to $a^{2,**}$ where $a^{1,(1)}=a^{1,**}=\pi^1_c(a^{2,**})$.
Games with multiple optimal points with the exact same return are very rare in practice, and R2G could be easily adapted to it if the reward structure is known prior to the training.
However, we do realize that R2G is not suitable for all game structures but a promising additional candidate in the current deep MARL algorithm landscape.

\section{Computation Platform and Cost}
\label{appdix:compute}
All experiments were performed on a 2.6GHz, 28 core Intel(R) Xeon(R) E5-2690 v4 CPU. 
Experiments on Differential Games for each method take approximately 1 hour.
Experiments on Particle World take approximately 1 day.
Experiments on RoboSumo take approximately 3 days.

\end{document}